\documentclass{article}

\usepackage{PRIMEarxiv}

\usepackage[utf8]{inputenc} 
\usepackage[T1]{fontenc}    
\usepackage{hyperref}       
\usepackage{url}            
\usepackage{booktabs}       
\usepackage{amsfonts}       
\usepackage{nicefrac}       
\usepackage{microtype}      
\usepackage{lipsum}
\usepackage{fancyhdr}       
\usepackage{graphicx}       
\graphicspath{{media/}}     

\usepackage{subcaption}
\usepackage{multirow}
\usepackage{tcolorbox}
\usepackage{graphicx}
\usepackage{subcaption}
\usepackage{ulem}
\usepackage{longtable}

\title{ARCANA: A Benchmark for Abstraction and Analogical Reasoning
}

\author{%
  Ruqi Chen \\
  University of Science and Technology of China \\
  Hefei, China \\
  State Key Laboratory of General Artificial Intelligence,
  BIGAI, Beijing, China \\
  \texttt{crq2020\_08@mail.ustc.edu.cn}
  \And
  Zilei Wang\thanks{Corresponding authors.} \\
  University of Science and Technology of China \\
  Hefei, China \\
  \texttt{zlwang@ustc.edu.cn}
  \And
  Wei Wang\footnotemark[1] \\
  University of Science and Technology Beijing \\
  Beijing, China \\
  State Key Laboratory of General Artificial Intelligence,
  BIGAI, Beijing, China \\
  \texttt{wangwei@nlpr.ia.ac.cn}
}

\begin{document}
\maketitle

\begin{abstract}
Analogical reasoning is widely considered a fundamental component of human intelligence. Few benchmarks are designed to explicitly test far analogical reasoning—reasoning between domains that are superficially dissimilar yet share deep relational or structural similarities. In this work, we introduce an analogical reasoning benchmark called Abstraction and Reasoning Corpus of ANAlogy (ARCANA). The tasks of ARCANA require performing far analogical mappings from real-world objects or physical phenomena to abstract grid-based transformations, different from the well-known ARC-AGI benchmark which primarily emphasized innate human priors.
We validate ARCANA through a human study, demonstrating that humans solve these tasks reliably. In contrast, evaluations across a range of large language models show a distinct performance gap compared to humans, revealing systematic failure modes that remain challenging for current models. To address this gap, we propose \textsc{ConceptSIR}, a reasoning framework based on concept slippage, introspection, and reflection. Experiments demonstrate that \textsc{ConceptSIR} enables standard LLMs to solve several challenging ARCANA tasks that were unsolved by direct prompting in our evaluation. Overall, our results suggest that far analogical reasoning remains an open problem for contemporary LLMs. 
All the materials are available at \href{https://sites.google.com/view/arcanacogtest}{sites.google.com/view/arcanacogtest}. The complete codebase, including benchmark, and ConceptSIR, is publicly available at \url{https://github.com/density-11/arcana-test}. All ARCANA data and task instances can be accessed and explored at
\url{https://arcanacogtest.org/}.
\end{abstract}

\keywords{Analogy \and Reasoning \and ARC-AGI \and ARCANA \and LLM \and ConceptSIR}

\section{Introduction}
\label{intro}

Analogical reasoning has long been regarded as a central component of human intelligence and a characteristic of general intelligence more broadly \cite{GENTNER1983155,hofstadter2001analogy,hofstadter1995copycat,holyoak1989analogical}. Humans routinely draw analogies between domains that are superficially dissimilar yet share deep relational structure, allowing them to generalize knowledge flexibly from limited experience. This capability for far analogy—mapping abstract relations across distant domains—plays a critical role in human learning, reasoning, and problem solving.

In recent years, substantial progress has been made in evaluating the general reasoning abilities of artificial intelligence systems, such as Humanity’s Last Exam \cite{phan2025humanity} and SWE-Bench Pro \cite{deng2025swe}. However, these benchmarks are not designed to explicitly probe far analogical reasoning. Among the most challenging existing benchmarks, ARC-AGI-1\&2 evaluate abstraction, reasoning, and analogy through grid-based tasks that require inferring latent rules from a small number of examples \cite{chollet2025arc}. 
In general, ARC-AGI-1\&2 primarily focus on testing near analogy based on relative surface-level feature similarity (e.g., color, shape, position, number, center, copy, and count). While instances of far analogy may arise implicitly, explicit far analogical reasoning—particularly that requiring \textbf{mappings from real-world concepts or physical phenomena to abstract symbolic representations}—is rare and limited in scope. This leaves a substantial gap in the systematic evaluation of fluid intelligence and far analogical reasoning.

\begin{figure*}[ht]
    \centering
    \subcaptionbox{ARC-AGI-1\label{fig:contarc1}}{\includegraphics[width=0.245\linewidth]{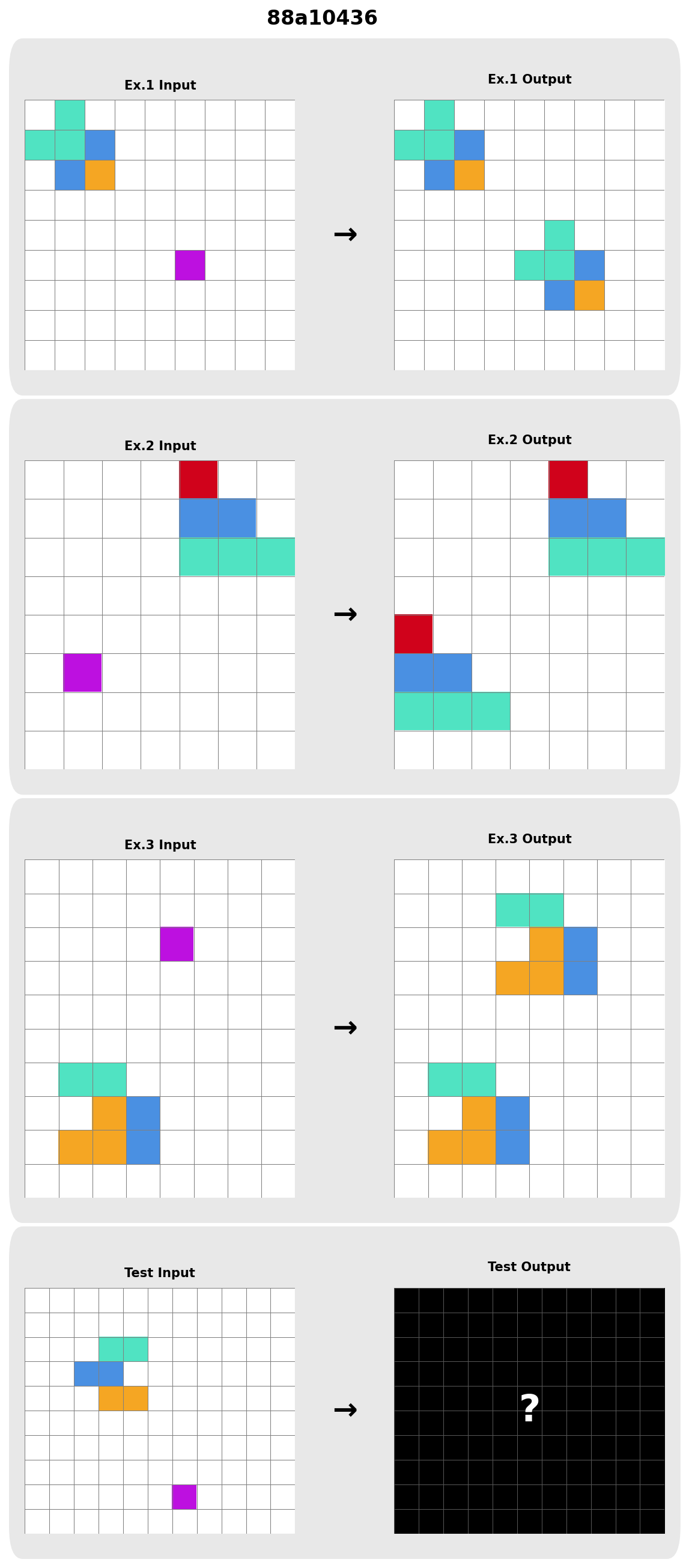}}
    \hfill
    \subcaptionbox{ARC-AGI-2\label{fig:contarc2}}{\includegraphics[width=0.334\linewidth]{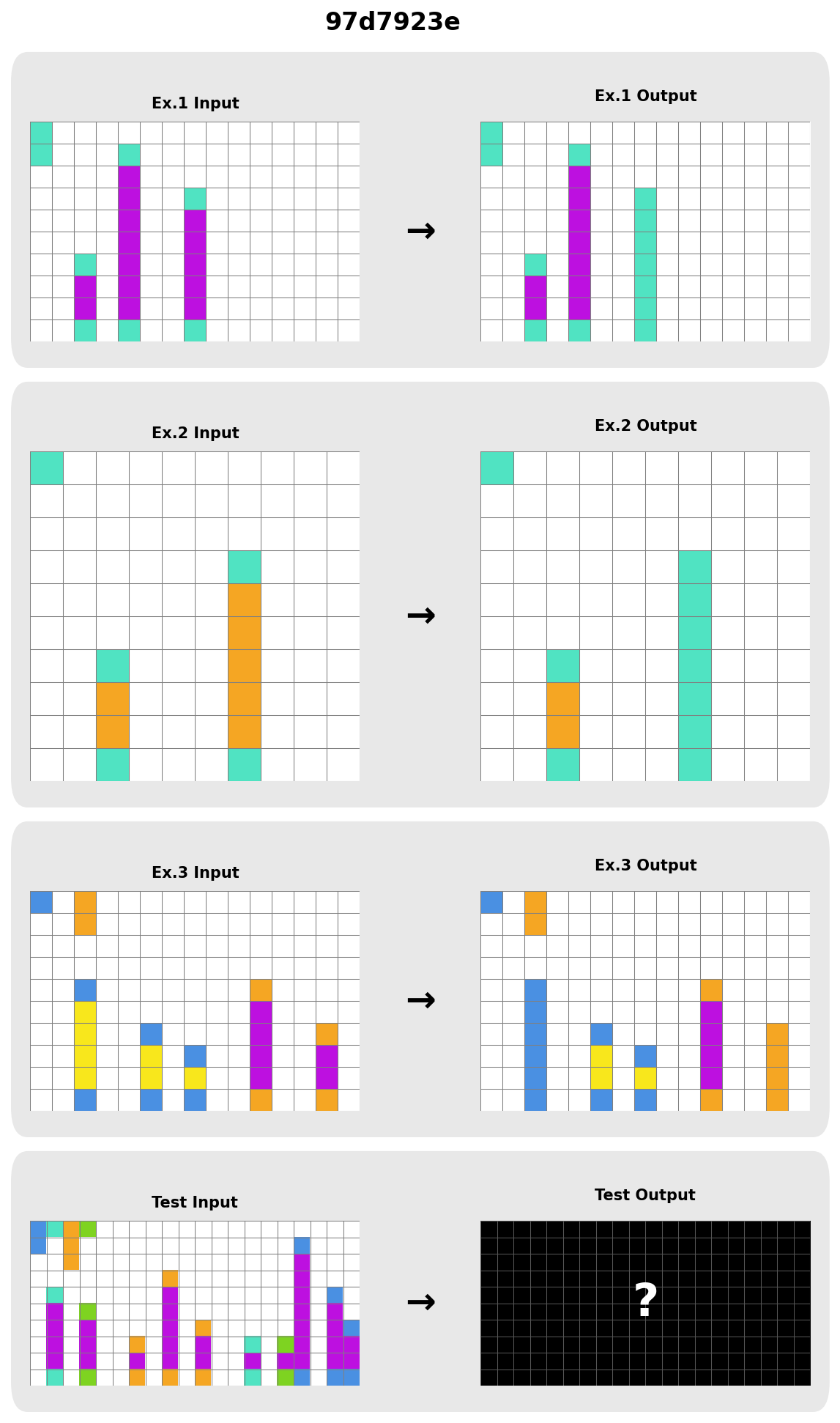}}
    \hfill
    \subcaptionbox{ARCANA\label{fig:contarcana}}{\includegraphics[width=0.26\linewidth]{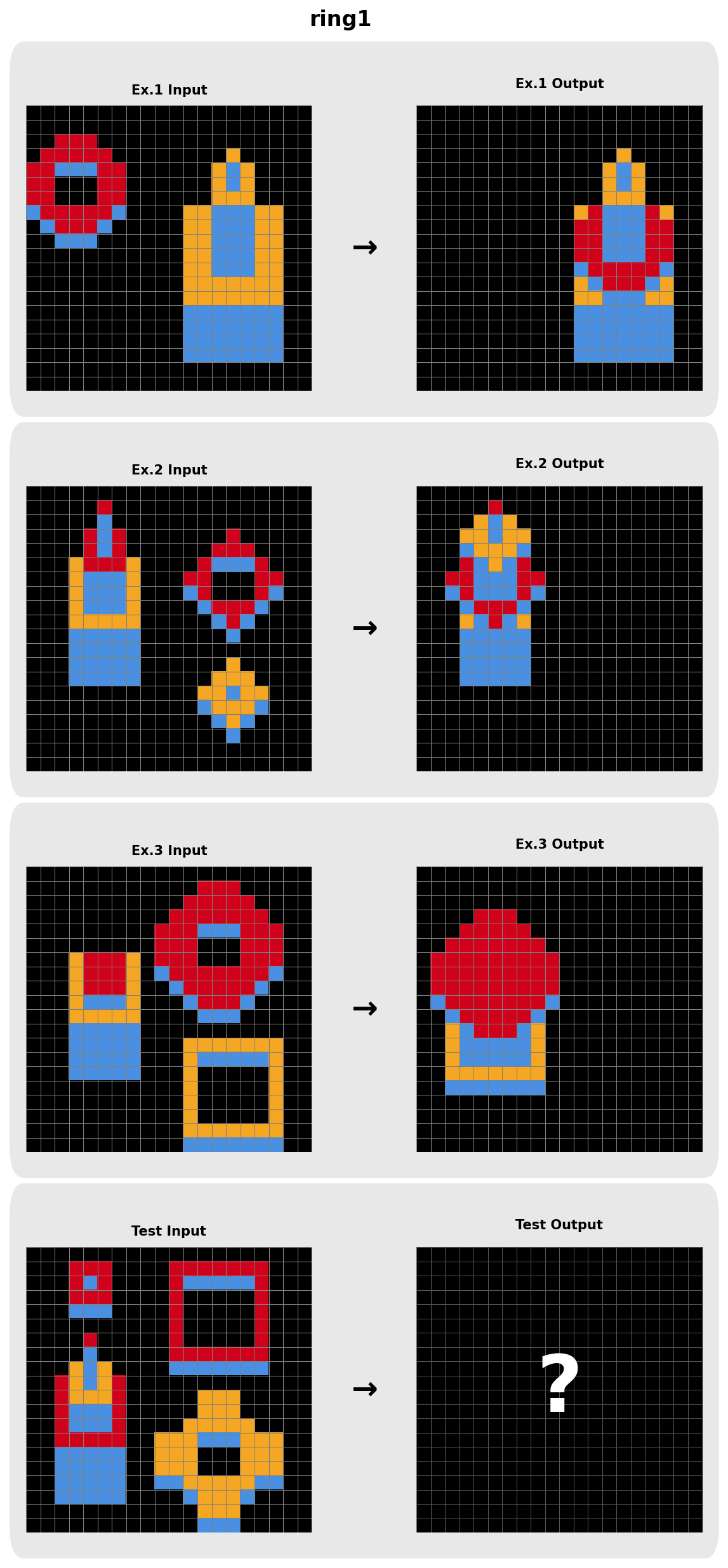}}
    \caption{Visualization of three tasks from ARC-AGI-1, ARC-AGI-2, and ARCANA. (a) Copy the 3\texttimes 3 pattern to the purple block which is the  center of the destination. (b) Fill in the columns with different colors, and the block numbers of the same color in the upper left corner determine the columns to be filled. (c) 
    ARCANA focuses on far analogical reasoning involving threading 3D rings onto a column, where the rings settle at equilibrium positions determined by the geometry of their inner diameters.
    }
    \label{fig:cont}
\end{figure*}

Inspired by ARC-AGI-2 and to address this gap, we introduce ARCANA (Abstraction and Reasoning Corpus for ANAlogy), a benchmark explicitly designed to evaluate far analogical reasoning. 
ARCANA inherits the form of ARC-AGI-1\&2, each task of which consists of 3-5 examples and 1-2 tests.
As shown in Figure \ref{fig:cont}, unlike ARC-AGI-1\&2 which largely emphasizes innate human priors (e.g., copying the 3\texttimes 3 pattern to the purple grid, filling in the grids with different colors using discovered rules), ARCANA focuses on far analogical mappings from real-world objects or physical phenomena to ARC grid transformations (e.g., threading 3D rings onto a column with physical constraints).
The benchmark draws from a broad and diverse set of conceptual sources, including physical principles (e.g., buoyancy, levers, gears, melting, the Doppler effect, Kepler’s second law), natural phenomena (e.g., planets, sunset), and object-level abstractions (e.g., cats, ducks, fireworks, 3D rings, 3D sushi).

To assess the validity and difficulty of ARCANA tasks, we conduct a human study with undergraduate participants. The results show that humans reliably solve ARCANA tasks, indicating that the benchmark captures forms of reasoning that are natural for human cognition. Meanwhile, we evaluate ARCANA across a range of large language models (LLMs), the results of which reveal lower performances, suggesting that far analogical reasoning remains a challenge for current models and exposing systematic failure modes beyond simple pattern matching.

Inspired by the Copycat model of analogical reasoning proposed by \cite{hofstadter1995copycat}, we propose \textsc{ConceptSIR}, an analogical reasoning framework based on concept slippage, introspection, and reflection. \textsc{ConceptSIR} is designed to approximate key aspects of the human problem-solving process when constructing and solving ARCANA tasks. Empirical results show that \textsc{ConceptSIR} enables large language models to solve several ARCANA tasks that are otherwise unsolved by standard prompting strategies, providing insight into the potential of our \textsc{ConceptSIR}.

In summary, this work makes three contributions:
(1) we introduce ARCANA, a benchmark explicitly targeting far analogical reasoning;
(2) we present a human and model evaluation that reveals a substantial performance gap; and
(3) we propose \textsc{ConceptSIR}, a reasoning framework that improves model performance while highlighting remaining challenges.
Together, our results suggest that far analogy remains an open problem for contemporary AI systems and that the reasoning pipeline of \textsc{ConceptSIR} offers a promising direction for studying analogical reasoning.

\section{Related Work}
\label{sec:relatedwork}

\subsection{Cognitive Theories of Analogical Reasoning}
\label{subsec:cogtheory}
Although the computational nature of analogy is not yet fully understood, a number of cognitive theories have been proposed to formalize how analogical reasoning operates. Structure-Mapping Theory (SMT) characterizes analogy as a mapping from a source to a target that prioritizes the preservation of relational structure over surface features \cite{GENTNER1983155}. SAGE, a later extension of SMT, incorporates frequency information over alignable structures, enabling probabilistic generalization across analogical instances \cite{forbus2017extending}. The Analogical Constraint Mapping Engine (ACME) is a classic computational model that formulates analogical mapping as a process of constraint satisfaction \cite{holyoak1989analogical}.

Among these theories, our \textsc{ConceptSIR} draws particular inspiration from Copycat \cite{hofstadter1995copycat, hofstadter2001analogy}, which involves a Slipnet, a workspace and a population of perceptual agents. Although originally developed for letter-string analogy problems, Copycat has been influential as a general cognitive theory of analogical reasoning. In fact, we believe that ARC-AGI tasks are similar to 2-dim Copycat analogy tasks to some extent: Copycat letter-string analogy tasks primarily focus on letters, numbers and symbols, while ARC-AGI tasks focus on numbers and 2-dim concepts.

\begin{figure*}[t]
    \centering
    \includegraphics[width=0.85\linewidth]{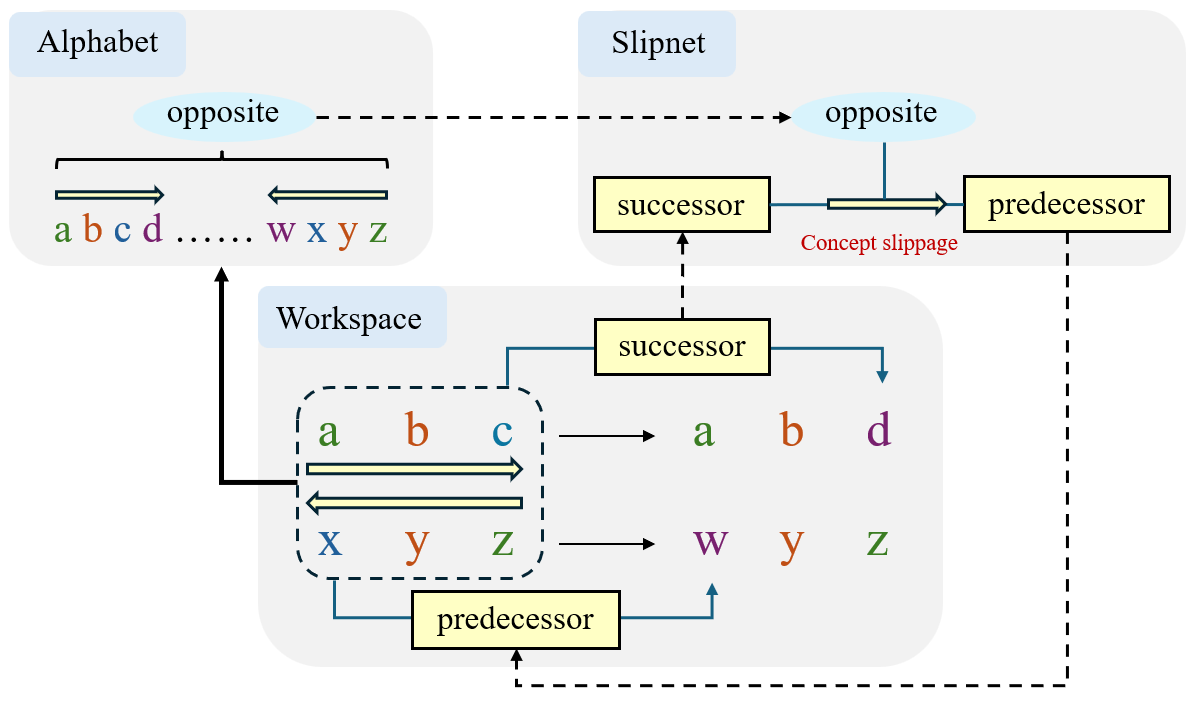}
    \caption{One task from Copycat theory. In workspace, the task is given as: if $a\ b\ c$ changing to $a\ b\ d$, then what will $x\ y\ z$ change to? The concept ``opposite” activated from $a\ b\ c$ and $x\ y\ z$ in the alphabet will promote the concept ``successor” to \textbf{slip to} the concept ``predecessor”. This will finally result in the answer $w\ y\ z$.}
    \label{fig:Conceptslippage}
\end{figure*}

Concept Slippage, a core mechanism in Copycat \cite{hofstadter2001analogy,hofstadter1995copycat}, describes how activating one concept can trigger related concepts and higher-level relational structures, thereby expanding the search over possible interpretations.
One task from Copycat theory is shown in Figure \ref{fig:Conceptslippage}. Given an example: $a\ b\ c \to a\ b\ d $, what will $x\ y\ z$ change to? If we assume the alphabet is not cycle, i.e. the next letter of $z$ is not $a$ but nothing, $x\ y\ z$ might change to $x\ y$, which is obviously not a preferred answer. If we notice that $a\ b\ c$ and $x\ y\ z$ are the beginning and last three letters of the alphabet, the concept ``opposite” will be \textbf{activated} in the Slipnet. Therefore, the concept ``successor” might \textbf{slip to} the concept ``predecessor”, resulting in the answer $w\ y\ z$. The process of one activated concept causing another related concept is \textbf{concept slippage}. 

\subsection{Analogical Reasoning Benchmarks and Models}
\label{subsec:anabench}
A variety of cognitive tests and benchmarks have been developed to assess analogical reasoning ability, among which Raven’s Progressive Matrices (RPM) are a canonical example and are widely used to measure human fluid intelligence \cite{john2003raven, zhang2019raven}. Other benchmarks, such as the Abstraction and Reasoning Corpus (ARC) \cite{chollet2019measure, chollet2025arc}, place a stronger emphasis on abstraction and analogical reasoning, posing persistent challenges for contemporary artificial intelligence systems. Compared to ARC-AGI-1, ARC-AGI-2 \cite{chollet2025arc} introduces substantially more complex and abstract conceptual compositions, significantly increasing task difficulty. Developed from these two benchmarks, ARC-AGI-3\cite{foundation2026arc} is introduced as an interactive benchmark for studying agentic intelligence from novel environments and in which agents must explore, infer goals, build internal models of environment dynamics, and plan effective action sequences without explicit instructions. Nevertheless, systematic evaluation of far analogical reasoning grounded in real-world concepts remains limited, which motivates us to propose ARCANA.

Among models for analogical reasoning, large language models (LLMs) have emerged as the dominant approach for solving ARC-AGI-1\&2, owing to their extensive pretraining and strong general-purpose reasoning capabilities. Beyond LLM-based approaches, alternative methods include train-from-scratch models such as Vision Transformer (ViT)-based picture-to-picture mappings \cite{hu2025arc}, as well as approaches inspired by the Minimal Description Length (MDL) principle \cite{liao2025arc}.
In parallel, several methods have been proposed for analogical reasoning beyond the ARC setting, including vector-based matching approaches that leverage BART embeddings \cite{lu2019emergence}, as well as models that integrate encoder–decoder architectures with Structure-Mapping Theory to induce structured correspondences \cite{crouse2021neural}.

\section{ARCANA: the Abstraction and Reasoning Corpus for ANAlogy}
\label{sec:arcana}

\begin{figure*}[t]
    \centering
    \subcaptionbox{Cake and candle\label{fig:candlearcana}}{\includegraphics[width=0.26\linewidth]{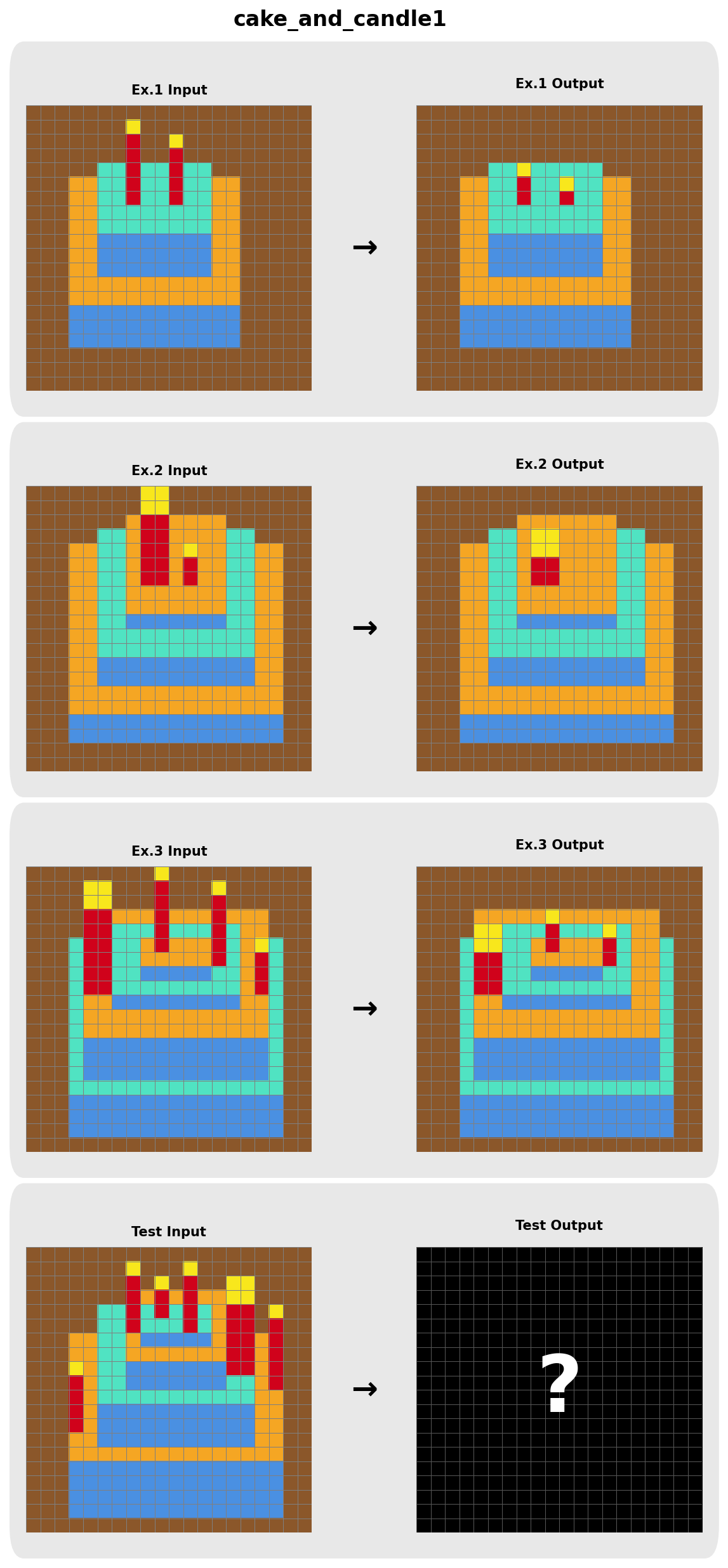}}
    \hfill
    \subcaptionbox{Melt\label{fig:melt1arcana}}{\includegraphics[width=0.26\linewidth]{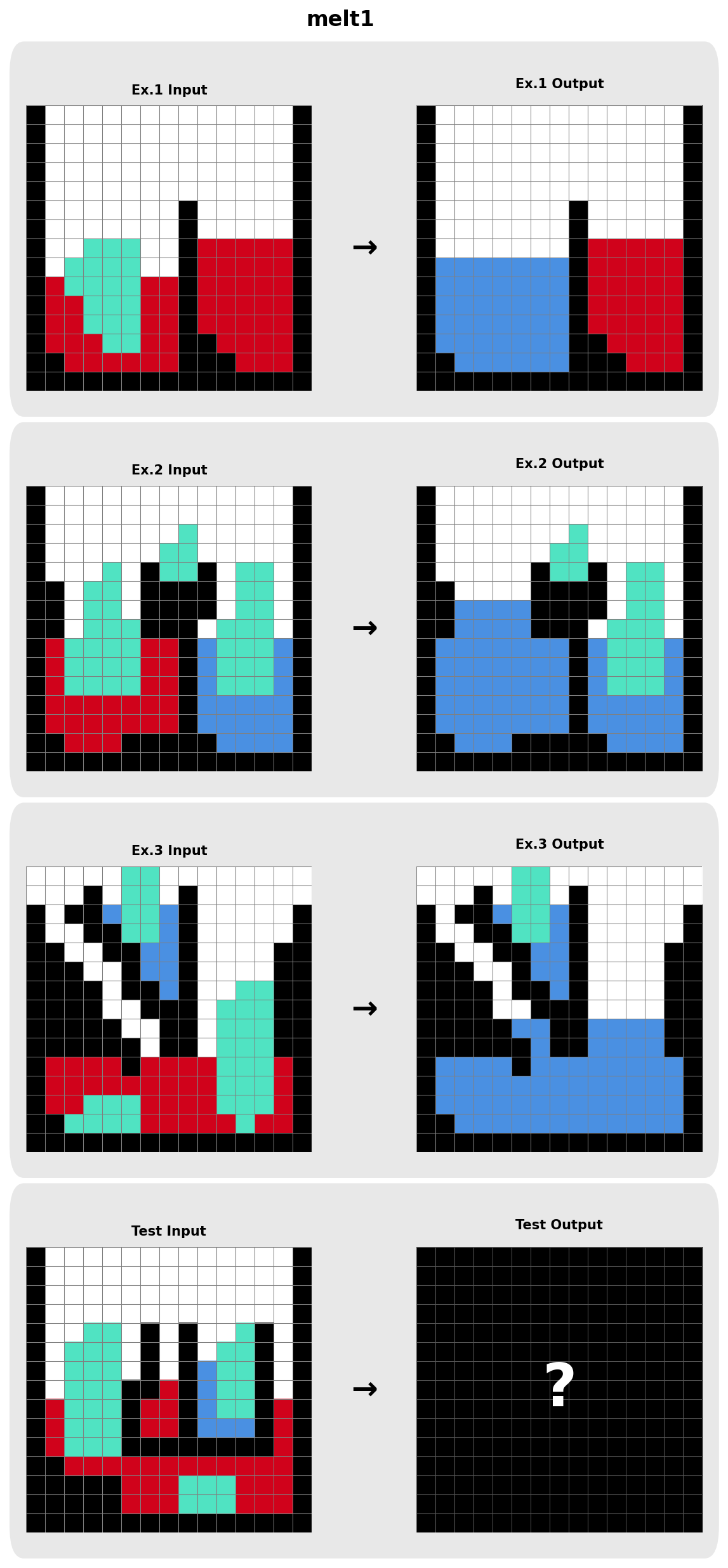}}
    \hfill
    \subcaptionbox{Charge\label{fig:chargearcana}}{\includegraphics[width=0.26\linewidth]{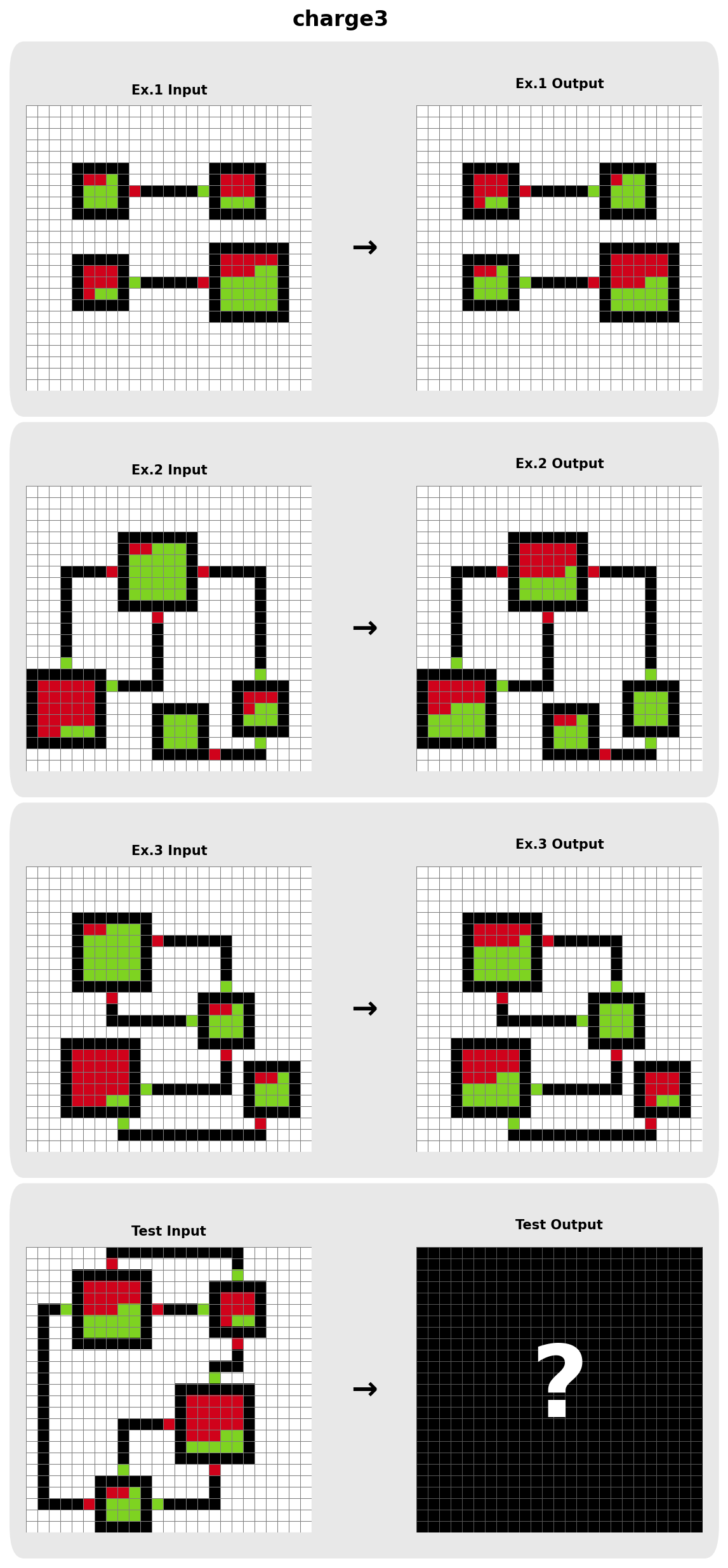}}
    \caption{
    Visualization of three ARCANA task instantiations, illustrating distinct real-world targets and their corresponding abstract concepts. (a) \textbf{Candles on a cake:} candles with different cross-sectional areas and lengths are placed on a 3D cake; after burning by a certain amount, they reveal the cake behind them. (b) \textbf{Melting:} teal cells denote ice, blue cells denote cold water, and red cells denote hot water; ice melts only upon contact with hot water, and the melted ice together with the hot water becomes cold water with volume preserved. (c) \textbf{Charge/data transfer:} green cells (charge/data) propagate along directed wires from the source container connected to the red terminal to the sink container connected to the green terminal; at each time step, each wire transfers one cell if the destination container has remaining capacity.
    }
    \label{fig:constarcana}
\end{figure*}

\subsection{Dataset Creation}
\label{subsec:datasetcreation}

\subsubsection{Target-concept-target Generation Loop}
\label{subsubsec:tctloop}

ARCANA tasks are grounded in diverse \underline{real-world objects, concepts, and physical phenomena}, which we refer to as \textbf{targets}. Each target can be abstracted into a small set of higher-level \textbf{concepts} that govern the grid transformation. For instance, the ``threaded rings" \textbf{target} (shown in Figure \ref{fig:contarcana}) can be summarized by three \textbf{concepts}—\textbf{move}, \textbf{shelter}, and \textbf{3D}—where 3D rings move and partially occlude (``shelter") the column pattern. Once such concepts are \textbf{extracted} from targets, they can be \textbf{slipped} and \textbf{re-grounded} to \textbf{other targets} (e.g., candles burning on a cake, as shown in Figure \ref{fig:candlearcana}). With this \textbf{target-concept-target generation loop}, we have created several different targets that nevertheless instantiate the similar underlying rules or concepts on ARC grids.

Although some tasks in ARCANA might be mapped from physical phenomena, ARCANA still holds a crucial difference from physical reasoning benchmarks. Many tasks in ARCANA are non-physical, including \texttt{cat1}, \texttt{shoes1}, etc. Even physical inspired tasks are highly abstracted away from real physics: \texttt{melt1}, \texttt{melt2}, \texttt{melt4} encode conservation of colored cells and a horizontal surface rather than density, or strict fluid dynamics; \texttt{buoyancy2} concerns grid relations rather than buoyant-force calculation; and even \texttt{Kepler\_second\_law\_or\_area1} can be solved through a grid invariant without knowing Kepler’s law. Therefore, we address that ARCANA evaluates explicit transfer from real-world concepts to abstract grids, and physical phenomena are only one possible source of analogy; the source concept should not be confused with the capability being evaluated.

The target-concept-target generation loop can also be viewed as an explicit process of analogy. We source targets from diverse domains (e.g., water, windmills, flags, and physical laws) and classify tasks into 34 concept categories plus an ``other" class. Example concepts include wave propagation, 3D cuboid motion and shadows, fluid flow, state transitions, object-induced rotation, light propagation, and game-rule dynamics. Many tasks are compositional, combining multiple concepts (e.g., motion with rotation).

\subsubsection{Grounding Targets into Grids}
\label{subsubsec:groundingt2g}

\paragraph{Concept attention.}
Bridging targets and discrete ARC grids requires \textbf{concept attention}: \underline{emphasizing} task-relevant concepts of a target while \underline{suppressing} concepts that are secondary or irrelevant to the core rule. For example, in the \textit{melt} target (shown in Figure \ref{fig:melt1arcana}), ice melts only upon contact with hot water (red cells); after melting, both the ice and the hot-water become cold water (blue cells). The final cold-water volume (the number of grid cells) equals the sum of the initial ice volume and hot-water volume. We therefore \textbf{focus on the melting condition} and this discrete \textbf{conservation rule}, while \textbf{ignoring physical nuances} such as \textbf{density differences across phases and temperatures}.

\paragraph{Concept expansion.}
Beyond concept attention, we sometimes \underline{augment a target with additional concepts} (e.g., time) to fully specify its dynamics and complicate the tasks. We refer to this as \textbf{concept expansion}. For instance, in the \textit{charge} target (shown in Figure~\ref{fig:chargearcana}), green cells (analogous to charge or data) propagate along directed ``wires'' from a source container at the red terminal to a sink container at the green terminal. At each \textbf{time step}, each wire transfers one cell if the sink container has remaining capacity. Introducing an \textbf{explicit temporal process} (and treating it as effectively continuous) is necessary to correctly solve the test. Overall, concept expansion substantially increases the difficulty and diversity of ARCANA tasks.

\begin{figure}[t]
    \centering
    \includegraphics[width=0.7\linewidth]{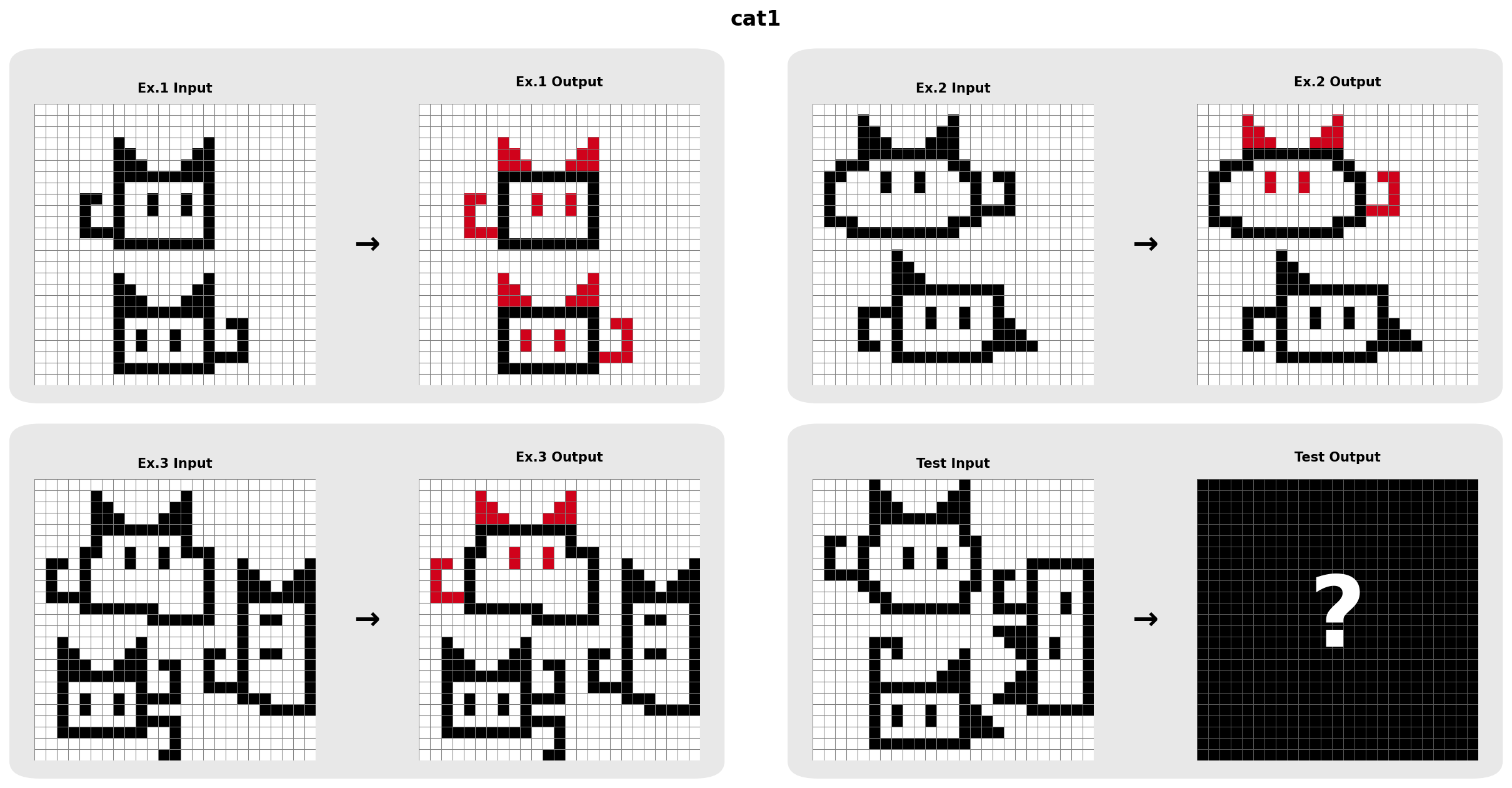}
    \caption{Visualization of \texttt{cat1} task in \texttt{animal} category. The task requires recognizing \textbf{topology} of the true cat, and changing the color of eyes, ears and tail into red.}
    \label{fig:arcanacat}
\end{figure}

\paragraph{Generation principles} 
We construct the benchmark according to four principles that are intentionally simple and human-grounded:
(1) \textbf{Human-familiar targets.} The targets are common and recognizable to humans; given the target, most people who are familiar with it can solve the task accurately with little effort.
(2) \textbf{Concept-faithful grounding.} The ARC grids inherit the target's core concepts, and the grid-target analogy is intended to be explicit.
(3) \textbf{Multiple targets, one rule.} A task may admit multiple plausible analogical targets, but they must agree on the same underlying transformation rule, such as claw and chopsticks.
(4) \textbf{Human-aligned correctness.}  When multiple rules are consistent with the examples, we choose the one most natural to humans, best aligned with everyday concepts, and lowest in cognitive burden.


\subsection{Discussion on ARCANA Tasks}
\label{subsec:arcanadiscuss}

ARCANA comprises \textbf{200 tasks} spanning \textbf{35 distinct concepts}, with each task targeting a unique analogical mapping. We have provided the categorization of ARCANA tasks in Appendix \ref{appsec:arcanadetail}. 

All tasks are named by the abbreviated \textbf{target} from the designer's perspective followed by a numeric suffix, while different tasks are guaranteed to correspond to different rules. For example, in \texttt{water\_transform} category, \texttt{melt1}, \texttt{melt2} and \texttt{melt4} tasks correspond to the following rules respectively: (i) ice melts if it touches hot water (as shown in Fig. \ref{fig:constarcana}), (ii) ice melts only if sunlight shines on it, and (iii) ice melts only if there is no light-blocking glass above it. These variants test subtly different capabilities. \texttt{melt3} task is under \texttt{heat} category, the rule of which is that ice cream melts under the sun. Unlike the other three, its solution does not involve water-conservation-style reasoning; therefore, we do not place it in the \texttt{water\_transform} category as the others. \texttt{melt2, melt3, melt4} are shown in Appendix \ref{appsec:arcanadetail}, Fig ~\ref{fig:arcanamelt2}, Fig ~\ref{fig:arcanamelt3}, Fig ~\ref{fig:arcanamelt4}.

Interesting results have emerged on certain tasks and categories during human study and LLMs testing, which we will discuss later. In this section we would like to clarify the difficulty and characteristics of ARCANA from the tasks themselves.

In the \texttt{animal} category, all tasks require recognition of \textbf{topological structure}, e.g. \texttt{cat1} task in Fig \ref{fig:arcanacat}. Such topological recognition is naturally hard to represent in code, which would be challenging for LLMs while easy for human. There are other similar instances in \texttt{pulley} category and 3d-pattern category. We therefore assumed that providing the visualization of the tasks for LLMs testing rather than merely providing the matrices in text might bring about better results in such categories, and accordingly designed the experiment setup in LLMs testing. The results of LLMs testing also confirmed our assumption.

\begin{figure}[t]
    \centering
    \includegraphics[width=0.7\linewidth]{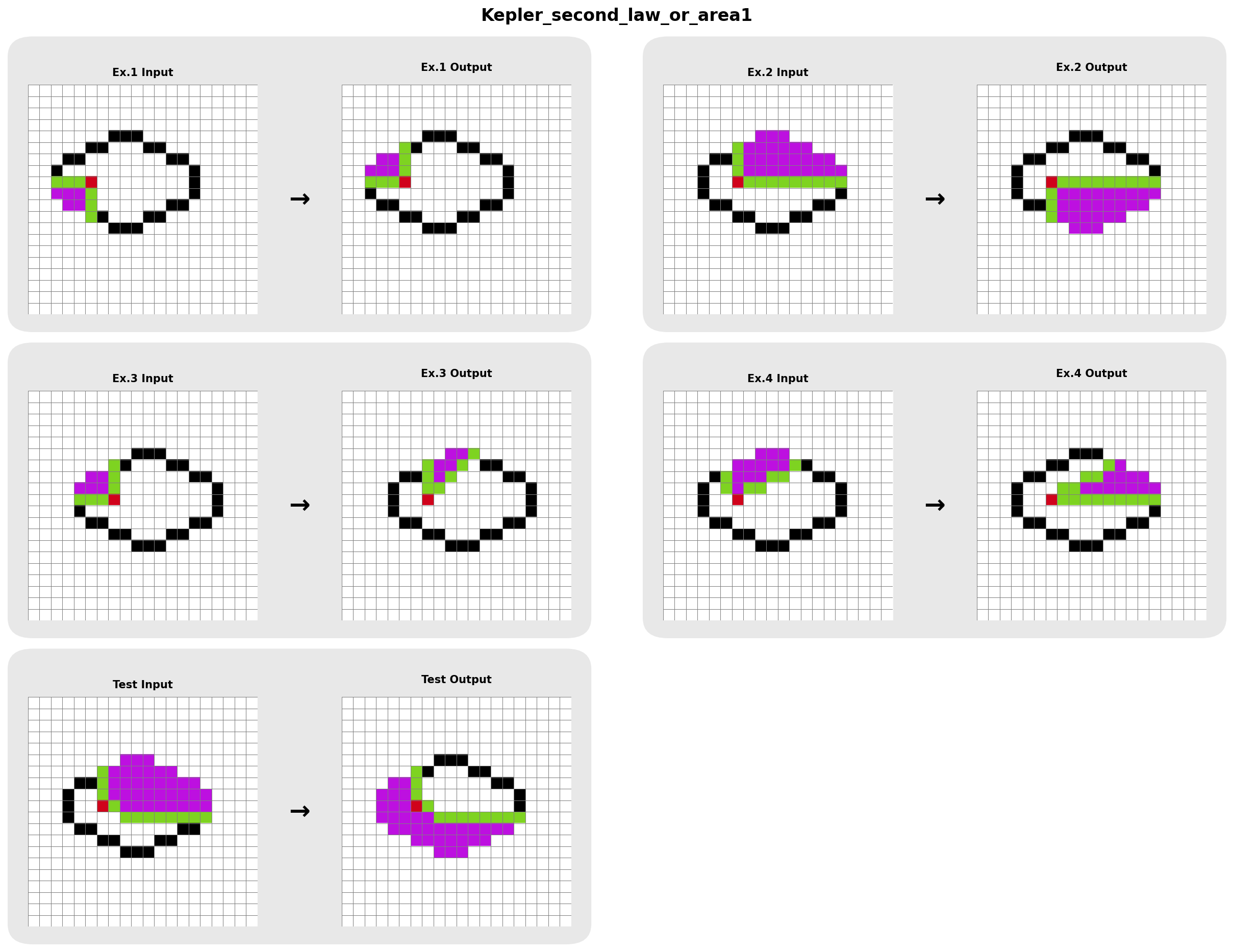}
    \caption{Visualization of \texttt{Kepler\_second\_law\_or\_area1} task in \texttt{other} category. We have provided the answer of the test in the figure. The same contour of each matrix represents the orbit of the planet which should be regarded as an ``ellipse”, and the red grid represents the Sun (star). In either input or output matrix, the green ``lines” represent the line segments joining the planet and the Sun, indicating the beginning and the end positions of one clockwise rotation. The purple grids represent the area swept out by the line segments, remaining consistent from input to output. All examples are designed to possess a unique solution.}
    \label{fig:arcanakepler}
\end{figure}

Another difficulty of ARCANA tasks lies in abstraction and analogical reasoning, which is for both humans and LLMs. Some tasks are highly abstract, inheriting such core concepts in higher levels such as number conservation (\texttt{Kepler\_second\_law\_or\_area1} task, \texttt{lid2} task) and number multiple relationship (\texttt{Doppler1} task, \texttt{cuboid\_and\_shadow3} task). Faithful groundings on ARC grids for such complicated targets appear to be challenging for the representational boundary of the simple input and output matrices and the complicated physical process. Take the grounding of \texttt{Kepler\_second\_law\_or\_area1} task in Fig \ref{fig:arcanakepler} as an example (Kepler's second law: a line segment joining a planet and the Sun sweeps out equal areas during equal intervals of time). In each matrix, the same contour should be regarded as an ``ellipse” orbit of the planet, and the red grid represents the Sun. In either the input or output matrix, the green ``lines” represent the line segments joining the planet and the Sun, indicating the beginning position and the end position of one rotation. The end position of the input is the same as the beginning position of the output. The purple grids represent the area swept out by the line segments. The areas remain consistent from input to output, reflecting the rule of Kepler's second law. One confusion might be raised from the representations of the ``lines”: some of the green lines are irregular in examples 4 and the test, and might result in non-unique answers. When designing this task, we have guaranteed all irregular lines are definitely given in all input matrices, and the lines that should be inferred are all straight (horizontal, vertical, or diagonal). No matter how irregular the lines could be, straight lines are always preferred. Accordingly, all examples and the test are designed to possess corresponding unique answer.

Displaying the \texttt{Kepler\_second\_law\_or\_area1} task here actually requires preparation of being doubted. We believe there will still be many readers considering the task as far-fetched despite the detailed description of our designing approach. Objectively speaking, the ellipses and the oblique lines are hard to represent in matrices due to the representation boundary of the ARC grids, leading to confusion and difficulty in recognizing the target and further successful solutions. In our human study, there was still one and only one out of six participants who successfully solved the task, encouraging us to display the task in this section. From designers' perspective, we do not consider this task as the best and the most delicate task, but an experimental and a little ``far-fetched” one. However, we think the task is the most abstract one in ARCANA and visualize it as an example that indicates the difficulty of ARCANA and the difference between ARCANA and ARC-AGI benchmarks.

The \texttt{Kepler\_second\_law\_or\_area1} task might also be solved if the participants have noticed the conservation rule of the purple area. The process of solving the task could be regarded as the scientific discovering process of discovering Kepler's second law. Moreover, not all tasks in ARCANA are such abstract and difficult like this, which could also be proved by the results of human study in the next section. There are also tasks requiring meticulous reasoning especially for the tasks with the concepts augmented by time (as discussed in Section \ref{subsubsec:groundingt2g}). Through this discussion, we wish we had shed light on the difference between ARCANA and ARC-AGI benchmarks and the importance of far analogy.

\section{Human Study on ARCANA}
\label{sec:humanstudy}

\subsection{Experiment Setup}
\label{subsec:hsexsetup}

Since solving ARCANA tasks often benefits from recognizing the intended real-world targets, unfamiliarity with the targets can increase failure rates. To reduce such idiosyncratic variance and obtain a more reliable estimate, each task is randomly assigned to at least six participants.
We recruit undergraduate participants via an online platform. In addition to final grid answers, we ask participants to (i) provide a brief one-sentence description of the inferred target, preferably in analogical terms, and (ii) rate the task's thinking and drawing difficulty on a five-level scale (very easy to very hard), enabling finer-grained analyses of human reasoning. To ensure data validity, participants must pass a tutorial of the online platform and an attention check; submissions from participants who fail the attention check are excluded. After filtering, we obtain 1659 question-level responses from 80 participants.All details are provided in Appendix \ref{appsubsec:humanexdetails}.

The reported human accuracy may vary across populations, because performance on far-analogy tasks partly depends on familiarity with the targets. To reduce cultural variation, we primarily selected broadly recognizable targets and concepts, although it is not enough to guarantee equal familiarity across populations, even on the participants recruited. Our undergraduate sample was chosen to provide participants with a reasonable level of familiarity with the concepts represented in ARCANA. For example, lower performance among young children or participants unfamiliar with a physical law may reflect limited source-domain knowledge rather than difficulty with analogy itself. Conversely, these participants might perform well on far analogies grounded in domains more familiar to them.

Therefore, the reported accuracy should be interpreted in the context of the sampled undergraduate population and may not directly generalize to others. Nevertheless, the results provide evidence that, when the underlying targets are familiar, humans can successfully perform the type of far analogical mapping tested by ARCANA.

\subsection{Experiment Analysis}
\label{subsec:hsexanalysis}

ARCANA is designed from human-centered priors, therefore human performance provides a natural reference point. Moreover, ARCANA tasks may admit multiple plausible targets that nonetheless instantiate the same underlying rule (Principle~3), therefore automated evaluation could be unreliable. We \textbf{manually} evaluate every response. All submissions are categorized into five types: \textbf{CC} (answer Correct, analogy Correct), \textbf{CW} (answer Correct, analogy Wrong), \textbf{WC} (answer Wrong, analogy Correct), \textbf{WW} (answer Wrong, analogy Wrong), and \textbf{WT} (answer Wrong with Typo). 
\textbf{WT} captures minor typographical mistakes: the participant expresses the intended rule but leaves one or two erroneous cells/blocks that are typically irrelevant to the core transformation. Although typos exhibit several forms which makes systematic categorization challenging, we still provide a rough taxonomy and concrete examples in the Appendix \ref{appsubsec:typotaxonomy}. If the mistakes affect the rule expression, we instead label the response as \textbf{WC}. Typos should not be considered as ``tricks” to ``improve” the results of human study as we were criticized before: on the one hand, we always report strict exact-match accuracy without typo handling; on the other hand, the results of Human ($\geq 2$), which is the main evaluation metrics of ARC-AGI-1 and ARC-AGI-2, are not significantly influenced by typos considered. 

We acknowledge that all results are evaluated manually by one author ensuring the same criteria were applied consistently. Nevertheless, this may possibly introduce subjective bias and precludes inter-rater agreement analysis. To make our evaluation more persuasive, we conservatively classify the ambiguous cases as incorrect and consistently report strict results without typo handling. Moreover, further analysis on Human ($\geq 2$) and average accuracy (in Appendix \ref{appsubsec:llmreslevelacc}) shows typo judgments do not drive the core conclusions.

The general results are shown in Table \ref{tab:humanstudy}. Human (Average, pass@1) represents the proportion of correct \textbf{answers} in all 1659 answers, and Human ($\geq 2$) represents the proportion of \textbf{tasks} in all 200 tasks solved by at least two participants. Because typo responses are qualitatively distinct from genuine failures, we report two accuracy variants: \texttt{+Typo} counts answers with typographical mistakes as correct, whereas \texttt{-Typo} treats them as errors. All participants are asked to submit the answers only once, i.e. pass@1.

\begin{table}[t]
    \centering
    \begin{tabular}{c|c|c}
        \hline
         ~ & +Typo & -Typo \\
         \hline
        Human (Average, pass@1) & 68.96\% & 58.35\% \\
        \hline
        Human ($\geq 2$, pass@1) & 96\% & 93.5\% \\
        \hline
    \end{tabular}
    \caption{The general results of human study. Human (Average, pass@1) represents the proportion of correct answers in all answers, and Human ($\geq 2$) represents the proportion of tasks solved by at least two participants. \texttt{+Typo} counts answers with typographical mistakes as correct, whereas \texttt{-Typo} treats them as errors.}
    \label{tab:humanstudy}
\end{table}

\begin{figure}[htpb]
    \centering
    \includegraphics[width=0.85\linewidth]{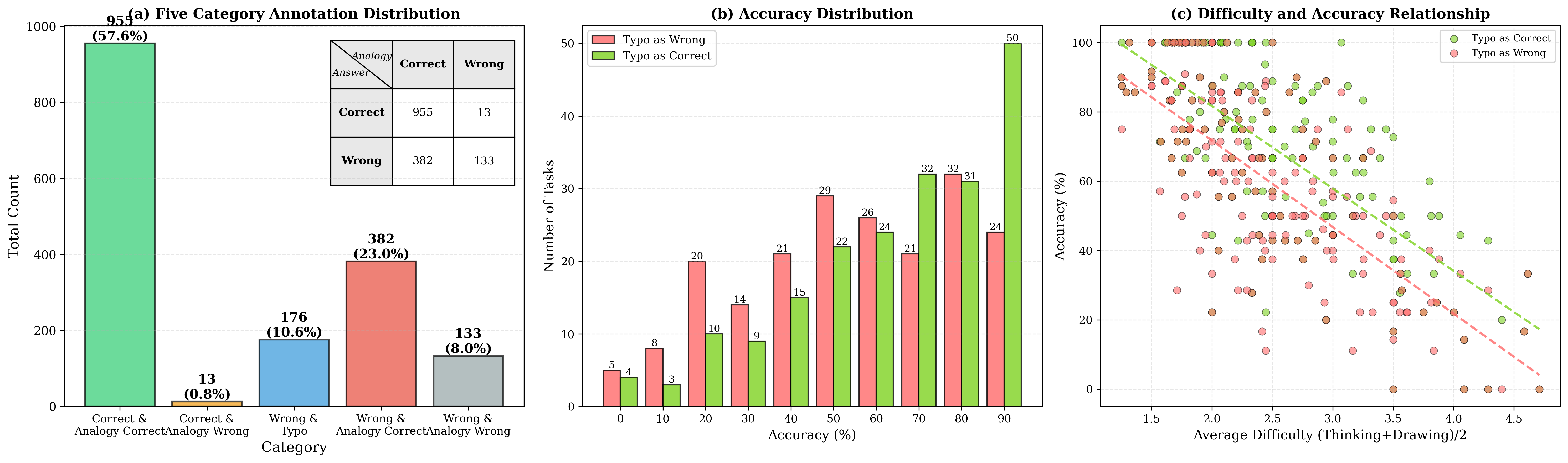}
    \caption{
    Human performance and subjective difficulty on ARCANA. (a) Response distribution over five manual categories (CC, CW, WT, WC, WW), with a $2\times2$ table summarizing answer vs.\ analogy correctness (typo excluded). (b) Per-task human accuracy histogram under two scoring conventions (typo counted as wrong vs.\ correct). (c) Per-task accuracy versus average subjective difficulty (thinking and drawing; mapped to a $1$--$5$ scale), with fitted trend lines for both conventions.
    }
    \label{fig:eachresultdistribution}
\end{figure}

The detailed results of \textbf{Human (Average, pass@1)} are shown in Fig \ref{fig:eachresultdistribution}. The distribution of the manual categories is shown in Figure~\ref{fig:eachresultdistribution}(a). To quantify the relationship between analogy and answer, we compute the Pearson coefficient $\phi$ on the $2\times2$ table after excluding typo responses. We obtain $\phi=0.391$, indicating a moderate positive association between producing a valid analogical interpretation and producing a correct grid output.
The fact that $\phi$ does not exceed $0.6$ (i.e., not a strong association) is also unsurprising: even after a participant identifies a plausible analogy, successfully solving the task can still require additional \textbf{concept expansion} and \textbf{careful, error-free operations} on the grid. Overall, these results are sufficient to conclude that analogy and answers are meaningfully related, and that analogy can serve as a useful aid for solving ARCANA tasks.

We next analyze task-level human accuracy and subjective difficulty to assess whether ARCANA is human-aligned. We also report two accuracy variants: counting typo as wrong versus as correct. Figure~\ref{fig:eachresultdistribution}(b) shows that ARCANA exhibits a reasonable, non-trivial difficulty profile for humans: most tasks fall into intermediate-to-high accuracy ranges, and treating typo as correct shifts the distribution upward. Together, these results support the validity of ARCANA as a human-grounded benchmark. Figure~\ref{fig:eachresultdistribution}(c) further relates accuracy to participants' subjective difficulty ratings, showing that most tasks are perceived as moderately difficult and that perceived difficulty is negatively correlated with accuracy. To quantify uncertainty in the human study, we compute $95\%$ confidence intervals using a hierarchical bootstrap over tasks and human responses. The \textbf{task-level macro-average human accuracy} is $59.13\%$ under strict evaluation ($95\%$ CI: $54.98\%$--$63.31\%$) and $69.67\%$ with typo tolerance ($95\%$ CI: $65.67\%$--$73.65\%$).

Following ARC-AGI-2, we report Human ($\geq 2$) accuracy for ARCANA: $93.5\%$ under strict evaluation and $96\%$ under typo-tolerant evaluation, suggesting that most tasks are well-posed. However, conversely speaking, even with typo included there are still 8 out of 200 tasks successfully solved by less than 2 participants, including \texttt{Kepler\_second\_law\_or\_area1} task, \texttt{wind\_and\_balloon2} task, etc. Although filtering out these tasks is reasonable (also applied in ARC-AGI-2), whether a task meets this threshold can be affected by participant variability due to the special characteristics of far analogy. For example, if all six participants are not familiar with physics or have forgotten Kepler's second law (which is supposed to be on the senior high school textbooks), the solving process of the task might become another process of Kepler's scientific discovering, which requires the ability of analogy on a higher level than solving process of those who are familiar of the physical law. Fortunately, there is still one out of six participants familiar with Kepler's second law and successfully solved the task. Also, accordingly, this could explain the relatively ``low” accuracy of the Human (Average). We therefore retain all tasks to preserve benchmark coverage and diversity.

More details are provided in the Appendix \ref{appsec:humanstudy}. Overall, humans achieve $68.96\%$ average accuracy on ARCANA, including $10.61\%$ typo-level matches, while the Human ($\geq 2$) accuracy reaches $96\%$, including $2.5\%$ typo-level matches. We further observe a moderate positive association between analogy correctness and answer correctness, suggesting that analogy provides a meaningful scaffold for solving ARCANA tasks. Finally, the resulting task-level accuracy and difficulty distributions consistently indicate that ARCANA is broadly solvable for humans and aligned with human reasoning, supporting the validity of the benchmark.

\section{LLMs Testing on ARCANA}
\label{sec:llmtesting}

\subsection{Experiment Setup}
\label{subsec:llmexset}

We have evaluated all ARCANA tasks on four models: GPT~5.2, Gemini~3 Pro, Claude Opus~4.5, and Grok~4.1 Fast. 
Most former LLM methods on ARC-AGI benchmarks employed codes as the solutions, however, according to the discussion of \texttt{animal} category in Section \ref{subsec:arcanadiscuss}, we assume that codes might be insufficient to express rules of all ARCANA tasks. Therefore in our LLMs testing, the chosen models are required to directly output the final answer grids. Under this output format, we could assume that LLMs may also produce minor typographical mistakes. Accordingly, all LLM responses are evaluated manually with the same criteria as in the human study, and results are reported under two conventions: counting typos as correct (\texttt{+Typo}) versus as wrong (\texttt{-Typo}). To further improve performance, we employ two prompting strategies: chain-of-thought and analogical prompting. Due to the substantial stochasticity in model outputs, we follow the common practice in ARC evaluations and report \texttt{pass@3} (top-3). All tests are finished through OpenRouter. Temperature was set to 1. Reasoning effort was not explicitly configured in our OpenRouter requests; therefore, each model used its model-specific default reasoning configuration. In the language-only setting, grids are serialized as numerical matrices with integers representing colors. In the vision-language setting, images are supplied through the standard image interface using Base64 encoding or URLs. Models output the final grid as a numerical matrix in both settings.

Also, the typos of LLMs might be doubted about their validity as those in human study, however there are absolutely no reasons to consider typos as ``tricks” to ``improve” the results of LLMs (after all we are the designers!). The typos might result from the lack of sufficient tokens or the stochastic output. We aim at \textbf{exploring the ability for analogy of LLMs} to our best and prevent these effects, therefore we have also manually evaluated all answers from LLMs just like humans. We provide some typos of LLMs in Appendix \ref{appsubsec:llmtypo}

\begin{table*}[ht]
    \centering
    \begin{tabular}{c|c|c|c|c|c|c}
    \hline
    \multirow{3}*{Model} & \multicolumn{6}{|c}{ARCANA} \\
    \cline{2-7}
    ~ & \multicolumn{2}{|c|}{Total} & \multicolumn{2}{|c|}{CoT (language)} & \multicolumn{2}{|c}{CoT + Ana (VL)} \\
    \cline{2-7}
    ~ & +Typo & -Typo & +Typo & -Typo & +Typo & -Typo  \\
    \hline
    GPT 5.2 & \textbf{51.20\%} & \textbf{48.33\%} & \textbf{40.19\%} & \textbf{38.28\%} & \textbf{42.11\%} & \textbf{37.32\%} \\ 
    \hline
    Gemini 3 Pro & 43.06\% & 29.19\% & 34.45\% & 23.92\% & 28.71\% &  18.66\% \\ 
    \hline
    Grok 4.1 Fast & 24.88\% & 18.66\% & 19.62\% & 14.35\% & 15.79\% & 11.00\% \\ 
    \hline
    Claude Opus 4.5 & 8.61\% & 7.18\% & 7.66\% & 6.22\% & 6.70\% & 5.74\%\\ 
    \hline
    Human (Average, pass@1) & 68.96\% & 58.35\% & - & - & - & -\\
    \hline
    Human ($\geq 2$, pass@1) & \uline{96\%} & \uline{93.5\%} & - & - & - & -\\
    \hline
    \end{tabular}
    \caption{
Experimental results on ARCANA. \texttt{CoT (language)} uses language-only chain-of-thought prompting; \texttt{CoT + Ana (VL)} adds an analogical prompt with visual inputs. \texttt{+Typo}/\texttt{-Typo} indicate whether minor typographical errors are counted as correct. \texttt{Total} is the union of tasks solved under the two setups (LLMs: 12 attempts for \textbf{pass@3} in total across prompts, two settings, each with two prompt variants, each run three times; humans: 1 attempt for \textbf{pass@1}).
    }
    \label{tab:baseline}
\end{table*}

\subsection{Experiment Analysis}
\label{subsec:llmexanalysis}

The experimental results are shown in Table~\ref{tab:baseline}. Overall, LLM performance on ARCANA is below human performance in Table~\ref{tab:baseline}: the \texttt{pass@3} upper bounds of LLMs remain lower than the \texttt{pass@1} human (average) and human ($\geq 2$). Nevertheless, the strongest models still solve a non-trivial subset of tasks, suggesting that current LLMs possess meaningful (albeit limited) capability for ARCANA-style problem solving. 
We emphasize that these results are influenced by evaluation budget (cost per task) and by the inherent stochasticity of LLM outputs.
More details and category-level analyses are provided in the Appendix \ref{appsec:LLMdetail}. 

\textbf{Language vs.\ vision--language prompting.}
Generally, vision-language (VL) prompting yields lower performance than language-only prompting for most models (with one notable exception: GPT~5.2 improves under VL when typos are counted as correct). This trend is consistent with prior findings that adding visual inputs can reduce LLM performance \cite{palmarini2024abstract}, potentially due to misalignment between the vision and language modalities. Importantly, however, these aggregate comparisons do not imply that VL is strictly worse. The \texttt{Total} column is higher than either prompting alone, indicating that the two prompting modalities succeed on different subsets of tasks. Category-level inspection also suggests that some concepts benefit substantially from visual grounding (e.g., Gemini on \texttt{electric\_clock} category and \texttt{moon\_and\_tide1} task, GPT on \texttt{pendulum} category). Together, these observations support our viewpoint that \textbf{ARCANA is inherently visual in nature, yet not merely visual problems}.

\textbf{Correct analogies support LLMs solving ARCANA.}
Under analogical prompting, we collect and manually evaluate model-generated targets, and find that correct analogies tend to co-occur with correct answers, suggesting that analogy can materially support ARCANA task solving. The analogy accuracy could be find in Appendix \ref{appsubsec:llmdiffonvision}. We also observe substantial differences in analogical ability across models resulting in different result accuracy. Focusing on the two strongest models (GPT~5.2 and Gemini~3 Pro), the \texttt{pulley} category is particularly diagnostic: Gemini achieves 100\% analogy correctness while GPT achieves 0\%; correspondingly, Gemini reaches 67\% answer accuracy in \texttt{pulley} whereas GPT remains at 0\%. Overall, these results suggest that analogical reasoning is a meaningful bottleneck for ARCANA performance.

\textbf{Human-model gaps and category-wise performance.}
A paired task-level bootstrap over the $200$ ARCANA tasks between LLM baselines and humans shows that the $95\%$ bootstrap confidence intervals of LLM-Human differences are strictly below zero under both strict and typo-tolerant evaluation (as shown in Table \ref{tab:human_llm_bootstrap}). Even the strongest baseline remains $10.13$ percentage points below humans under strict evaluation and $17.67$ percentage points below under typo-tolerant evaluation. This indicates a statistically reliable gap between current LLMs and human performance on ARCANA.
We further examine performance by task category. Overall, human accuracy is relatively stable across categories, whereas LLM performance is more uneven. Humans outperform LLMs in most categories, but there are also categories in which some models match or exceed the human average (typically those dominated by precise, fine-grained manipulations, e.g., \texttt{shelter\_then\_move} category). At the same time, several categories expose clear gaps where LLMs lag far behind humans (e.g., \texttt{water\_deep\_concept} category and \texttt{animal} category), indicating that substantial differences remain between current LLMs and human analogical reasoning.

\begin{table}[t]
\centering
\small
\begin{tabular}{lcc}
\toprule
Model & Strict $\Delta$ / $95\%$ CI & Typo-tolerant $\Delta$ / $95\%$ CI \\
\midrule
GPT~5.2 
& $-10.13\%$ [$-17.06\%$, $-3.31\%$] 
& $-17.67\%$ [$-24.20\%$, $-11.12\%$] \\
Gemini~3 Pro 
& $-29.63\%$ [$-35.66\%$, $-23.58\%$] 
& $-26.17\%$ [$-32.80\%$, $-19.47\%$] \\
Grok~4.1 Fast 
& $-40.13\%$ [$-45.53\%$, $-34.46\%$] 
& $-44.17\%$ [$-50.24\%$, $-38.08\%$] \\
Claude Opus~4.5 
& $-51.63\%$ [$-56.16\%$, $-46.85\%$] 
& $-60.67\%$ [$-65.59\%$, $-55.73\%$] \\
\bottomrule
\end{tabular}
\caption{Paired task-level bootstrap comparison between LLM baselines and human per-task accuracy on ARCANA. $\Delta$ denotes LLM accuracy minus human accuracy; all confidence intervals are below zero.}
\label{tab:human_llm_bootstrap}
\end{table}

More details are provided in the Appendix \ref{appsec:LLMdetail}. Nonetheless, the scale and breadth of modern LLMs suggest substantial potential for approaching human-like analogy. To narrow this performance gap and probe the limits of LLM analogical reasoning, we introduce a new reasoning framework, \textsc{ConceptSIR}, in the next section and study its empirical behavior.

\section{\textsc{ConceptSIR}: A Reasoning Framework for Concept Slippage, Introspection, Reflection}
\label{sec:conceptsir}

\begin{figure*}[t]
    \centering
    \includegraphics[width=0.85\linewidth]{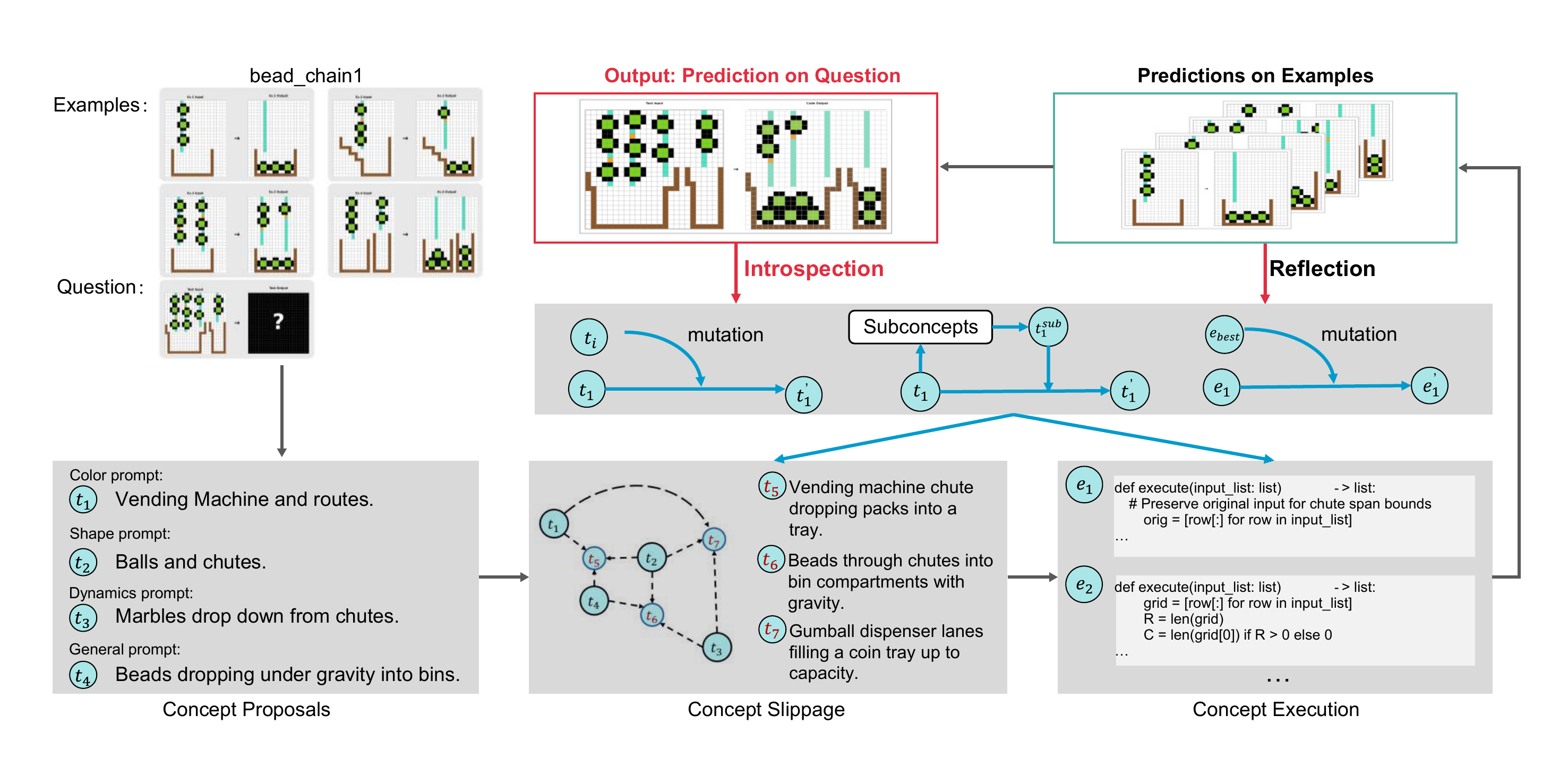}
    \caption{An overview of \textsc{ConceptSIR} framework. Given an ARCANA task, the framework first proposes initial concepts from multi-dimensional prompts. These concepts evolve via Concept Slippage to generate a diverse set of candidates, which are then implemented as executable code under Concept Execution. The system iteratively refines solutions through Reflection, employing mechanisms such as code mutation and concept mutation, while Introspection ensures the final prediction on the test question aligns with the established rules and human preference.}
    \label{fig:ConceptSIR}
\end{figure*}

Inspired by cognitive theories of analogy-making, especially Hofstadter and Mitchell's Copycat architecture \cite{hofstadter2001analogy,hofstadter1995copycat}, we propose \textsc{ConceptSIR}, an LLM-driven analogical reasoning framework built around Concept Slippage, Introspection, and Reflection. Following Copycat, where Workspace structures activate Slipnet concepts that may slip to related concepts and trigger Codelets, \textsc{ConceptSIR} treats the LLM's knowledge as an implicit Slipnet, context-conditioned generation as concept slippage, and concept execution as Codelets. Introspection and reflection then use execution feedback to iteratively refine conceptual hypotheses and executable rules. The framework is shown in Figure~\ref{fig:ConceptSIR}. Additional details are provided in Appendix \ref{appsec:conceptsirdetail}.

\subsection{Concept Proposal and Slippage}
\label{subsec:sirslippage}


Motivated by our human study and the breadth of LLM pretraining, we introduce a concept proposal and slippage stage to elicit diverse analogical targets. We first prompt candidates along four human-prior dimensions---color, shape, dynamics, and holistic recognition---and then use a slippage prompt to expand them into a broader concept neighborhood. We assume that it is particularly useful for LLMs, whose outputs can be overly context-dependent; increasing the diversity of candidate targets helps avoid getting trapped in a ``local minima'' of concepts. From a Bayesian perspective, Copycat-style slippage can be viewed as combining conceptual priors with task evidence: the Slipnet encodes prior knowledge, while structures activated in the Workspace provide likelihood-like constraints on which concepts become relevant. Analogously, LLMs generate new concepts conditioned on both knowledge learned from pretraining and the given task context. Thus, \textsc{ConceptSIR} implements concept slippage as an additional prompting operation that leverages LLMs' context-conditioned generation to activate a more diverse set of related concepts.

\subsection{Concept Execution}
\label{subsec:sirexe}

To reliably apply an inferred rule underlying all given demonstrations to the question grid, we represent the execution step as code rather than natural language. Natural-language rules are often ambiguous or internally inconsistent, making it difficult to ensure faithful and deterministic application to grids. Using code provides an explicit, verifiable realization of a candidate rule and is broadly consistent with prior ARC solvers. Concretely, for each candidate target produced by the slippage stage, we prompt the LLM to synthesize a corresponding program as one possible instantiation of the target's rule, which is then executed and checked against the examples.

A contradiction might lie in the fact that the rules of some ARCANA tasks are difficult to represent in code. In ConceptSIR, we applied a relatively consistent form of expression but not a general one. The form of concept execution could be extended by atomic grid operations, providing a more general concept execution.

\subsection{Introspection}
\label{subsec:sirintrospection}

Introspection iteratively aligns the model's analogical interpretation and its execution with human cognition (Principle~4). Some ARC tasks admit multiple rules consistent with the demonstrations, but humans typically favor one interpretation as the intended solution with the lowest cognitive burden. For example, in a \texttt{clock} task, both ``rotate the pointer on the right dial by $90^\circ$'' and ``rotate the hour hand by $90^\circ$'' may fit all examples, yet the latter is more natural because ``hour hand" composes more directly with the concept of a ``clock". 
After each code update, we run the program on the examples and questions, using introspection to identify abnormal behaviors and producing targeted correction instructions; once all examples are matched, we perform a final introspection pass on the question output to filter invalid or human-misaligned solutions.

\subsection{Reflection}
\label{subsec:sirreflection}

Inspired by genetic algorithms \cite{mitchell1998introduction}, we perform two levels of probabilistic mutations in parallel: \textbf{code mutation} and \textbf{concept mutation}. 

At the \textbf{code} level, mutation primarily focuses on implementation techniques and the handling of special or edge cases. We calculate the discrepancy between the code outputs and the ground truth on example inputs, identifying the code with the smallest error as the optimal candidate. We then employ LLMs to analyze the strengths and weaknesses of this optimal code. For all other code candidates, mutation is triggered probabilistically during modification. Upon code mutation, the LLM is tasked with analyzing the pros and cons of its current code and combining them with the insights from the optimal code to propose improvements, combining the strengths of both while discarding their weaknesses. When humans attempt to solve a problem, they often generate multiple candidate solutions and then iteratively refine them by retaining what works and discarding what does not. This observation naturally motivated our idea of code mutation.

At the \textbf{concept} level, mutation includes broader concept slippage and finer-grained concept augmentation. Concept slippage is aligned with the process in Section \ref{subsec:sirslippage}, with the purpose of generating more diverse results and preventing the model from becoming trapped in cognitive corners. While concept augmentation is to complement a primary concept with its associated subconcepts, thereby facilitating more effective rule implementation (e.g., adding a ``pile-up" concept on top of ``gravity" for \texttt{bead\_chain1} task). Together, these mechanisms promote both diversity and implementability of candidate solutions.

\section{Experimental results of \textsc{ConceptSIR}}
\label{sec:sirexresult}

\subsection{Quantitative Analysis on \textsc{ConceptSIR}}
\label{subsec:sirquan}

Based on stage-wise observations, we use Gemini~3 Pro for concept proposal, concept slippage and introspection, and GPT~5.2 for code-centric generation and refinement. The design is heuristically based on preliminary analogy experiments and our LLM results (we found that Gemini performs best in the concept proposal and slippage stages, while GPT excels in code-centric processes). We tried multiple configurations (e.g., GPT-only vs. GPT+Gemini), while acknowledging substantial room for improvement.

We conduct a focused evaluation of \textsc{ConceptSIR} on a deliberately challenging subset of ARCANA in order to fully explore the capability of each modules. The subset consists of $17$ tasks from $11$ categories, many of which are unsolved or only rarely solved by direct LLM testing in Section~\ref{sec:llmtesting}. These tasks were randomly sampled with a preference for tasks with low direct-LLM accuracy. Under a strict exact-match setting with one run per task and a budget of $5$ loops, \textsc{ConceptSIR} solves $9/17$ tasks across $7$ categories, requiring only $2.22$ loops on average for successful cases. These results suggest that the structured cycle of concept proposal, slippage, execution, and refinement can help LLMs go beyond direct prompting, both enabling solutions on difficult tasks and improving proposal efficiency in several cases. 

\subsection{Qualitative Analysis on \textsc{ConceptSIR}}
\label{subsec:sirqual}

We clarify that \textsc{ConceptSIR} is not a general framework due to the code execution, prediction-error-dependent mutation, and the preliminary pipeline. Nevertheless, the qualitative result analysis shows that the modules in \textsc{ConceptSIR} might be critical in leading LLMs to the correct answer. Furthermore, the results also shed light on the shortages of \textsc{ConceptSIR} or even the LLMs.

\textbf{Concept slippage contributes to program generation.}
We study tasks where failure is caused primarily by selecting an incorrect analogical target, e.g., \texttt{refraction2} task and \texttt{bead\_chain1} task. In \textsc{ConceptSIR}, the concept proposal stage elicits multiple candidate targets, and concept slippage diversifies and refines them, enabling Gemini to recover the intended interpretations (``refraction'' for \texttt{refraction2} task and ``beads falling into a container'' for \texttt{bead\_chain1} task). Conditioned on these targets, GPT solves \texttt{refraction2} task in two reflection steps and produces two correct programs for \texttt{bead\_chain1} task within three steps. In contrast,  \texttt{refraction2} task cannot be solved by all the LLM baselines, while \texttt{bead\_chain1} task is solved only once by GPT~5.2 across 12 attempts.

\textbf{Introspection and reflection contribute to program modification.}
In our case studies, we find that introspection and reflection play a critical role in refining candidate programs into correct and generalizable solutions. For the \texttt{bead\_chain1} task, the target concept of ``beads falling into a container” is under-specified; reflection introduces an implicit discrete packing constraint (e.g., ``bays behave like discrete parking slots”), which captures the pile-up mechanism required to implement the rule. For \texttt{candle1} task, none of the LLM baselines succeeded, while human participants solved $5/6$ instances when typos are counted as correct. Even with the target concept provided, candidate programs can fit all training examples but fail on the test case, e.g., by deleting candle structures or producing inconsistent shadows. Introspection detects these failure modes and provides targeted correction instructions, while reflection revises the program without breaking consistency on the examples. In this case, reflection reveals that the candle should be ``burned down'' by a fixed number of cells, rather than simply shifting an orange segment downward.

\textbf{LLMs perform analogy depending both on images and matrices.} In our LLM baseline tests, Gemini~3 Pro have performed a correct analogy and found out the ratio on \texttt{pulley3} task with VL prompt. However, in \textsc{ConceptSIR} experiments, when we only provided the visualization of the task, Gemini~3 Pro failed to perform a correct analogy, resulting in false code generation. After we added the matrices, Gemini~3 Pro succeeded to generate the target ``pulley”, even though there still exists problems in code generation. The observations above support the assumption that LLMs perform analogy depending on both images and matrices.

\textbf{LLMs tend to be trapped in a ``local minima”.} In a preliminary test with GPT~5, we have tested \texttt{bead\_chain1} task and set the budget of the loops to 15 (which is already solved by GPT~5.2 within 3 steps). In the 15 loops, GPT~5 appeared to re-modify the same locations and fail to notice the concept ``pile up”. The phenomenon is the same as the observations on the CL-bench \cite{dou2026cl}: LLMs demonstrate partial compliance by correctly handling explicit and surface-level requirements, yet systematically fail when tasks demand deeper reasoning. This phenomenon might be eliminated with stronger LLMs (e.g., \textsc{ConceptSIR} with GPT~5.2 has already succeeded to generate the correct answers), however, we still hold the bold view that the ``local minima” or the ``context ignored”  exist in SOTA LLMs, possibly originating from the design of \textsc{ConceptSIR} or even the incomplete world model of transformer itself.


\section{Limitations}
\label{sec:limitations}

ARCANA is currently limited in scale and representation. When we constructed ARCANA, we have tried our best to distinguish each task from each other, i.e., similar tasks could be generated from existing tasks to expand the benchmark. We mainly focus on the diversity but not the quantity of ARCANA, while leaving scaling up an unsettled problem for the community. Moreover, although ARCANA covers diverse targets, the space of real-world concepts and analogical sources is effectively unbounded, making larger benchmarks difficult without community effort. We view ARCANA as a starting point for such expansion. Furthermore, ARC-style grid abstraction provides conciseness, precision, and cognitive challenge, but limits expressivity for structures such as oblique lines, circles, and ellipses (e.g., the ellipses of \texttt{Kepler\_second\_law\_or\_area1} task shown in Figure \ref{fig:arcanakepler} are not strictly faithful expressions), motivating richer representations for future far-analogy benchmarks.

We also note several limitations of \textsc{ConceptSIR}. Future work should evaluate it on a larger set of ARCANA tasks beyond the current challenging subset. Although code execution offers a faithful and reproducible way to apply hypothesized rules, it is not fully expressive for all ARCANA targets, especially topology-heavy ones (e.g., \texttt{animal} category); extending the executor to atomic grid operations with explicit where-to-apply instructions, or expanding the framework to a true agent which could judge output the answer as code or direct matrix may improve generality. Moreover, our code-mutation procedure selects the ``best'' candidate solely by prediction error, which is a rigid proxy for solution quality. Additional modules, such as explicit memory or human feedback, may further improve robustness and generalization. More limitations are also mentioned in Section \ref{sec:sirexresult}.

\section{Conclusion}

We introduce ARCANA, a benchmark for evaluating far analogical reasoning, and analyze both human and LLM performance. We further propose \textsc{ConceptSIR}, an LLM-driven framework based on concept slippage, introspection, and reflection, showing that it can solve tasks missed by direct LLM baselines, while also leaving substantial room for improvement. We hope ARCANA and \textsc{ConceptSIR} help establish far analogy as an important frontier for more human-like, general reasoning.

\bibliographystyle{unsrt}  
\bibliography{references}  

\newpage

\appendix

\section{Motivation of ARCANA}
The grid format is following Chollet’s motivation \cite{chollet2019measure}: grids provide a concise, precise, and cognitively challenging format for evaluating abstraction and reasoning, as we stated in Section \ref{sec:limitations}. More importantly, as Hofstadter and Sander argue in Surfaces and Essences \cite{hofstadter2013surfaces}, abstraction is deeply grounded in analogy. We therefore use the grid as an abstract target domain to which relational structures from real-world concepts can be transferred.

We have consistently acknowledged ARC-AGI as an important inspiration for ARCANA. In particular, ARC-AGI-2 task 28a6618f, which admits a natural water-related interpretation, directly inspired our design; among the public ARC-AGI-1\&2 tasks, we found only one other similarly explicit case, 36a08778.

\begin{figure*}[ht]
    \centering
    \subcaptionbox{28a6681f\label{fig:candlearcana}}{\includegraphics[width=0.45\linewidth]{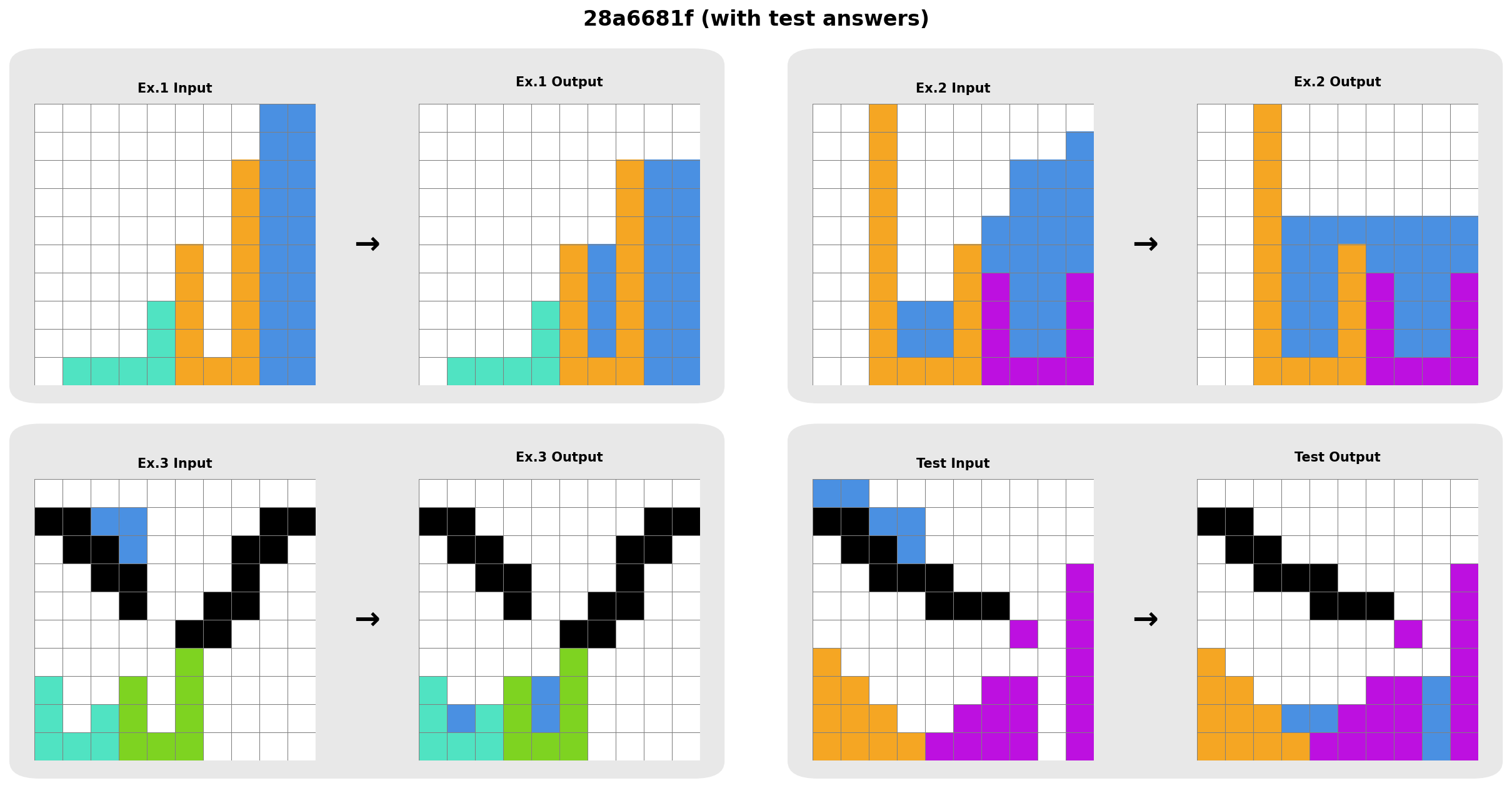}}
    \hfill
    \subcaptionbox{36a08778\label{fig:meltarcana}}{\includegraphics[width=0.45\linewidth]{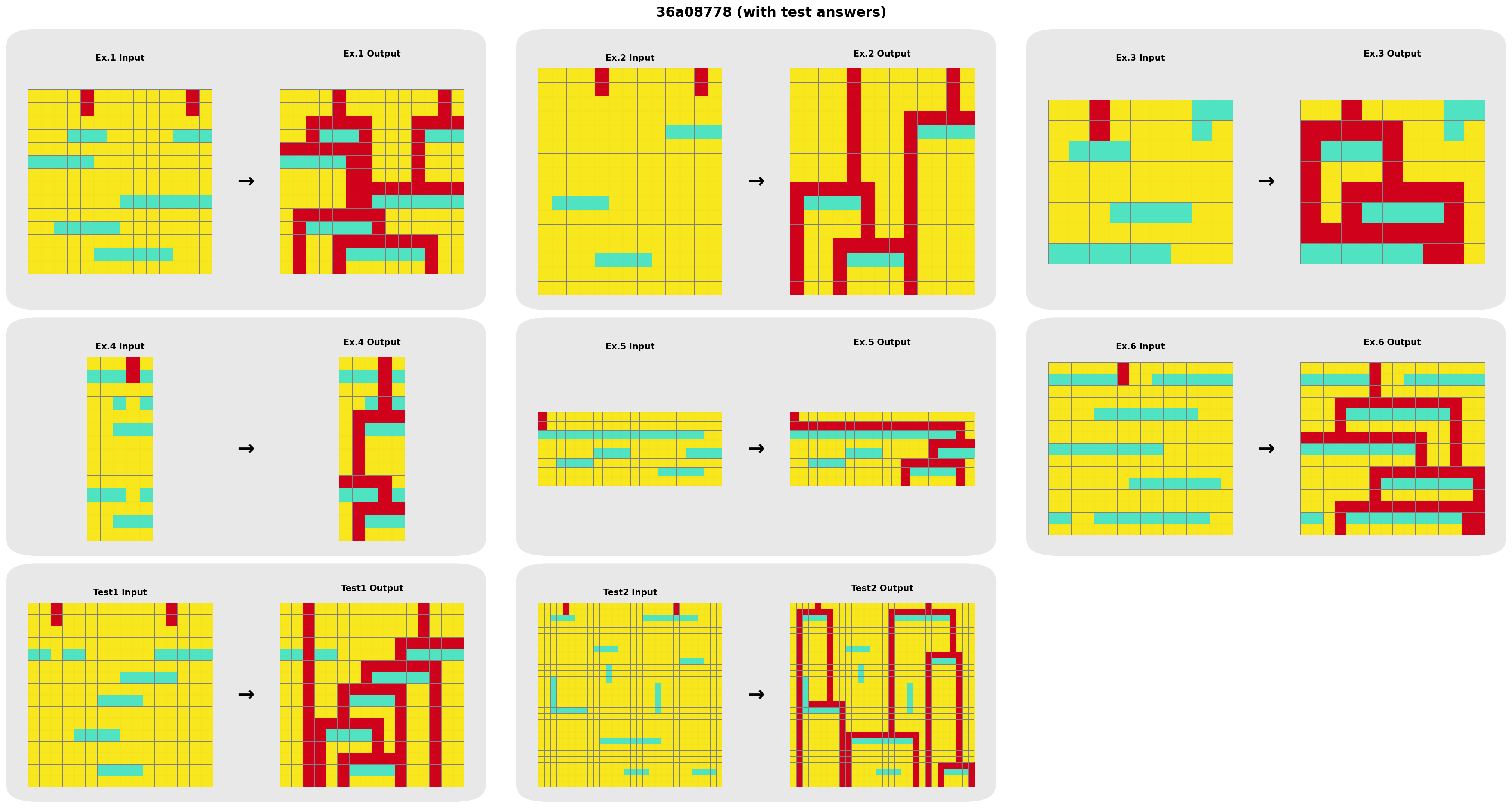}}
    \caption{
Visualization of two ARC-AGI-2 task instances that may admit explicit
far-analogy interpretations. Both tasks can be understood through analogies
to water: (a) water flow and (b) a water stream.
}
    \label{fig:constarcana}
\end{figure*}

\section{Human study details}
\label{appsec:humanstudy}

\subsection{Quantification details.}
\label{appsubsec:humanstatistic}

Excluding typo responses, we compute the Pearson coefficient $\phi$ for a $2\times2$ contingency table of answer correctness versus analogy correctness:
\begin{equation}
\phi = \frac{P_{CC}-P_{CA}P_{AC}}{\sqrt{P_{CA}(1-P_{CA})P_{AC}(1-P_{AC})}},
\end{equation}
where $CA$ denotes all correct answers, $AC$ denotes all analogy-correct answers, and $P$ denotes proportions in the filtered set. In our data, $\phi=0.391$. A chi-square test derived from $\chi^2(1)=n\phi^2$ yields $\chi^2(1)\approx 226.9$ with $p \ll 0.001$. We also report an odds ratio of $OR\approx 25.6$ (95\% CI $\approx [14.3,45.8]$) and Yule's $Q\approx 0.93$.

\subsection{Experimental details.}
\label{appsubsec:humanexdetails}
\begin{figure}[ht]
    \centering
    \includegraphics[width=0.4\linewidth]{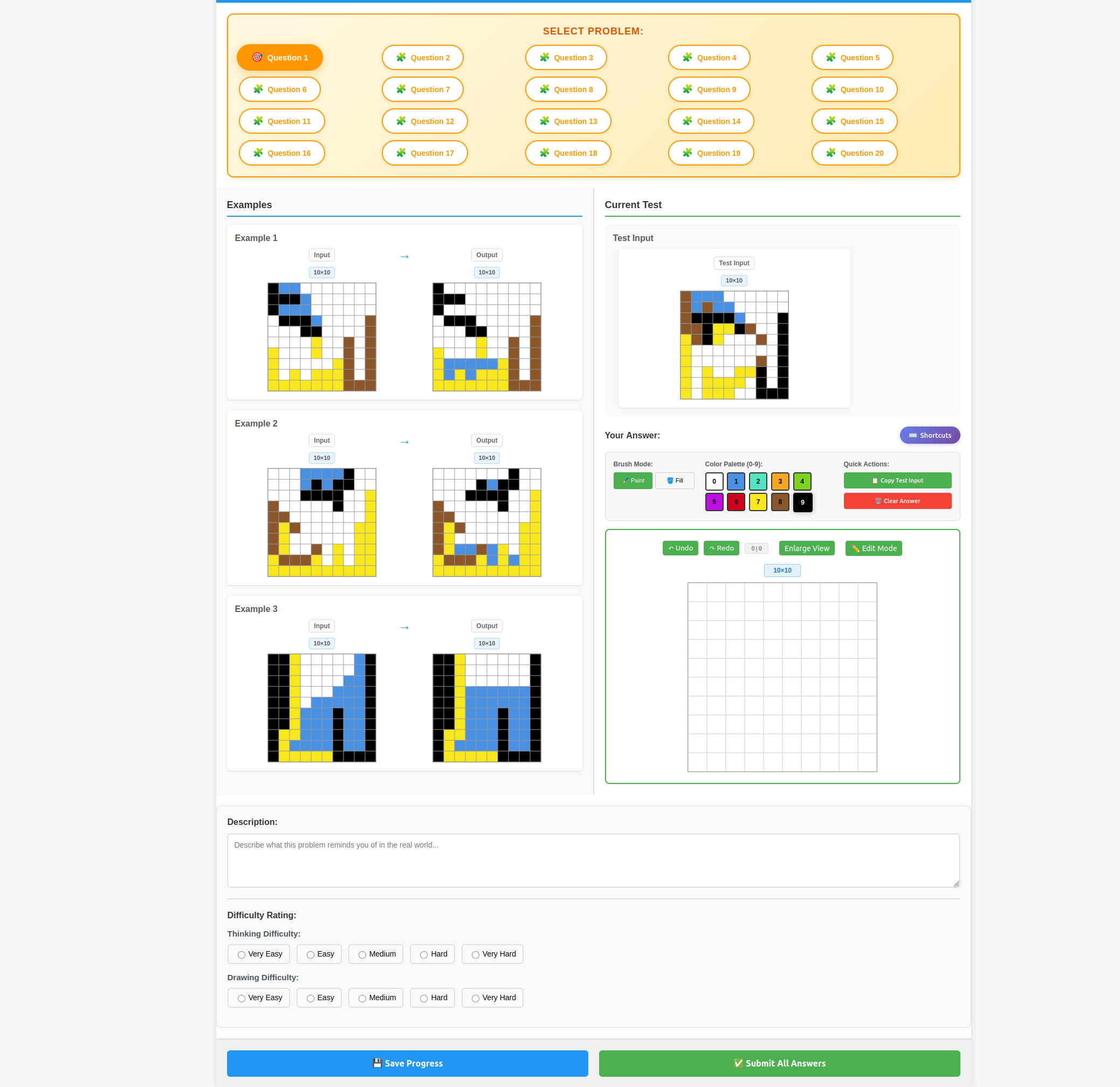}
    \caption{A screenshot of the formal task interface.}
    \label{fig:humanstudy}
\end{figure}

Participants were recruited through an official university platform and completed the study on a custom-built website. A screenshot of the task interface of human study is shown in Figure \ref{fig:humanstudy}. Participants are asked to infer the rule from input–output examples, constructed the test grid, and described what the task reminded them of in the real world. The grid editor and analogy field appeared on the same page, with no enforced response order. Participants could save progress and had 120 hours to complete the study. The website source code has been released in github.

\textbf{Informed Consent for Participants.}

Please read the following information carefully before deciding whether to participate. Your participation is voluntary, and you may stop or withdraw at any time without penalty. Note that eligibility for the study payment is described below.

\paragraph{Purpose}
This study aims to collect human responses on a set of short reasoning tasks. The data will be used to analyze human problem-solving behavior and to support research in cognitive science and related fields.

\paragraph{Procedures}

If you agree to participate, the testing platform will guide you through the following sequence of pages and steps:

\textbf{Introduction page}: An introduction page describes the study and tips for participation.

\textbf{Tutorial page}: A short interactive tutorial that demonstrates how to complete the tasks. You must read and follow the tutorial carefully to learn the response format and controls.

\textbf{Attention check page}: A short verification page to confirm that you understood the instructions in the tutorial. You must complete the attention check correctly to continue.
Formal test page: If you pass the attention check, you will be assigned the formal test sequence of tasks.

\textbf{Important}: You must read the tutorial and complete the attention check correctly to proceed to the formal test. Participants who do not complete the tutorial or who fail the attention check will not proceed to the formal test and will not be eligible for compensation.

\paragraph{Data Collected}
Task responses (the answers you provide for each task).

Timestamps for task start and completion.

Non-identifying metadata such as browser type and assignment ID. These identifiers are used for assignment management and quality control only and are not direct personal identifiers.

\paragraph{Confidentiality}
All data will be treated as confidential. We will only collect necessary data for data management. These will not be used to direct personal identification in research outputs and will be handled in accordance with the project's data-management policies. Collected research data will be stored on secure drives and used only for approved research purposes. Access to raw data will be restricted to authorized research personnel. Any publications or publicly shared datasets will contain only aggregated or anonymized results.

\paragraph{Compensation}
All participants who successfully complete and submit the assigned test session will receive a fixed base payment for their time. In addition to this base payment, participants will earn a small bonus for each task answered correctly. Your total payment will be calculated as the base payment for the completed submission plus the per-item bonuses for correctly answered tasks.

The exact base amount and the bonus rate per correct answer will be displayed on the testing platform and in any payment instructions before you begin. To be eligible for payment (including any earned bonuses), you must complete the assigned session and submit your responses according to the study instructions; withdrawing before completion or failing to submit may affect eligibility. Payments will be issued by the method described by the study coordinator and may be subject to verification and routine quality-control checks.

\paragraph{Risks and Benefits}
There are minimal risks associated with this study. You may experience mild fatigue from performing tasks. There is no direct benefit to you, but your participation contributes to scientific understanding of reasoning.

\paragraph{Voluntary Participation}
Your participation is entirely voluntary. You may decline to participate or stop at any time but without compensation. Please note that payment eligibility is governed by the completion rules described above.

\begin{figure}[t]
    \centering
    \includegraphics[width=0.4\linewidth]{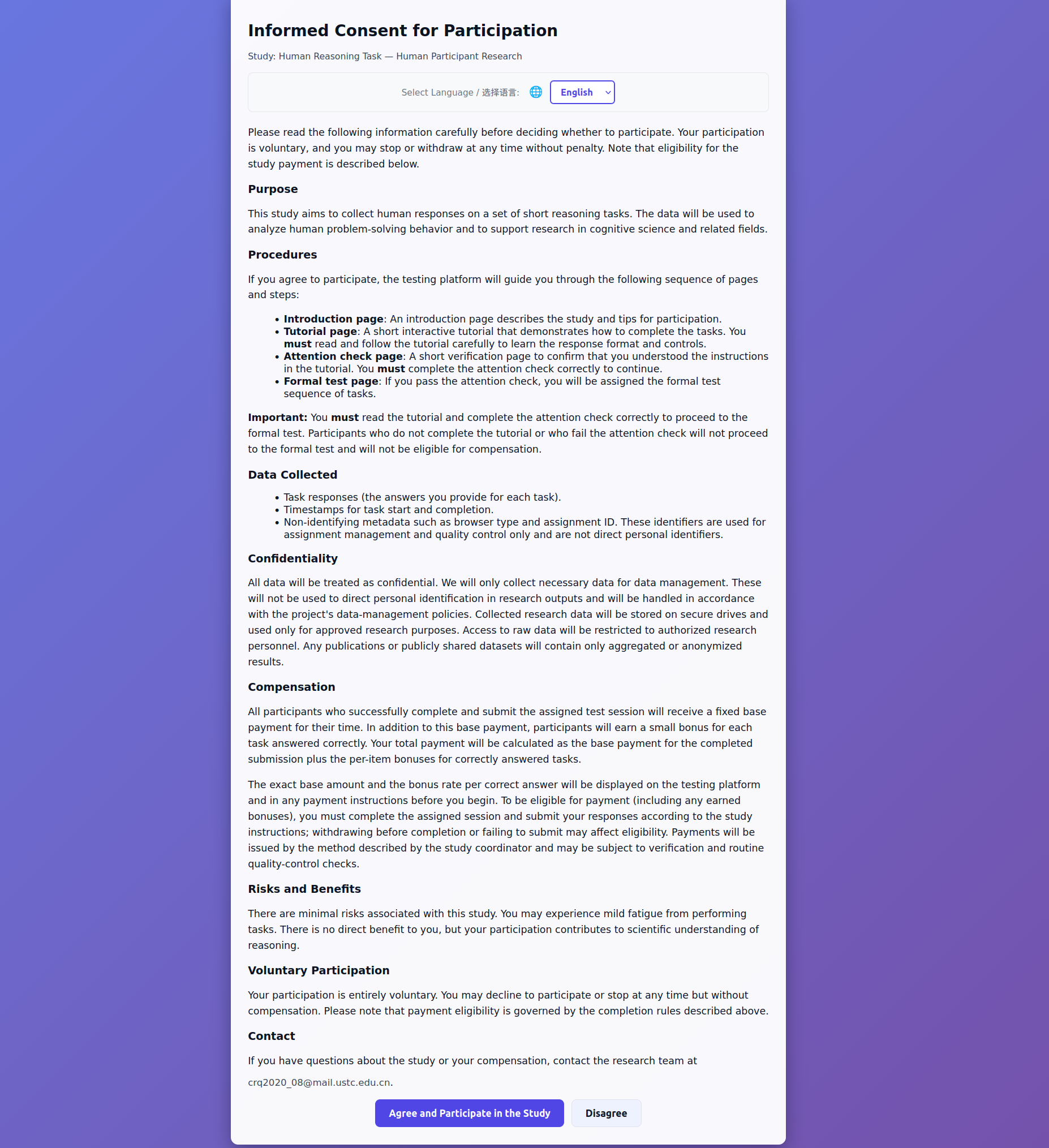}
    \caption{A screenshot of informed consent for participants.}
    \label{fig:informed}
\end{figure}

\textbf{Instructions and Tips}

Instructions:
\begin{enumerate}
    \item You will be presented with visual pattern recognition problems
    \item Each problem contains example input-output pairs to help you understand the pattern
    \item Your task is to identify the underlying rule and apply it to new test cases
    \item Use the color palette and drawing tools to create your answers
    \item After solving all problems, describe what each problem reminds you of in the real world
    \item Once you submit your answers, you cannot return to edit them
    \item You can save your progress at any time and return later, but please finish the formal test within 120 hours.
\end{enumerate}

Tips:
\begin{enumerate}
    \item Study the examples carefully to understand the transformation pattern.
    \item Look for spatial relationships, color changes, and geometric patterns.
    \item Click on any example to enlarge it - easier to see details in small grids.
    \item Take your time - you have 120 hours to complete the test. If you experience mild fatigue, take a short break and return later.
    \item More compensation for more correct answers!
\end{enumerate}

\begin{figure}[t]
    \centering
    \includegraphics[width=0.4\linewidth]{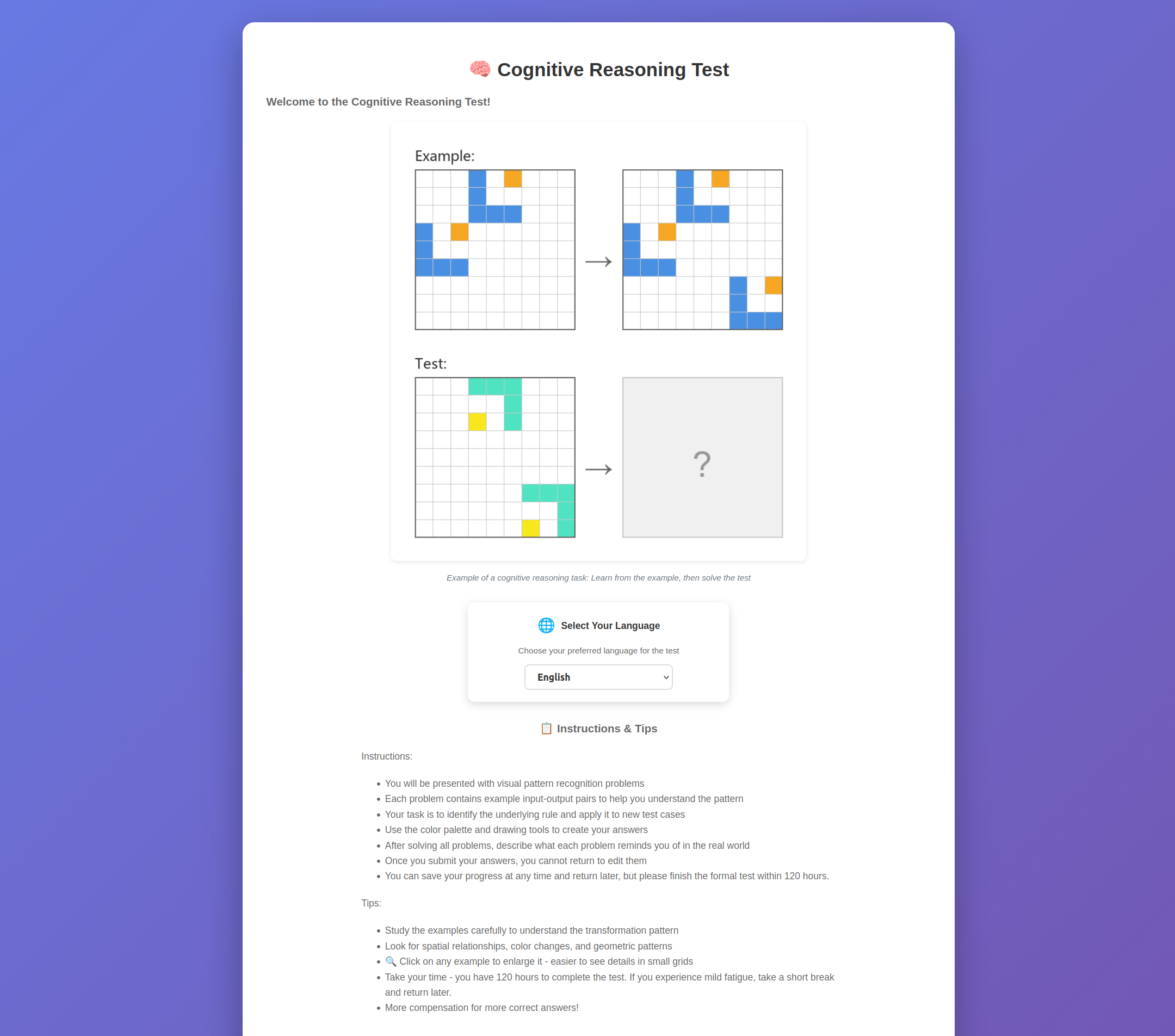}
    \caption{A screenshot of instructions page.}
    \label{fig:welcome}
\end{figure}

\textbf{Tutorials}

\paragraph{Welcome to Tutorial! Step 1 of 6} Welcome to the Cognitive Reasoning Tutorial! This interactive guide will walk you through all interface features.

\textbf{Example Area:} Here you'll see example input-output pairs that help you understand the pattern. Click any example to enlarge it for better visibility, and drag the grid around to explore different parts between input and output of the example.

\paragraph{Problem Selection Step 2 of 6} Use these tabs to switch between different problems. Each problem has its own examples and tests.

\paragraph{Test and Test Navigation Step 3 of 6} This area shows the current test case - the input grid you need to transform. Study it carefully along with the examples.

\textbf{Click to Enlarge:} Click on the test input to open a draggable floating window with an enlarged view. You can move this window around and keep it open wile working on your answer!

Sometimes a problem has multiple tests, use the tabs above to switch between them. Each test requires a separate solution. Watch how the "Current Test" area updates when you switch tests!

\begin{figure}[t]
    \centering
    \includegraphics[width=0.4\linewidth]{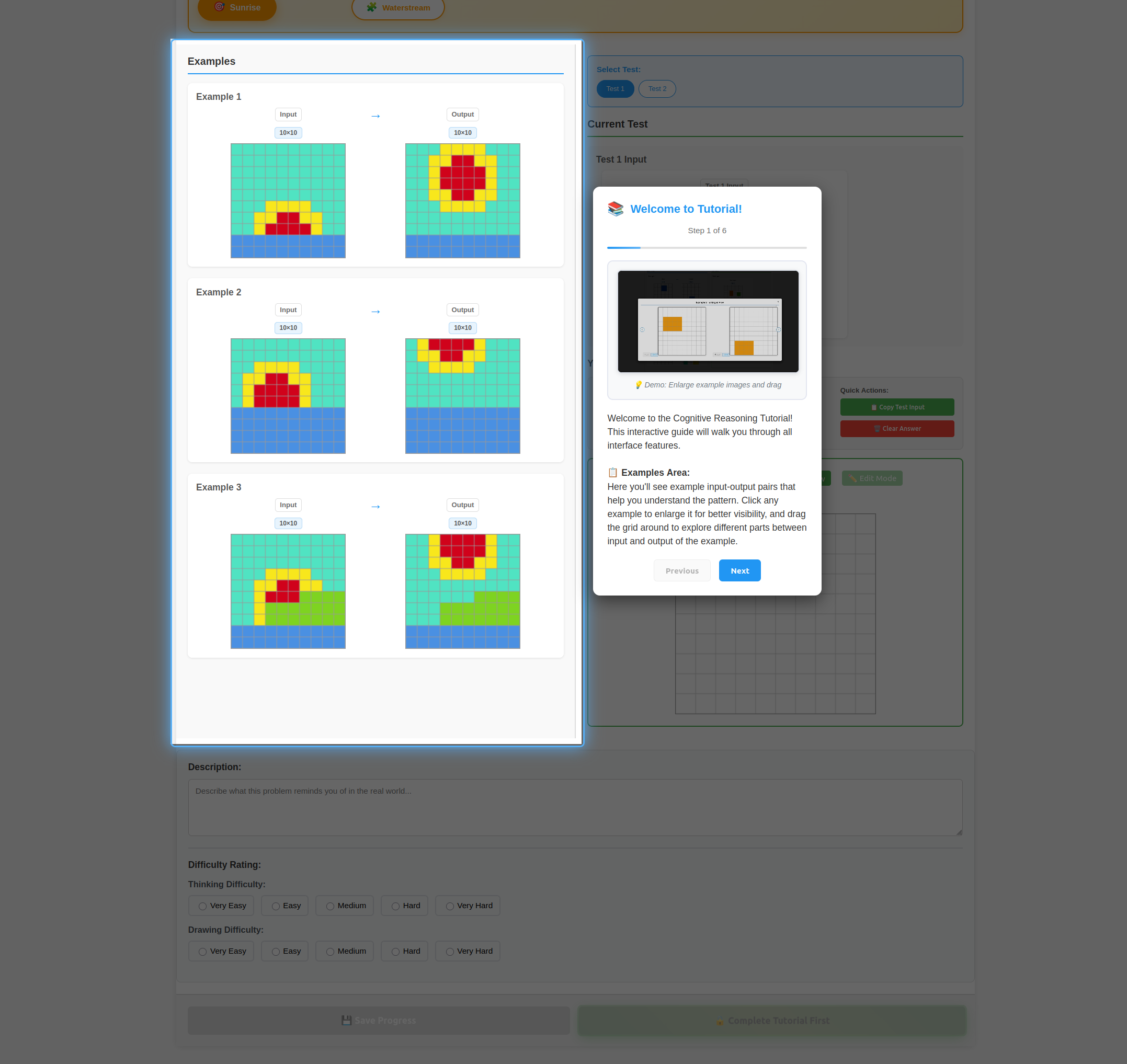}
    \caption{A screenshot of tutorial.}
    \label{fig:tutorial}
\end{figure}

\paragraph{Drawing Tools Step 4 of 6} Use these tools to create your answer:
\begin{enumerate}
    \item Paint/Fill mode buttons
    \item Color palette
    \item Copy and Clear buttons
    \item Enlarge View button
    \item \textbf{Edit Mode button} (unlocked after tutorial)
\end{enumerate}

\textbf{Tip:} To get started quickly, click the \textbf{Copy Test Input} button to copy the test input into the answer grid as a starting point. When you have finished your answer, use the \textbf{Enlarge View} button to open an enlarged side-by-side comparison of the test input and your answer so you can check accuracy.

Here are also some keyboard shortcuts for efficient solving:

\textbf{Color Selection:}
\begin{enumerate}
    \item \texttt{0-9} Select color directly
    \item \texttt{W/A/S/D} Navigation on the color palette
\end{enumerate}

\textbf{Drawing Tools:}
\begin{enumerate}
    \item \texttt{Q} Paint mode (single cell)
    \item \texttt{E} Fill mode (flood fill)
\end{enumerate}

\textbf{Edit History:}
\begin{enumerate}
    \item \texttt{Ctrl+Z} Undo last action
    \item \texttt{Ctrl+Y/Ctrl+Shift+Z} Redo action
    \item \texttt{Ctrl+Shift+X} Copy current problem input
\end{enumerate}

Now enjoy your drawing! \textit{To explore more features, please access Edit Mode after completing the tutorial.}

\paragraph{Description Field and Difficulty Rating Step 5 of 6}
After solving each problem, please describe what the problem reminds you of in the real world. This helps researchers understand how people interpret these patterns and will not affect your accuracy.

Additionally, please rate the \textbf{Thinking Difficulty} and \textbf{Drawing Difficulty} of each problem on a 5-point scale from 'Very Easy' to "Very Hard". These ratings are required and help us understand your perception of the difficulty distribution of the problems.

\paragraph{Tutorial Complete! Step 6 of 6} Congratulations! You've mastered all the interface features:
\begin{enumerate}
    \item \textbf{Examples:} Viewing and enlarging example patterns
    \item \textbf{Navigation:} Switching between problems and tests
    \item \textbf{Drawing Tools:} Paint/Fill modes and color palette
    \item \textbf{Description:} Writing pattern observations
\end{enumerate}

\textbf{Important: Save Your Progress!}
\begin{enumerate}
    \item Remember to \textbf{save} before closing the page (use the Save Progress button or Ctrl+S). When you reopen the page, your saved progress will automatically loaded
    \item There is NO auto-save, you must manually save your work. Closing the page without saving will result in \textbf{losing} your unsaved progress.
\end{enumerate}

\textbf{Ready to begin?} Click "Begin Attention Check" to start with the attention verification before the main test!

\textbf{Edit Mode Tutorial}

\begin{figure}[t]
    \centering
    \includegraphics[width=0.4\linewidth]{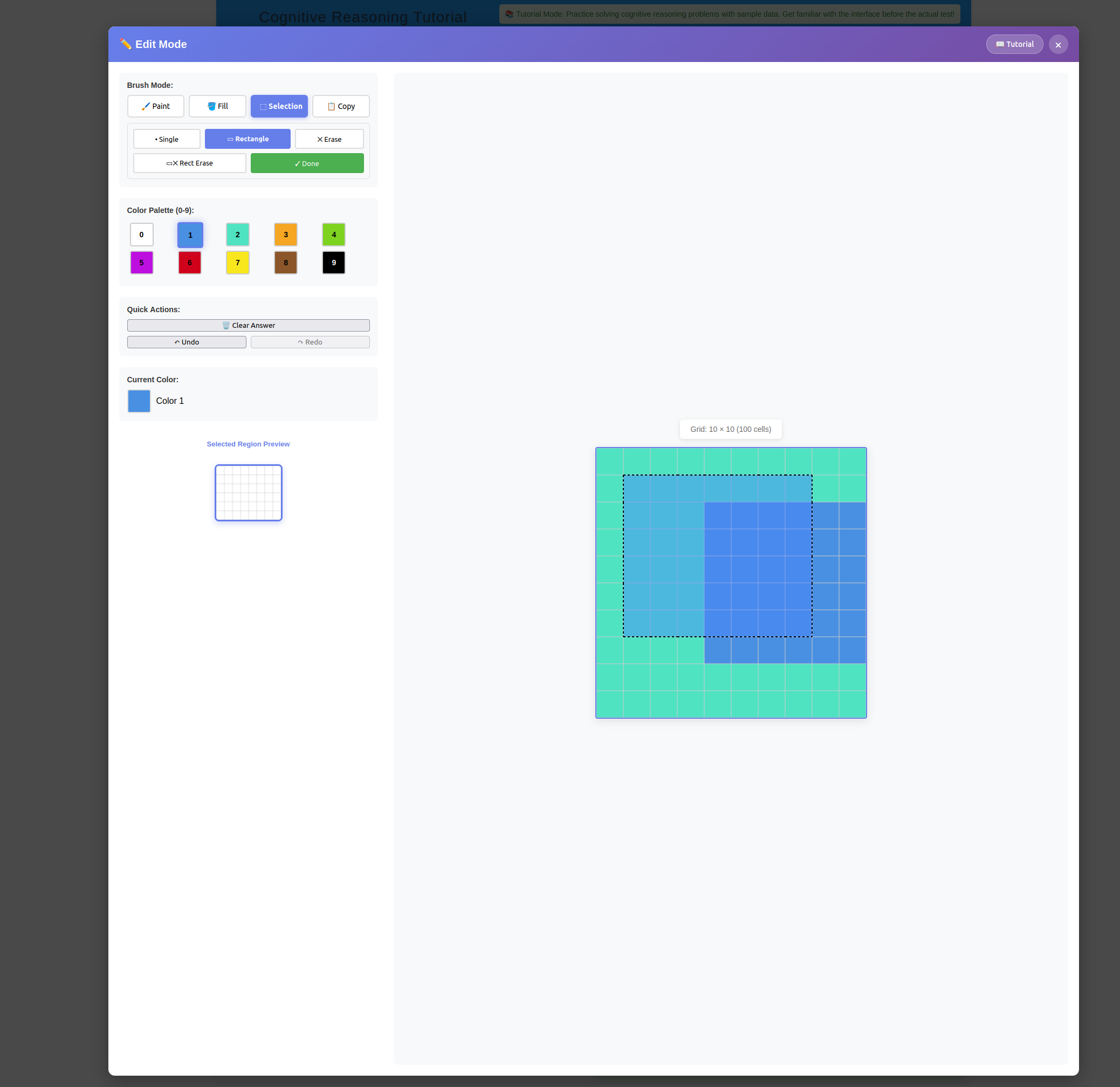}
    \caption{A screenshot of edit mode.}
    \label{fig:editmode}
\end{figure}
\paragraph{Welcome to Edit Mode! Step 1 of 8} Welcome to Edit Mode! This focused editing mode retains most of the features and shortcuts while adding powerful new \textbf{selection and copy} capabilities. Let's explore these new features to make your grid editing more efficient!

\paragraph{Single Cell Selection Step 2 of 8} When in selection mode, you can use \textbf{Single} selection to choose any shape of region. Click each cell to add it to your selection one by one, then click the \textbf{Done} button when finished.

A thumbnail of your selected region will appear below the toolbar. Click the thumbnail to view an enlarged window for easy inspection of your selection.

\paragraph{Rectangle Selection Step 3 of 8} You can also use \textbf{Rectangle} selectionmode to quickly select a rectangular region.
\begin{enumerate}
    \item \textbf{Left-click} to set the starting corner
    \item \textbf{Left-click} again to set the ending corner (diagonal)
    \item \textbf{Right-click} to cancel the current selection
\end{enumerate}

\paragraph{Erasing  Selections Step 4 of 8} If you want to remove parts of your selection, you can use \textbf{Erase} mode to delete individual cells, or use \textbf{Rectangle Erase} mode to quickly remove rectangular regions.

\textbf{Useful Shortcuts:}
\begin{enumerate}
    \item \texttt{A/D} Switch between selection submodes
    \item \texttt{F} Finish/confirm selection
\end{enumerate}

\paragraph{Copy \& Paste Step 5 of 8} When you have a selected region, you can choose \textbf{Copy} mode to apply it.

In copy mode, your selected region appears on the grid. You can \textbf{drag your mouses} to any position on the grid to move the selection. Once you find the perfect spot, click \textbf{Confirm Copy} to complete the operation.

Switch to any other mode to see the copied result!

\paragraph{Mirror Transforms Step 6 of 8} You can use \textbf{Horizontal Mirror} and \textbf{Vertical Mirror} to quickly flip your selection and create symmetrical patterns!

This is especially useful for creating symmetric designs and saves tons of time on repetitive operations.

\begin{figure}[t]
    \centering
    \includegraphics[width=0.4\linewidth]{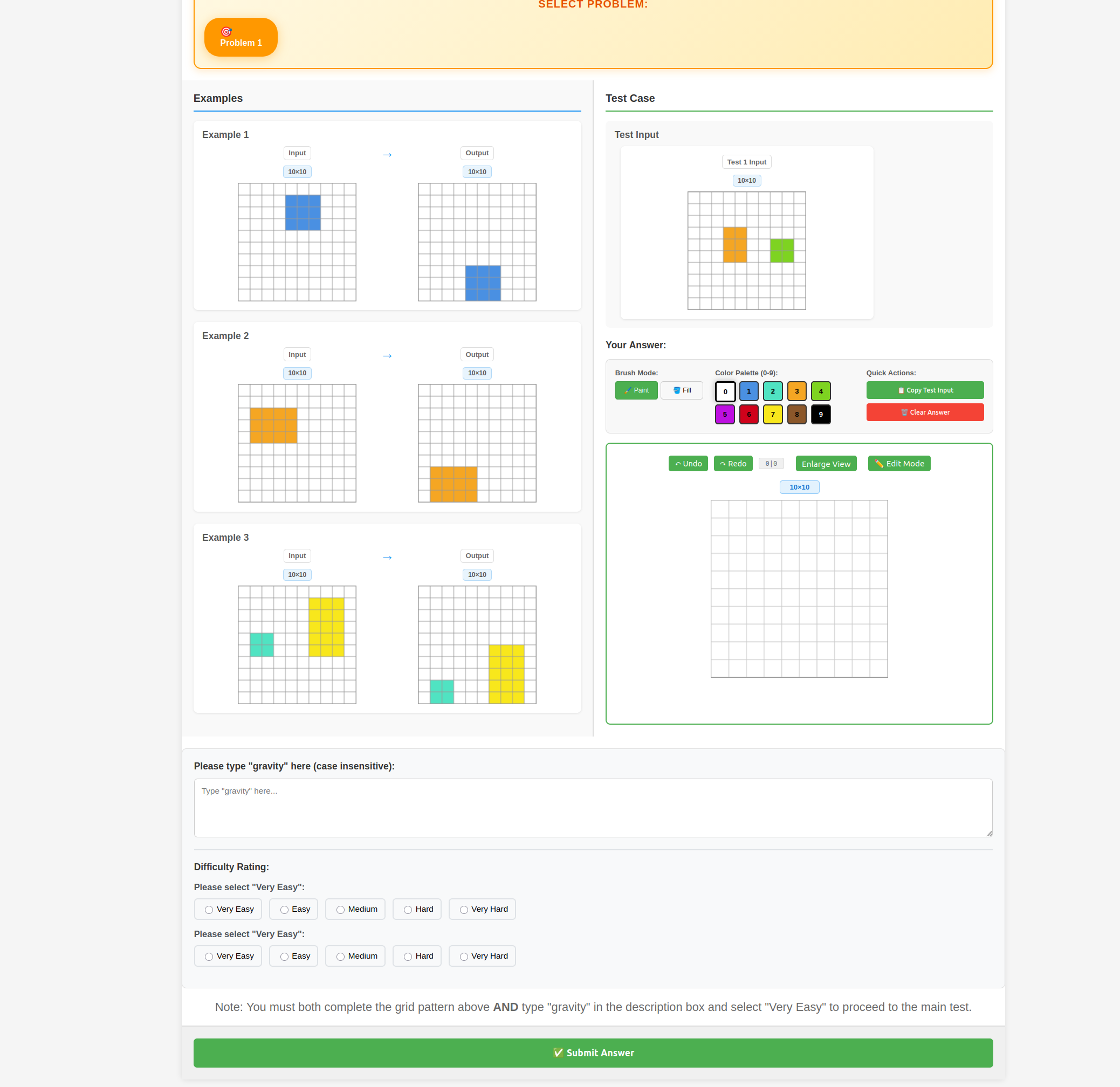}
    \caption{A screenshot of attention check.}
    \label{fig:attcheck}
\end{figure}

\textbf{Useful Shortcuts:}
\begin{enumerate}
    \item \texttt{W} Vertical mirror (flip up-down)
    \item \texttt{S} Horizontal mirror (flip left-right)
\end{enumerate}

\paragraph{Rotation Transforms Step 7 of 8} You can also use \textbf{Rotate Clockwise} and \textbf{Rotate Counter-Clockwise} to rotate your selection by 90 degrees.

After rotation, the selection automatically adjusts its position to stay within grid boundaries - smart and convenient!

\textbf{Useful Shortcuts:}
\begin{enumerate}
    \item \texttt{A} Rotate counter-clockwise 90 degrees
    \item \texttt{D} Rotate clockwise 90 degrees
\end{enumerate}

\paragraph{Tutorial Complete! Step 8 of 8} Congratulations on completing the Edit Mode tutorial! You've now mastered:
\begin{enumerate}
    \item \textbf{Selection Tools} Single/Rectangular selection \& erase
    \item \textbf{Copy \& Paste} Flexible region copying
    \item \textbf{Mirror Transforms} Horizontal/Vertical flips
    \item \textbf{Rotation Transforms} Clockwise/Counter-clockwise rotation
\end{enumerate}

Use these powerful new features with their shortcuts to edit grids more efficiently and create more complex patterns!

\textit{Tip: Press \texttt{Q} or \texttt{E} to quickly switch between these modes!}

\textbf{Attention check}

The participants are required to finish a task concerned with gravity, and correctly type the "gravity" in the description frame. For difficulty rating, the participants are required to choose "Very Easy" for both thinking difficulty and drawing difficulty.

All these text are provided in both English and Chinese, and you may find it out in the source code.

\subsection{Analysis details.} 
\label{appsubsec:humandetails}

All responses are evaluated by one author using consistent criteria, and we acknowledge the possibility of subjective annotation bias. Inter-rater agreement could not be measured. An analogy description is marked correct if it captures the core transformation rule; responses that do not express the core rule, including “I do not know,” are marked incorrect. Ambiguous cases and essential errors are conservatively classified as incorrect, and we also report strict exact-match accuracy without typo handling.

Overall, humans achieve $68.96\%$ average accuracy on ARCANA, including $10.61\%$ typo-level matches, while the Human ($\geq 2$ humans) accuracy reaches $96\%$, including $2.5\%$ typo-level matches. The results on Human ($\geq 2$ humans) show that typo judgments do not drive the conclusions of ARCANA's validity. In Figure~\ref{fig:eachresultdistribution}(a), each histogram bin aggregates tasks by their per-task accuracy (binned in $10\%$ intervals). 
For subjective difficulty, participants separately rate thinking and drawing difficulty using a five-level scale (very easy, easy, medium, hard, very hard), which we map to scores $1$-$5$. For each task, we average the two scores to obtain a single difficulty value. Figure~\ref{fig:eachresultdistribution}(b) plots each task as one point and overlays fitted trend lines under the two Typo conventions.

\subsection{ARCANA tasks have only one and clear ground truth.}
\label{appsubsec:onetruth}

In ARCANA, the ground truth is designed with the deterministic grid-level transformation rule for each task. In our human study, we provided participants with the ground-truth solutions after testing received no feedback suggesting that the answers were ambiguous. When some instances feel “unclear” to participants during testing , we believe this is largely driven by variability in far-analogy mapping (i.e., differences in concept familiarity and in identifying the intended target concept) rather than ambiguity in the grid-level rule itself. The observed accuracy primarily reflects participants’ capabilities.

In alignment with the construction principles of ARCANA, we will explain the ARCANA evaluation methods in detail.

\paragraph{"Multiple plausible analogical targets" do not imply multiple valid grid outputs.}
In the paper we mainly refer to the "multiple plausible analogical targets" but not "solutions". We mean that different participants may describe the same task using different analogical targets, while still implying the same underlying grid-level transformation rule (Construction Principle 3: multiple targets, one rule). For example, in \texttt{cat1} task, some participants described the patterns as "cats", while others described them as "little devils". These different targets can still yield the same correct grid transformation, and we treat both target descriptions as plausible as long as they faithfully reflect the intended rule.

\paragraph{For the grid answers, ARCANA is designed to have a unique ground-truth output under strict evaluation.}
For every task, we aim to ensure a single intended grid-level solution. Any designer-perceived "alternative" rules that might appear plausible are excluded by adding or modifying examples that serve as counter-evidence, so that only one rule remains consistent with the examples. While a single designer cannot guarantee absolute completeness, we made a strong effort to consider plausible alternatives and add counter-evidence examples where needed. Additionally, after the human study we provided participants with the ground-truth solutions, and we received no feedback indicating that any answers were ambiguous.

\paragraph{When multiple rules fit the examples, we select the human-aligned one—this is different from accepting multiple outputs.}
The issue of "multiple rules consistent with the examples" (Construction Principle 4: Human-aligned correctness) is closer to the cases where multiple rules might fit all the examples. In such cases, we resolve the under-specification during dataset construction by selecting the rule that is most natural to humans and best aligned with everyday concepts (i.e., lowest in cognitive burden), rather than treating multiple grid outputs as valid at evaluation time. For instance, in clock1 task, both "rotate the pointer on the right dial by $90^{\circ}"$ and "rotate the hour hand by $90^{\circ}$" may fit all examples, yet we select the latter as more natural and concept-aligned (as noted in the appendix). This motivation also informs why \textsc{ConceptSIR} includes an introspection step.

\paragraph{We do NOT accept unexpected grid solutions as correct.}
Accordingly, grid answers are scored by strict exact-match against the single ground-truth output. If we ever observe a non-matching output that seems plausibly valid, we treat it as evidence that the task is under-specified and would revise the task by adding/modifying examples and re-testing, rather than accepting multiple outputs at evaluation time. In contrast, for analogy targets, multiple descriptions are acceptable as long as they correctly reflect the task rule.

Thus, human judgment is used only for \textbf{typo annotation} (and ambiguous cases are counted as incorrect), not for validating alternative grid-level solutions or as a "trick" to make the accuracy results seem better. If multiple outputs appear valid, we treat it as a dataset issue to be fixed rather than an evaluation-time exception.

\subsection{Typo taxonomy}
\label{appsubsec:typotaxonomy}

\begin{figure*}[htpb]
    \centering
    \subcaptionbox{Typo answer.}{\includegraphics[width=0.35\linewidth]{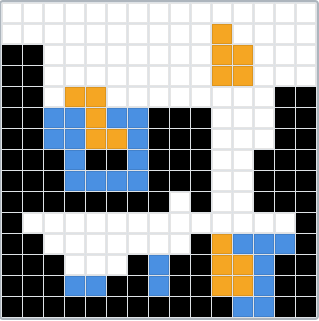}}
    \hfill
    \subcaptionbox{Correct answer.}{\includegraphics[width=0.35\linewidth]{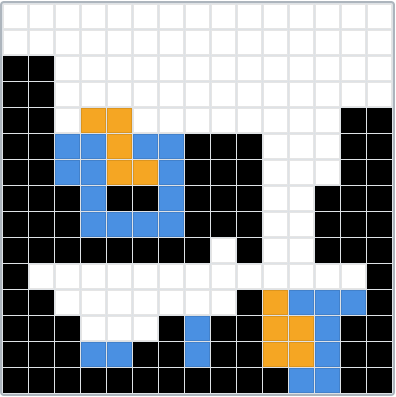}}
    \caption{A typo of \textbf{missed erase / incomplete edit after a state change} in \texttt{archimedes1}.}
    \label{fig:typo1}
\end{figure*}

\begin{figure*}[htpb]
    \centering
    \subcaptionbox{Typo answer.}{\includegraphics[width=0.35\linewidth]{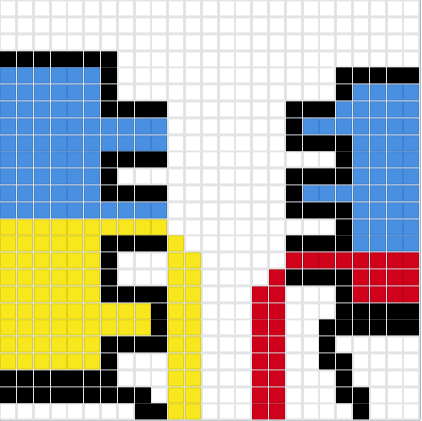}}
    \hfill
    \subcaptionbox{Correct answer.}{\includegraphics[width=0.35\linewidth]{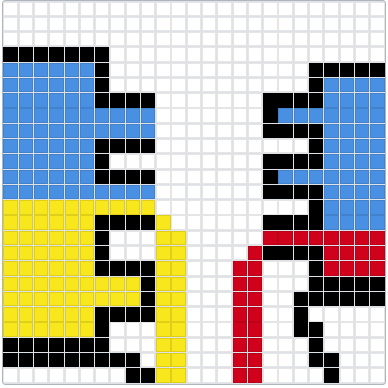}}
    \caption{A typo of \textbf{Accidental clicks or small omissions in complex structures} in \texttt{bottle3}.}
    \label{fig:typo2}
\end{figure*}

\begin{figure*}[htpb]
    \centering
    \subcaptionbox{Typo answer.}{\includegraphics[width=0.35\linewidth]{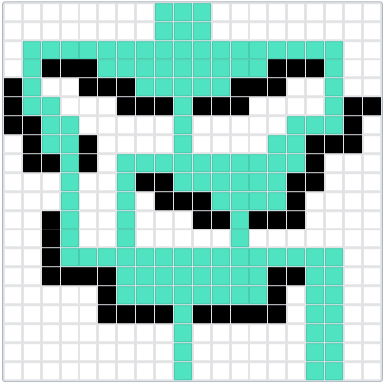}}
    \hfill
    \subcaptionbox{Correct answer.}{\includegraphics[width=0.35\linewidth]{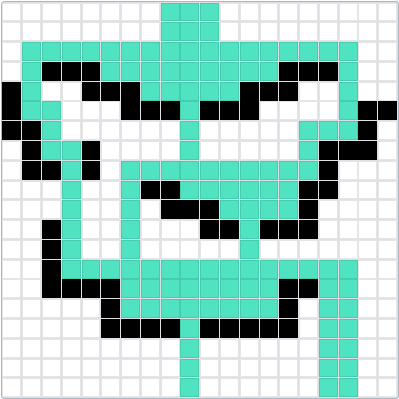}}
    \caption{A typo of \textbf{Color slip that flips a single cell to an adjacent/nearby color} in \texttt{water\_stream1}.}
    \label{fig:typo3}
\end{figure*}

\begin{figure*}[htpb]
    \centering
    \subcaptionbox{Typo answer.}{\includegraphics[width=0.35\linewidth]{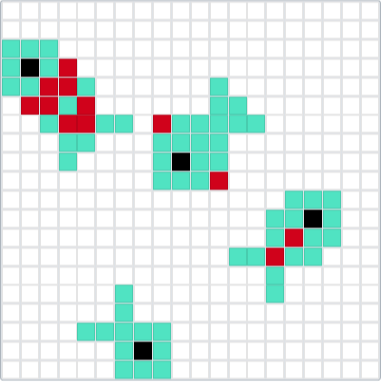}}
    \hfill
    \subcaptionbox{Correct answer.}{\includegraphics[width=0.35\linewidth]{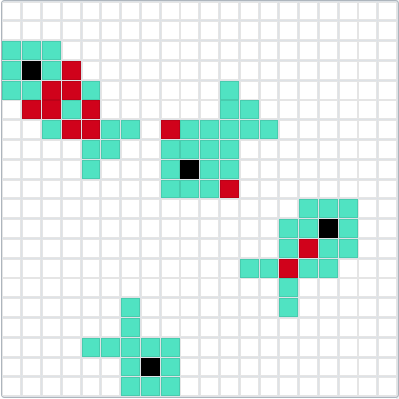}}
    \caption{A typo of \textbf{Off-by-one / off-by-two after translation or rotation of a pattern} in \texttt{fish1}.}
    \label{fig:typo4}
\end{figure*}

\begin{figure*}[htpb]
    \centering
    \subcaptionbox{Typo answer.}{\includegraphics[width=0.35\linewidth]{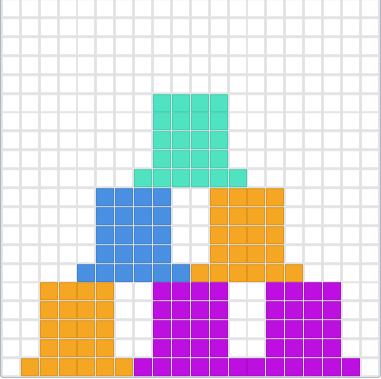}}
    \hfill
    \subcaptionbox{Correct answer.}{\includegraphics[width=0.35\linewidth]{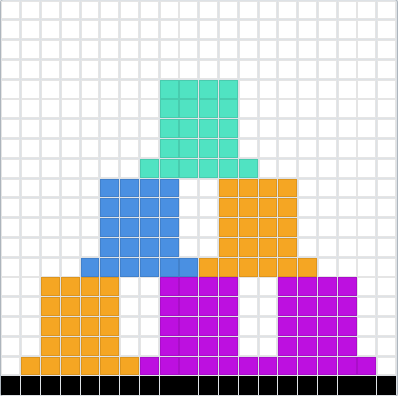}}
    \caption{A typo of \textbf{Failure to apply a copy operation, causing minor irrelevant errors} in \texttt{cup2}.}
    \label{fig:typo5}
\end{figure*}

To make our use of "typo" more interpretable, we provide a rough taxonomy and concrete examples below (task names in parentheses). We believe that explicitly enumerating recurring patterns makes typo handling more transparent. Furthermore, we provide some typos in Figure ~\ref{fig:typo1}, Figure ~\ref{fig:typo2}, Figure ~\ref{fig:typo3}, Figure ~\ref{fig:typo4}, Figure~\ref{fig:typo5}.

\begin{enumerate}
    \item \textbf{Missed erase / incomplete edit after a state change.}
    \begin{enumerate}
        \item Object falls into water but the pre-fall object is not erased (\texttt{archimedes1}).

\item Dye droplet falls into water but the falling droplet is not erased (\texttt{dye1}).

\item Magnet gets attracted but the previous magnet is not erased (\texttt{magnetic\_or\_chemical\_reaction1}).
    \end{enumerate}

\item \textbf{Accidental clicks or small omissions in complex structures.}
\begin{enumerate}
    \item A bottle-wall cell is painted as “empty” blue without affecting the pouring result (\texttt{bottle3}).

\item One source cell of water is missing when drawing the water flow (\texttt{bottle3}).
\end{enumerate}

\item \textbf{Color slip that flips a single cell to an adjacent/nearby color.}
\begin{enumerate}
\item When filling water inside a container, a floor cell is accidentally colored as water (\texttt{water\_stream1}).

\item During a rotation, the rotation-center cell changes from rope color to wall color (\texttt{ballistic\_pendulum1}).
\end{enumerate}

\item \textbf{Off-by-one / off-by-two after translation or rotation of a pattern.}
\begin{enumerate}
    \item After translating the fish pattern, the tail is off by one cell (\texttt{fish1}).

\item After wheel rotation, an extra wheel-center cell appears (\texttt{gear2}).
\end{enumerate}

\item \textbf{Failure to apply a copy operation, causing minor irrelevant errors.}
\begin{enumerate}
    \item Cups are stacked correctly, but a one-cell-high floor is not drawn (\texttt{cup2}).

\item One or two single cells in complicated pattern transformation (\texttt{kaleidoscope1}).
\end{enumerate}

\end{enumerate}

For ambiguous cases where it is difficult to determine whether an error should be considered a typo, we conservatively treat the answer as incorrect. Moreover, we always report strict exact-match accuracy without typo handling.

\section{LLMs evaluation details}
\label{appsec:LLMdetail}

\subsection{Cost and tokens}
\label{appsubsec:llmcost}

We provide the cost and tokens of every LLMs and human in Table \ref{tab:cost and tokens} for comparison. The currency in the table is USD. All LLM experiments are accomplished via Openrouter.

\begin{table*}[htpb]
    \centering
    \begin{tabular}{c|c|c|c|c|c}
    \hline
    ~ & GPT 5.2 & Gemini 3.1 pro & Grok 4.1 Fast & Claude Opus 4.5 & human \\
    \hline
    cost/task & \textbf{11.84} & 4.33 & 1.16 & 1.40 & 6.93 \\
    \hline
    tokens/task & 571K & 475K & \textbf{2.46M} & 169K & - \\
    \hline
    \end{tabular}
    \caption{
    The average cost and tokens LLMs and human spend per task.
    }
    \label{tab:cost and tokens}
\end{table*}


\subsection{LLMs results and baseline prompts}
\label{appsubsec:llmprompts}

While chain-of-thought prompts might be easy to understand, analogical prompts actually include the prompt about associating the task with real-world objects or phenomena which is targets. Beyond the two high-level prompting setups reported in Table~\ref{tab:baseline} (i.e., \texttt{CoT (language)} and \texttt{CoT, ana (VL)}), we additionally use two prompt variants within each setup. For each variant, we compute \texttt{pass@3} and obtain a set of tasks solved by that variant. The reported score for a setup is then computed from the union of these task-level success sets (i.e., a task is counted as solved if it is solved by either prompt variant under \texttt{pass@3}). We also observe occasional ``truncated" outputs (the matrix is cut off mid-generation) and formatting errors (e.g., mismatched brackets). Cost and token statistics include all attempts, including failures.

Because our evaluation uses a fixed number of trials (\texttt{pass@3}) under specific prompting setups, the reported accuracies should be viewed as an empirical upper bound under our testing budget rather than an intrinsic ceiling of each model. Differences in cost per task can also affect how aggressively one can sample, refine, or verify outputs, further entangling performance with budget. In addition, run-to-run variability is substantial; therefore, the analysis that follows is intended to surface plausible tendencies and failure modes observed in our runs, not to claim universal properties of the models.

The prompts used in baseline are provided below:

CoT system prompt 1:

\begin{tcolorbox}[colback=green!5!white,colframe=green!50!black,arc=4mm]
   You are an expert ARC (Abstraction and Reasoning Corpus) task solver. I will provide you with the matrix of the question, you could decide the rule to solve the question. In matrix, every example is provided as input-output pair, and question is provided as input. There might be multiple questions, they all share the same rule as the examples. Please follow the principles and answer the given test.

  Principles:
  
    - Avoid ad-hoc exceptions that only fix one example but break generality.
    
    - If there are multiple tests, your answer could look like: [grid1, grid2]

    Output format:
    
    ANSWER:
    
    [Your answer for all tests - no explanations and no markdown formatting. If there are multiple tests, follow the principles to sequentially place the answers, and your answer should be like [grid1, grid2] where grid1 and grid2 are the answers of different questions.]
\end{tcolorbox}

CoT system prompt 2:
\begin{tcolorbox}[colback=green!5!white,colframe=green!50!black,arc=4mm]
You are an expert ARC (Abstraction and Reasoning Corpus) task solver. I will provide you with the matrix of the question, you could decide the rule to solve the question. In matrix, every example is provided as input-output pair, and question is provided as input. There might be multiple questions, they all share the same rule as the examples.

Your mandatory workflow:

1. Understand:

    - Carefully inspect all training input-output pairs.
    
    - Describe in works what changes from input to output.

2. Hypothesize:

    - Propose at least two distinct candidate rules that would explain the transformations.
    
    - For each candidate, give a clear natural language and short pseudocode.
    
    - Guarantee the diversity of the different candidate rules, try to come up with more rules from diverse aspects.

3. Test:

    - For each candidate rule, mentally apply it to every training input.

    - Explicitly compare the predicted output with the ground-truth output.
    
    - Mark each candidate as 'fully consistent', 'partially consistent', or 'inconsistent'.
\end{tcolorbox}
\begin{tcolorbox}[colback=green!5!white,colframe=green!50!black,arc=4mm]
4. Select:

    - Choose the simplest rule that is fully consistent with all training examples.
    
    -If none is fully consistent, refine or combine candidates until at least one fits all data.

5.Apply:

    - Apply the selected rule to each test input grid to obtain predicted test outputs.

Principles:

    - Never stop at the first plausible rule; always consider multiple hypotheses.
    
    - Avoid ad-hoc exceptions that only fix one example but break generality.
    
    - If there are multiple tests, your answer could look like: [grid1, grid2]
    
    - Prefer object-based reasoning (connected components, shapes, repetition, symmetry, etc.)

Output format:
    
REASONING:

[Your reasoning progress here, include all of your workflow.]

ANSWER:

[Your answer for all tests - no explanations and no markdown formatting. If there are multiple tests, follow the principles to sequentially place the answers, and your answer should be like [grid1, grid2] where grid1 and grid2 are the answers of different questions.]

\end{tcolorbox}

CoT, Analogy system prompt 1:
\begin{tcolorbox}[colback=green!5!white,colframe=green!50!black,arc=4mm]
You are an expert ARC (Abstraction and Reasoning Corpus) task solver. Now you will be provided with one ARC task, the rule of the task could be analogous to a target in real life which could be some objects or some natural phenomena. I will provide you with the matrix of the question, you could decide the rule to solve the question. In matrix, every example is provided as input-output pair, and question is provided as input. There might be multiple questions, they all share the same rule as the examples. Please follow the principles and answer the given test. You are encouraged to use analogy to complete the test.

You will be provided with the visualization of the examples and the questions for better analogy. In the visualization of examples, every column is an input-output pair, the first row is input and the second row is output. From left to right, the columns correspond to example1, example2, and so on. In visualization of questions, there is only one row for the reason that there is no ground truth. Also, from left to right, the columns correspond to question1, question2, and so on.

Relation between the color and the number is provided: 0: white, 1: blue, 2: teal, 3: orange, 4: green, 5: purple, 6: red, 7: yellow, 8: brown, 9: black.

Principles:

    - Avoid ad-hoc exceptions that only fix one example but break generality.
    
    - If there are multiple tests, your answer could look like: [grid1, grid2]

Output format:

ANALOGY:

[Your analogous target here, use one word or sentence to describe the rule.]

ANSWER:

[Your answer for all tests - no explanations and no markdown formatting. If there are multiple tests, follow the principles to sequentially place the answers, and your answer should be like [grid1, grid2] where grid1 and grid2 are the answers of different questions.]

\end{tcolorbox}

CoT, Analogy system Prompt 2:
\begin{tcolorbox}[colback=green!5!white,colframe=green!50!black,arc=4mm]
You are an expert ARC (Abstraction and Reasoning Corpus) task solver. Now you will be provided with one ARC task, the rule of the task could be analogous to a target in real life which could be some objects or some natural phenomena. I will provide you with the matrix of the question, you could decide the rule to solve the question. In matrix, every example is provided as input-output pair, and question is provided as input. There might be multiple questions, they all share the same rule as the examples. Please follow the principles and answer the given test.

Your mandatory workflow:

1. Understand:

    - Carefully inspect all training input-output pairs.
    
    - Describe in works what changes from input to output.

2. Hypothesize:

    - Propose at least four distinct candidate analogies and their corresponding rules that would explain the transformations.
    
    - For each candidate, give a clear natural language and short pseudocode.
    
    - Guarantee the diversity of the different candidate rules, try to come up with more rules from diverse aspects.

3. Test:

    - For each candidate rule, mentally apply it to every training input.
    
    - Explicitly compare the predicted output with the ground-truth output.
    
    - Mark each candidate as 'fully consistent', 'partially consistent', or 'inconsistent'.

4. Select:

    - Choose the simplest rule that is fully consistent with all training examples.
    
    -If none is fully consistent, refine or combine candidates until at least one fits all data.

5.Apply:

    - Apply the selected rule to each test input grid to obtain predicted test outputs.

You will be provided with the visualization of the examples and the questions for better analogy. In the visualization of examples, every column is an input-output pair, the first row is input and the second row is output. From left to right, the columns correspond to example1, example2, and so on. In visualization of questions, there is only one row for the reason that there is no ground truth. Also, from left to right, the columns correspond to question1, question2, and so on.

Relation between the color and the number is provided: 0: white, 1: blue, 2: teal, 3: orange, 4: green, 5: purple, 6: red, 7: yellow, 8: brown, 9: black.

Principles:

    - Never stop at the first plausible rule; always consider multiple hypotheses.
    
    - Avoid ad-hoc exceptions that only fix one example but break generality.
    
    - If there are multiple tests, your answer could look like: [grid1, grid2]
    
    - Prefer object-based reasoning (connected components, shapes, repetition, symmetry, etc.)

Output format:

REASONING:

[Your reasoning progress here, include all of your workflow.]

ANALOGY:

[Your analogous target here, use one word or sentence to describe the rule.]

ANSWER:

[Your answer for all tests - no explanations and no markdown formatting. If there are multiple tests, follow the principles to sequentially place the answers, and your answer should be like [grid1, grid2] where grid1 and grid2 are the answers of different questions.]
\end{tcolorbox}

\subsection{Response-level accuracy}
\label{appsubsec:llmreslevelacc}
Noted that the LLM accuracy and human accuracy are under different metrics in Table \ref{tab:arcana} (pass@12 vs. pass@1), we report the response-level accuracy across all outputs of the LLMs. We believe this will provide an empirical estimate of single-response accuracy across repeated runs.

\begin{table*}[ht]
    \centering
    \begin{tabular}{c|c|c|c|c|c|c}
    \hline
    \multirow{3}*{Model} & \multicolumn{6}{|c}{ARCANA} \\
    \cline{2-7}
    ~ & \multicolumn{2}{|c|}{Total} & \multicolumn{2}{|c|}{CoT (language)} & \multicolumn{2}{|c}{CoT + Ana (VL)} \\
    \cline{2-7}
    ~ & +Typo & -Typo & +Typo & -Typo & +Typo & -Typo  \\
    \hline
    GPT 5.2 & \textbf{18.86\%} & \textbf{17.03\%} & \textbf{17.54\%} & \textbf{15.87\%} & \textbf{20.18\%} & \textbf{18.18\%} \\ 
    \hline
    Gemini 3 Pro & 13.12\% & 8.57\% & 13.96\% & 9.17\% & 12.28\% &  7.97\% \\ 
    \hline
    Grok 4.1 Fast & 6.46\% & 4.39\% & 6.70\% & 4.47\% & 6.22\% & 4.31\% \\ 
    \hline
    Claude Opus 4.5 & 2.23\% & 1.71\% & 2.47\% & 1.99\% & 1.99\% & 1.44\%\\ 
    \hline
    Human (Average, pass@1) & 68.96\% & 58.35\% & - & - & - & -\\
    \hline
    \end{tabular}
    \caption{
Response-level accuracy on ARCANA. For each LLM, Total aggregates all 12 responses per task across four prompt variants and three runs. \texttt{CoT (language)} uses language-only chain-of-thought prompting; \texttt{CoT + Ana (VL)} adds an analogical prompt with visual inputs. Missing or failed generations are counted as incorrect. \texttt{+Typo} counts minor typographical errors as correct, whereas \texttt{-Typo} uses strict exact match.
    }
    \label{tab:baselineresponse}
\end{table*}

\subsection{Detailed analysis on analogy}
\label{appsubsec:llmanalogy}
\begin{figure*}[htpb]
    \centering
    \includegraphics[width=0.95\linewidth]{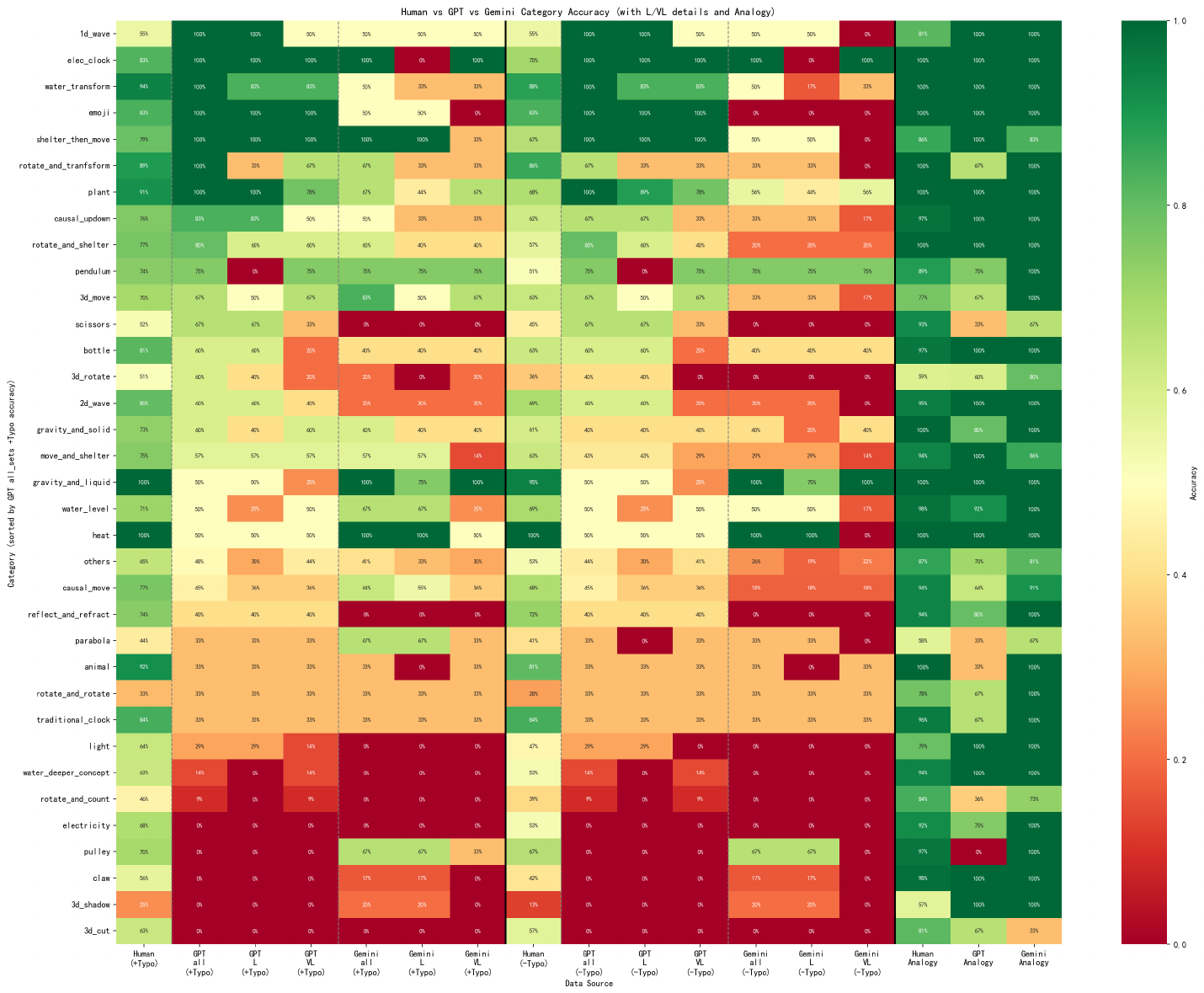}
    \caption{Accuracy distribution of GPT 5.2, Gemini 3 Pro and human across various task categories, including answers accuracy and analogy accuracy.}
    \label{fig:accheatmap}
\end{figure*}

\begin{figure*}[htpb]
    \centering
    \includegraphics[width=0.55\linewidth]{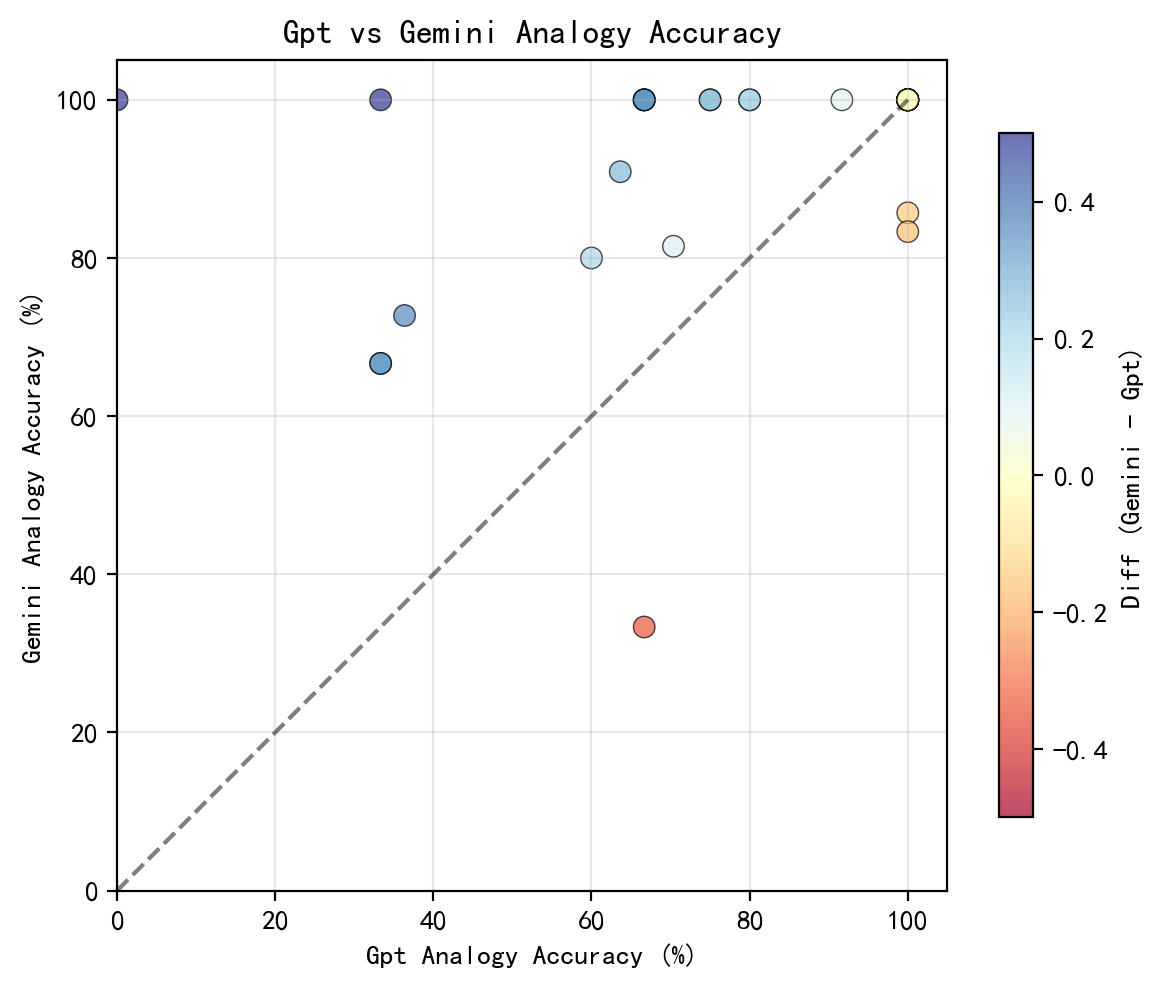}
    \caption{
    Category-wise analogy accuracy: GPT~5.2 vs.\ Gemini~3 Pro. Each point denotes one ARCANA category, with GPT~5.2 analogy accuracy on the x-axis and Gemini~3 Pro analogy accuracy on the y-axis. The dashed diagonal indicates parity; points above it correspond to categories where Gemini produces more correct analogical targets than GPT. Point color encodes the accuracy difference (Gemini minus GPT).
    }
    \label{fig:anagptgemini}
\end{figure*}

Results are provided in Figure~\ref{fig:accheatmap}, including the accuracy of analogy and answers on different categories for human, GPT 5.2, and Gemini 3 Pro.
We compare GPT~5.2 and Gemini~3 Pro analogy accuracy across task categories using a scatter plot (Figure~\ref{fig:anagptgemini}), where each point corresponds to a category. Points above the diagonal indicate categories in which Gemini's analogy accuracy exceeds GPT's.

Beyond \texttt{pulley} clarified in the main body, we observe additional category-specific gaps. For example, in \texttt{animal}, Gemini achieves higher accuracy than GPT, whereas GPT outperforms Gemini on several categories that require more precise manipulations (e.g., \texttt{move\_and\_shelter}, \texttt{shelter\_then\_move}, and \texttt{3d\_cut}). These differences suggest complementary capability profiles and may reflect differences in pretraining exposure.

We also note qualitative cases where a model's responses contain explicit real-world interpretations even under language-only prompting: Gemini describing a mechanical lifting system for \texttt{pulley1} under language-only prompting. This might result from the inner understanding or operation process of Gemini on the given task, which could be understanding only on language (for \texttt{pulley} category, the multiple relationship of moving distance could be helpful to recognize the analogical targets) or similarity between visual hidden states and language hidden states.

\subsection{Differences between language and vision-language prompting}
\label{appsubsec:llmdiffonvision}
To localize modality effects, we further inspect which task categories are solved under each prompting modality for Gemini and GPT. While most categories exhibit only small differences, we identify several cases with large modality gaps. In the \texttt{electric\_clock} category, Gemini achieves 0\% accuracy under language-only prompting but 100\% under vision--language prompting. Similarly, in the \texttt{pendulum} category, GPT improves from 0\% under language-only prompting to 75\% under vision--language prompting. For \texttt{moon\_and\_tide1}, Gemini attains 100\% under vision--language prompting but fails under language-only prompting, and its successful response explicitly attributes the pattern to lunar-phase variation.

Conversely, we also observe categories where performance decreases after introducing vision-language prompting. These decreases do not follow a consistent pattern across Gemini and GPT, suggesting that the impact of visual inputs is model-dependent and intertwined with output stochasticity, rather than reflecting a uniform modality effect.

A closer inspection of correct analogy responses suggests systematic differences between GPT~5.2 and Gemini~3 Pro. We reiterate that LLM outputs are highly random; our observations are based on six attempts per task and should be interpreted as descriptive hypotheses rather than deterministic conclusions.

\textbf{Case studies.} For \texttt{moon\_and\_tide1}, Gemini explicitly identifies ``the lunar cycle and its effect on tides,'' whereas GPT's best response remains at a surface-level grid description (e.g., ``Yellow controls a 2-row tide shift of the blue sea''). For \texttt{newton\_cradle1} and \texttt{newton\_cradle2}, Gemini readily recognizes the intended physical concept, while GPT tends to produce less grounded interpretations (e.g., ``Pipe insertion with opposite-side ejection''). Despite these failures, GPT can occasionally produce high-quality analogies: for \texttt{candy\_crush1}, it generates a correct match-3 interpretation (``swap, clear lines, gravity, cascade''). Notably, this task also exhibits substantial variation in human analogy accuracy, plausibly reflecting differences in participant background (our participants are undergraduates in Chinese universities).

\textbf{Lexical preferences.} We additionally analyze word frequencies in model-generated analogies. Gemini exhibits a strong preference for the term ``gravity,'' which appears in 59 out of 200 tasks, compared to 7 occurrences in human targets and 13 in GPT's targets. While gravity is relevant for some categories (e.g., water or pendulum), Gemini sometimes invokes it even when the core phenomenon is unrelated, suggesting that its analogical search may be biased toward a limited set of familiar physical explanations. We also observe that Gemini frequently uses terms such as ``system'' and ``mechanic,'' whereas these words are absent in GPT's analogy responses, indicating model-specific analogical styles and domain biases.

\textbf{Other models.} Finally, we briefly examine Grok and Claude. Their distributions of analogy-correct categories differ from GPT and Gemini, suggesting distinct strengths across categories; overall, however, both models exhibit weaker analogical reasoning performance in our setting. All quantitative analogical accuracy results are provided in Table \ref{tab:llmanaresult}.

\begin{table}[t]
\centering
\small
\begin{tabular}{lcc}
\toprule
Model & Correct / Total & Accuracy \\
\midrule
GPT-5.2          & 159 / 200 & 79.50\% \\
Gemini 3 Pro     & 183 / 200 & \textbf{91.50\%} \\
Claude Opus 4.5  & 75 / 200  & 37.50\% \\
Grok 4.1 Fast    & 29 / 200  & 14.50\% \\
\bottomrule
\end{tabular}
\caption{Analogy-recognition accuracy of the four baseline LLMs on the 200 ARCANA tasks.}
\label{tab:llmanaresult}
\end{table}

\subsection{Additional discussion on typo effects.}
\label{appsubsec:llmtypo}
\begin{figure*}[ht]
    \centering
    \subcaptionbox{Typo answer.}{\includegraphics[width=0.3\linewidth]{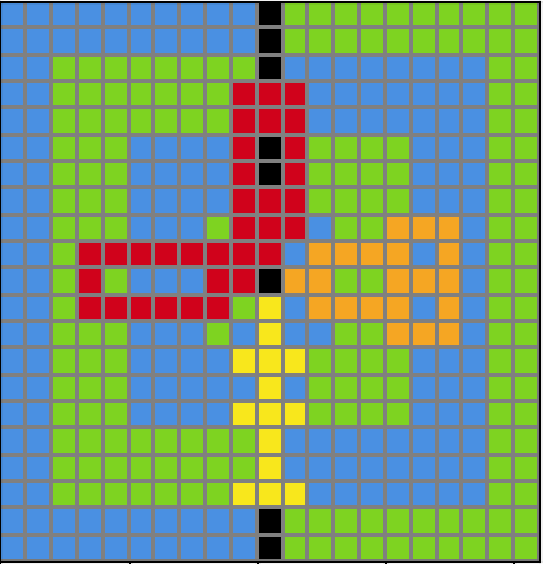}}
    \hfill
    \subcaptionbox{Correct answer.}{\includegraphics[width=0.3\linewidth]{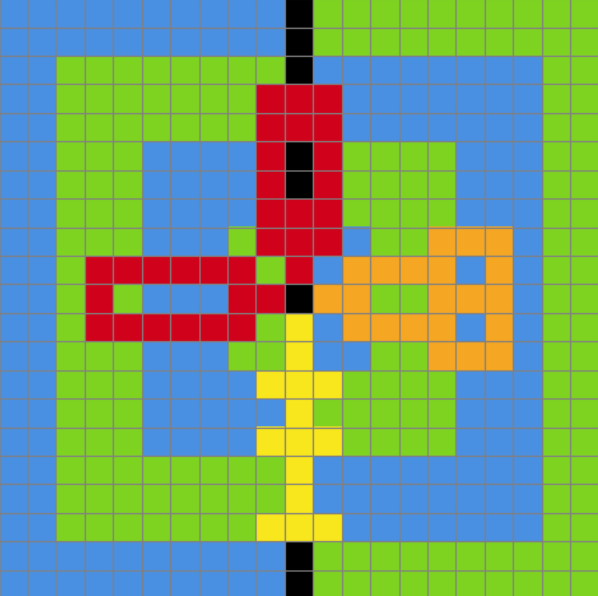}}
    \caption{A typo by GPT~5.2 in \texttt{fan1} task. The typo might be considered as one of the \textbf{Accidental clicks or small omissions in complex structures}.
    }
    \label{fig:llmtypo1}
\end{figure*}

\begin{figure*}[ht]
    \centering
    \subcaptionbox{Typo answer.}{\includegraphics[width=0.3\linewidth]{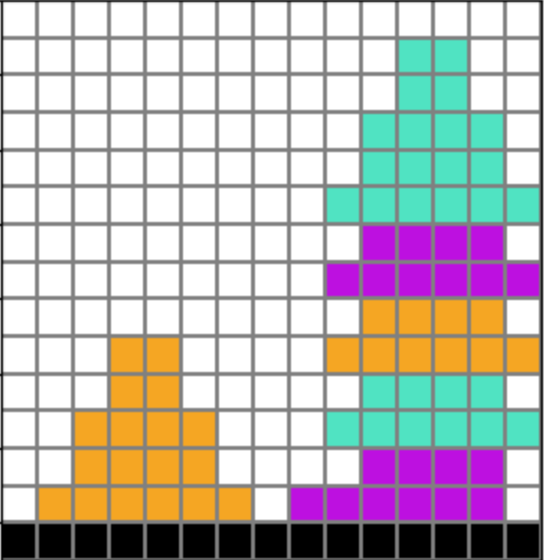}}
    \hfill
    \subcaptionbox{Correct answer.}{\includegraphics[width=0.3\linewidth]{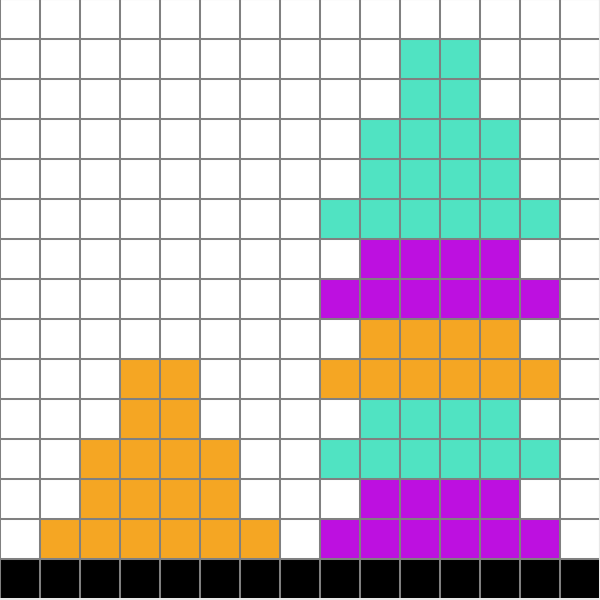}}
    \caption{A typo by Gemini~3 Pro in \texttt{cup1} task. The typo might be considered as one of the \textbf{Off-by-one / off-by-two after translation or rotation of a pattern}.
    }
    \label{fig:llmtypo2}
\end{figure*}

When LLMs are asked to response the grid answers directly, it is reasonable to assume that typos might occur due to the lack of sufficient tokens or the randomness. After evaluating all answers manually, we provide some typos of LLMs in Figure~\ref{fig:llmtypo1}, Figure~\ref{fig:llmtypo2}. In the first typo on \texttt{fan1} task in Figure~\ref{fig:llmtypo1}, GPT~5.2 mistook three grids irregularly,which is similar to the \textbf{Accidental clicks or small omissions in complex structures} in human typos. While in the second typo on \texttt{cup1} task in Figure~\ref{fig:llmtypo2} by Gemini~3 Pro, the upper part of the right cup tower are totally moved to right by one grid. In the answer matrix, the typo could be regarded as adding a zero in every first 13 rows between these two patterns. Therefore, the typo might be considered as one of the \textbf{Off-by-one / off-by-two after translation or rotation of a pattern} in human typos.

Typos of LLMs are slightly similar to the typos of humans, while remaining some innate difference. In Figure~\ref{fig:llmtypo2}, although the pattern is off-by-one after translation (more accurately, off-by-one-column), humans do not prefer to tear up a whole pattern of a cup, which is visually unnatural and obvious. In human typos shown in Figure~\ref{fig:typo4}, a single grid of the fish tail disappears, which is relatively hard to find. The difference might reveal the weakness in vision or the lack of visual check-up of LLMs.

When typo responses are counted as correct, both humans and Gemini exhibit a relatively larger increase in accuracy, whereas GPT shows a smaller change. We refer to this pronounced gain as a typo leap. Qualitatively, this pattern is consistent with a failure mode where the model captures the correct concept/rule but makes minor execution-level mistakes in the final grid.
Several factors may contribute to the cross-model differences. GPT's strong baseline accuracy may leave limited headroom for gains under \texttt{+Typo}. Conversely, Gemini's stronger analogy accuracy may lead to more conceptually correct solutions that are close to the correct output but contain small implementation slips. Given the stochasticity of LLM outputs, these explanations should be interpreted as hypotheses grounded in our observations.

\subsection{Category-level details.}
\label{appsubsec:llmcatdetail}
While humans generally outperform LLMs category-wise, we observe a small number of categories where certain models exceed the human average, suggesting complementary strengths in manipulation-heavy settings, as shown in Figure \ref{fig:accheatmap}. Conversely, we identify categories with pronounced deficits in relation to humans. For example, in \texttt{water\_deep\_concept} category (tasks grounded in deeper liquid-related concepts such as buoyancy and phase change), humans achieve 63\% accuracy, while Gemini achieves 0\% and GPT achieves 14\% under \texttt{+Typo}. In \texttt{animal}, humans achieve 92\% accuracy, while both Gemini and GPT achieve 33\%. Many \texttt{animal} tasks emphasize topological recognition (e.g., identifying patterns as ducks or cats) under varying bounding boxes and geometric transforms (rotation/mirroring), which appears easy for humans yet remains difficult for current models.

\begin{figure*}[htpb]
    \centering
    \includegraphics[width=0.6\linewidth]{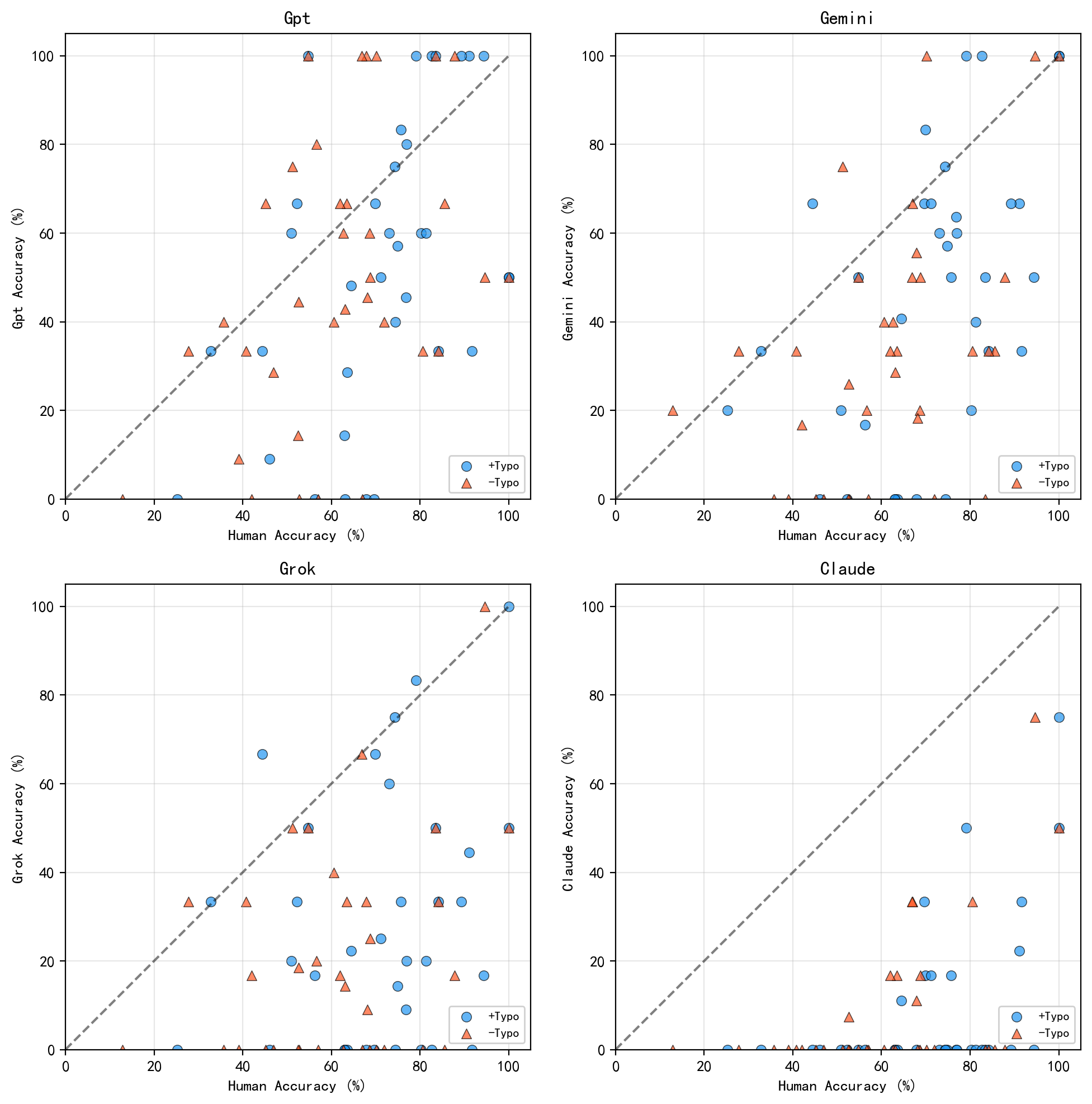}
    \caption{
    Category-wise accuracy of humans versus LLMs. Each subplot corresponds to one model (GPT~5.2, Gemini~3 Pro, Grok~4.1 Fast, and Claude Opus~4.5). Each point denotes a task category, with human accuracy on the x-axis and model accuracy on the y-axis. Circles report \texttt{+Typo} (typos counted as correct) and triangles report \texttt{-Typo} (typos counted as wrong). The dashed diagonal indicates parity; points below it correspond to categories where humans outperform the model, and points above it correspond to categories where the model exceeds the human average.
    }
    \label{fig:accscatter}
\end{figure*}

\subsection{Paired bootstrap comparison between humans and LLMs.}
\label{appsubsec:llmlagforhumans}
We further compare LLM baselines with humans using a paired task-level bootstrap over the $200$ ARCANA tasks. For each task, we compare the model's binary solve outcome with the human per-task accuracy and compute the mean paired difference. Across all four LLMs, the $95\%$ bootstrap confidence intervals of LLM--Human differences are strictly below zero under both strict and typo-tolerant evaluation; even the strongest baseline remains $10.13$ percentage points below humans under strict evaluation and $17.67$ percentage points below under typo-tolerant evaluation. This indicates a statistically reliable gap between current LLMs and human performance on ARCANA.

We compare each LLM baseline with humans using a paired task-level bootstrap over the $200$ ARCANA tasks. For task $i$, let $m_i\in\{0,1\}$ denote whether the LLM solves the task, and let $h_i$ denote the human per-task accuracy. We compute the paired difference $\Delta=\frac{1}{T}\sum_i(m_i-h_i)$ and estimate $95\%$ confidence intervals by resampling tasks with replacement. Negative values indicate that the LLM underperforms humans.


\section{Implementation details of \textsc{ConceptSIR}}
\label{appsec:conceptsirdetail}
\subsection{Framework details}
\label{appsubsec:conceptsirparts}

Based on stage-wise observations, we use Gemini~3 Pro for concept proposal and slippage, and GPT~5.2 for code-centric generation and refinement. The design is heuristically based on preliminary analogy experiments and our LLM results (we found that Gemini performs best in the concept proposal and slippage stages, while GPT excels in code-centric processes). We also have tried multiple configurations (e.g., GPT-only vs. GPT+Gemini), while acknowledging substantial room for improvement.

\paragraph{Concept proposal and slippage details.}
Our initial experiments suggest that LLM reasoning is highly sensitive to prompt context, which can reduce analogical diversity and hinder ARCANA solving. Concept Slippage provides a mechanism for counteracting this rigidity: by explicitly activating multiple candidate concepts, the model is encouraged to traverse higher-level relational abstractions and recruit a broader set of associated concepts. 

Given the scale of modern LLMs, it is plausible that their training data covers many everyday targets relevant to ARCANA. From a Bayesian perspective, we further hypothesize that LLM neuron activations can be viewed as a (possibly shallower) analogue of Concept Slippage in humans. We therefore augment prompts with diverse analogy candidates to better exploit the combinational creativity of LLMs and to activate a wider range of concepts in their latent knowledge base. The four proposal dimensions (color/shape/dynamics/holistic recognition) reflect common human priors, but other dimensions can be substituted depending on the problem domain.

\paragraph{Concept execution details.}
We treat code as a commitment to an operational semantics: it forces the model to specify edge-case handling and prevents post-hoc reinterpretation that is easy in free-form text. In practice, we generate one candidate program per target and evaluate it by executing on the example inputs and comparing outputs with ground truth; discrepancies are then passed to the subsequent reflection and introspection stages.

\paragraph{Introspection details.}
We view human preference as arising from aesthetic and cognitive priors (e.g., simplicity, symmetry, and conceptual coherence). Introspection serves as a mechanism for detecting when a program matches demonstrations for the ``wrong reasons'' (overly ad hoc predicates or unnatural object choices) and steering revisions toward more coherent, human-aligned abstractions.

\paragraph{Reflection details.}
We treat reflection as a family of code-modification operators, including syntax fixes and logic-level refinements, aimed at better instantiating the analogical rule. At each reflection step, we update multiple candidate programs in parallel. For each candidate, we compute its discrepancy on the examples (between predicted and ground-truth outputs) and select the lowest-discrepancy program as the current best. The LLM is then prompted to analyze the strengths and weaknesses of this best program; when updating the remaining candidates, we trigger code mutation with a fixed probability, asking the LLM to combine useful implementation patterns from the best program with the candidate's own partial strengths.

At the concept level, we similarly introduce mutations over analogical targets. Concept slippage proposes larger-variance alternative targets to avoid narrow local minima. Concept augmentation expands a target by attaching relevant subconcepts that may be essential to the core transformation; for example, in \texttt{bead\_chain1} task, gravity alone is insufficient without an additional stacking/pile-up constraint to reproduce the packed final configuration.

 Our framework is more similar to an agent, and we aim to shed light on the effectiveness of our framework based on the resulting reasoning trajectories. We also seriously state that our framework leaves a lot space for improvement, for example, concept execution form could be changed from code to other more robust forms, additional explicit memory modules could be added in order to conclude the ``experience" during concept proposal (more effective prompting aspects), more aligned rule with human (more human-aligned introspection), human feedback for better alignment, and so on.

 Moreover, on the assumption that analogical reasoning could benefit to most of problems solving, we believe our framework could be generalized to common analogical reasoning problems, further common tasks, since former work shows analogical prompting could improve the performance of models \cite{yasunaga2023large}. The concept proposal could be generalized to multi-modalities, such as vision and auditory, for handling more generalized problems (e.g., associating the music heard before for the sentiments in the past scenario). Execution can be generalized to more general forms like motion sequences and natural language. All of the above assumptions are just a beginning, and we believe there remains a lot aspects for exploration.

\subsection{Theoretical grounding: Copycat inspiration for ConceptSIR.}
\label{appsubsec:conceptsirgrounding}
\texttt{ConceptSIR} is explicitly inspired by Copycat. In Copycat theory by \textit{Douglas R. Hofstadter and Melanie Mitchell}, task-specific structures in the Workspace activate concept nodes in the Slipnet; activation can "slip" probabilistically to nearby concepts, which in turn triggers Codelets that operate in the Workspace to build/modify structures and relations on the task, further updating activations in an iterative loop.

ConceptSIR follows this spirit: the role of the Slipnet is played by the LLM’s prior knowledge; concept slippage occurs implicitly within the LLM conditioned on context; and the analog of Codelets is our executable concept instantiation (code execution). Mutation provides a mechanism to iteratively refine both concept proposals and executable rules based on feedback.

From a Bayesian perspective, concept slippage is grounded in the prior knowledge and the concepts activated by the task in the Copycat theory, which is loosely analogous to the prior and the likelihood in Bayesian theory. Similarly, LLMs generate new concepts conditioned on the knowledge learned from training data and the task contexts. We believe these two mechanisms could be analogous to each other at a high level, which makes using LLMs for concept slippage a practical instantiation.

\subsection{Prompts for each module}
\label{appsubsec:conceptsirprompt}
Here we provide prompts we used in each module.

Concept proposal system prompt:
\begin{tcolorbox}[colback=green!5!white,colframe=green!50!black,arc=4mm]
You are a helpful assistant that good at image association. Now I need you to play a role in solving ARC question. I will provide you a picture contains some grids, every column is an input-output pair, the first row is input and the second row is output. Every question can be analogous to objects or phenomenon in real world. Your task is to associate the pattern in the picture with real world, and give me your reason. 

Please mainly consider the color of the patterns (the shape of the patterns/the dynamic change of the patterns/the full image of the patterns) in the examples, use one word or one sentence to describe what the picture looks like in real world, in other words, what can be the picture analogous in real world, and give me your reason.
\end{tcolorbox}

Concept slippage system prompt:
\begin{tcolorbox}[colback=green!5!white,colframe=green!50!black,arc=4mm]
You are a helpful assistant that good at image association. Now I need you to play a role in solving ARC question. I will provide you a picture contains some grids, every column is an input-output pair, the first row is input and the second row is output. Every question can be analogous to objects or phenomenon in real world. Then I will provide you four possible proposals and their reasons, the proposals are about color, shape, dynamic change and full image between input and output respectively. You can combine these proposals and associate new concepts with these 
proposals, or just choose one of the proposals. Futhermore, you must provide me with the rule in analogy and reason. Give me the top three most possible targets and rules in JSON format.

Your output should look like this:

```json

    \{
    
        "Analogy\_list":[
        
            \{
                "Analogy": "", "Reason": "" 
            \},
            
            \{
                "Analogy": "", "Reason": ""
            \},
            
            \{
                "Analogy": "", "Reason": ""
            \} 
        ]

    \}

```
\end{tcolorbox}

Code generation (execution) system prompt:
\begin{tcolorbox}[colback=green!5!white,colframe=green!50!black,arc=4mm]
You are a helpful assistant that is good at solving ARC question. Now I will provide you an ARC question, the question can be analogous to objects or phenomenon in real world. Besides, I will provide a possible target and rule from former agent. The rule of the question usually is delicate, which might include moving a parsing region, counting the number of a certain structure, rotating an area, copy some structures that is meaningful, keeping some grids in certain color or number conserved, occlusion relationship, and so on. Therefore I need you to complete a code to execute the rule.

The code should include parsing part and execution part including different functions. In parsing part, the code will locate the target according to some certain rules, for example, according to location, colors and corresponding numbers, shapes, or combination of the possible properties. The parsing group could be an irregular list, which might include some grids that are not necessary to be compared. For example, we use the 10 grids to represent the unnecessarily compared grids because the normal distribution of the grids is 0-9, the parsed object is [[1,1,10,10],[1,1,10,10],[1,1,1,1],[1,1,1,1]] represent a square consisted of 1 but without the upper right 2*2 square. When parsing or moving, the operation should ignore the 10's grids. You may consider more properties according to different question. And when parsing, the pattern in input grids might be sheltered, which means the patterns in input grids and output grids might be part of the full pattern, and you might need to infer where the full patterns are, and use code to accomplish the occlusion relationship. The output of the code might be targets and corresponding point sets. In execution part, you need to use the rule and the parsing sets from parsing part to generate answer grid, input and output are both lists. Sometimes you might need to judge the moving direction step by step, take the falling along the steps as an example, you might need to judge the possible moving direction in every time step for one pattern or grid. It's only an example, you need to judge the algorithm according to different settings.

Code should NOT include ' if '\_\_name\_\_' == 'main' '. The code should only include the functions and a execute function to call these functions, whose name should be execute() to let me call the function easily. The input should be a list, and output should be a list, that is, the function should be 'def execute(input\_list : list) $\rightarrow$ list'. Considering safety, do NOT include 'os.system', 'subprocess.call', 'eval(', 'exec(', '\_\_import\_\_', 'open(', 'rmdir', 'remove'. Input matrix will ALWAYS be one input which is a two-dimention list. A possible input is: [[1, 0], [0, 1]], please pay attention. Mention that the code should NOT include endless loop.

Please ensure that the dimensions of the input and output are the same. Please do NOT return [output] like this where the output in the square brackets has the same dimension as the input.

I will provide you both the visualization of the question and the matrixs, every column is an input-output pair, the first row is input and the second row is output. Relation between the color and the number is provided: 0: white, 1: blue, 2: teal, 3: orange, 4: green, 5: purple, 6: red, 7: yellow, 8: brown, 9: black. The matrix is a list which might include three or more dicts. Every dict is an example, including input matrix and output matrix.

Please follow this format:

REASONING:

[Think step by step about:

1. Which targets should I consider to represent the rule? I might need to consider multiple targets, and I should consider all of the examples.
\end{tcolorbox}
\begin{tcolorbox}[colback=green!5!white,colframe=green!50!black,arc=4mm]
2. What characteristics should I use to parse the grids and get point sets? Color, shape, location, or combination of above? I should consider the difference on different examples.
3. What algorithm/approach should be used? Move, rotate, mirror, copy a certain structure, or combination of above? I should consider all of the examples.

4. What are the key steps in the implementation? This will guide me to finish the code.]

CODE:

[Your complete, executable Python code here - no markdown formatting]

EXPLANATION:

[Brief explanation of how the code works]
\end{tcolorbox}

Code rectification (check the errors) system prompt:
\begin{tcolorbox}[colback=green!5!white,colframe=green!50!black,arc=4mm]
You are a programming assistant. Now I have a execute() function and some small functions. When I call execute(input\_list), some error happened. I need you to rectify the code to get the code run successfully. The input should be a list, and output should be a list, that is, the function should be 'def execute(input\_list : list) $\rightarrow$ list'. Considering safety, do NOT include 'os.system', 'subprocess.call', 'eval(', 'exec(', '\_\_import\_\_', 'open(', 'rmdir', 'remove'. Input matrix will ALWAYS be one input which is a two-dimension list. A possible input is: [[1, 0], [0, 1]], please pay attention. Mention that the code should NOT include endless loop.

I will provide you the code and the error message, please follow this format:
REASONING:

[Think step by step about:

1. What's the reason why the error happens? Did I use a wrong input format?

2. How can I rectify the code?]

CODE:

[Your complete, executable Python code here - no markdown formatting]

EXPLANATION:

[Brief explanation of how the code works]
\end{tcolorbox}

Common code reflection:
\begin{tcolorbox}[colback=green!5!white,colframe=green!50!black,arc=4mm]
You are a helpful assistant that is good at solving Abstraction and Reasoning Corpus (ARC) Test. Now I will provide you an ARC question, the question can be analogous to objects or phenomenon in real world. Besides, I will also provide you a function execute() which is expected to get the output from input. The function is designed to represent the rules of the question. The rules of the question is usually delicate, which might include moving a

parsing region, counting the number of a certain structure, rotating an area, copy some structures that is meaningful, keeping some grids in certain color or number conserved, and so on. What I need you to do is to analyze where the function is wrong according to the output of the function and ground truth output.

In this execution, the analogy of the question might be augmented with the sub-concepts attached to the original analogy, for example, commonsense such as conditions, conservations, time, causality and so on. If the augmentation happens, I will provide the proposals that will direct you to come up with the instructions of modifying code. Please take them into consideration! 

In the following part, I will provide you the analogy of the question, the code of the function, explanation of how the code work, the input-output-groundtruth grids picture, the detailed difference grids between output and ground truth of each example and their coordinates, along with the matrices of their input, output and ground truth. 

The code includes parsing part and execution part, and at last I will call execute(input: list) -> list to get my output. the explanation is expected to inform how the code works and functions' function. The parsing group could be an irregular list, which might include some grids that are not necessary to be compared. For example, we use the 10 grids to represent the unnecessarily compared grids because the normal distribution of the grids is 0-9, the parsed object is [[1,1,10,10],[1,1,10,10],[1,1,1,1],[1,1,1,1]] represent a square consisted of 1 but without the upper right 2*2 square. When parsing or moving, the operation should ignore the 10's grids. You 
\end{tcolorbox}
\begin{tcolorbox}[colback=green!5!white,colframe=green!50!black,arc=4mm]
may consider more properties according to different question. And when parsing, the pattern in input grids might be sheltered, which means the patterns in input grids and output grids might be part of the full pattern, and you might need to infer where the full patterns are, and use code to accomplish the occlusion relationship. In execution part, you need to use the rule and the parsing sets from parsing part to generate answer grid, input and output are both lists. Sometimes you might need to judge the moving direction step by step, take the falling along the steps as an example, you might need to judge the possible moving direction in every time step for one pattern or grid. It's only an example, you need to judge the algorithm according to different settings. 

Input-output-groundtruth picture will include three grids, from left to right lies the input of the function, the output from code, and the ground truth output expected. For detailed comparison, I will provide you with the difference grids picture between output and ground truth, and the matrices of input and output and ground truth. In the difference picture, there will be two grids. The gray grids means the same part of the output and ground truth, and the other colored part means the different grids between output of the code and ground truth, including white, blue, teal, orange, green, purple, red, yellow, brown, black. The left grid shows what the code output is in these part, and the right grid shows what the ground truth is in this part. The coordinates are up to down, left to right, and start from 0. For example, (0,13) means the first row from up to down and the 14th column from left to right.

Relation between the color and the number is provided: 0: white, 1: blue, 2: teal, 3: orange, 4: green, 5: purple, 6: red, 7: yellow, 8: brown, 9: black.

Please follow the following format:

REASONING:

[Think step by step about:

1. How does the different parts of the function work? Analyze according to the explanation. 

2. On which input the code generate a wrong output? Why does it generate a wrong output? The problem is on the parsing part or the execution part, or both? I need to analyze all of the examples.

3. Focus on the difference grids, why are these grid wrong? Which part of the code are the main reason causing the difference? Or should I add some rules? If I should add a rule, can I find the corresponding explanation of the target of analogy in real world object or phenomenon?

4. Are algorithm/approach be falsely used or lacked of? Should I add, delete or change the operation of move, rotate, mirror, copy a certain structure, or combination of above? I should consider all of the examples.

5. What are the key steps in the implementation of modifying the function? This will help me instruct the latter agent to modify the code.
]

INSTRUCTIONS:

[The instructions of modifying the code -- try to be comprehensive.]
\end{tcolorbox}

Analyzing best code (code mutation) system prompt:
\begin{tcolorbox}[colback=green!5!white,colframe=green!50!black,arc=4mm]
You are a helpful assistant that is good at solving Abstraction and Reasoning Corpus (ARC) Test. Now I will provide you an ARC question, the question can be analogous to objects or phenomena in real world. Besides, I have a function execute(), I need to improve the code.

To achieve this, I will provide you a better code, you should analyze the advantages of the better code, the disadvantages of the code to be improved, and give me the instructions to improve the code.

In the following part, I will provide you the code to be improved, the analogy and reason of the code to be improved, the better code and the analogy and reason of it. Besides, I will provide you with the visualization of all of the examples of the question, every column is an input-output pair, the first row is input and the second row is output. The code includes parsing part and execution part, and at last I will call execute(input: list) $\rightarrow$ list to get my output. The reason is expected to inform you how the code is built and how the code works. 

Relation between the color and the number is provided: 0: white, 1: blue, 2: teal, 3: orange, 4: green, 5: purple, 6: red, 7: yellow, 8: brown, 9: black.

Please follow the following format:

REASONING:

[Think step by step about:

1. Analyze these two codes, understand how these two codes work according to the reasons and explanations.

2. Analyze the advantages of the better code, how these advantages are realized, and the details of realization. 

3. Analyze the advantages of the code to be improved, remember to reserve these advantages.

4. Analyze the disadvantages of the better code and the code to be improved, remember to avoid them.

5. What are the key steps in the implementation of modifying the function? This will help me propose the instructions.]
\end{tcolorbox}
\begin{tcolorbox}[colback=green!5!white,colframe=green!50!black,arc=4mm]
INSTRUCTIONS:

[The instructions of modifying the code -- try to be comprehensive.]
\end{tcolorbox}

Target generation in concept slippage (concept mutation, reflection) system prompt:
\begin{tcolorbox}[colback=green!5!white,colframe=green!50!black,arc=4mm]
You are a helpful assistant that good at image association. Now I need you to play a role in solving ARC question. I will provide you a picture contains some grids, every column is an input-output pair, the first row is input and the second row is output. Every question can be analogous to objects or phenomenon in real world. Then I will provide you a main analogy and its reason. You should start from this analogy and its reason, associate another analogy and explain the reason. Besides, I provide you more analogies and their reasons.  You can combine these proposals and associate new concepts with these proposals, but please mainly consider the main analogy. Furthermore, you must provide me with the rule in analogy and reason. Give me ONE most possible targets and rules in JSON format.

Your output should look like this:

```json

\{

    "Analogy": "", "Reason": ""
    
\}
```

\end{tcolorbox}

Code generation in concept mutation (reflection) prompt:
\begin{tcolorbox}[colback=green!5!white,colframe=green!50!black,arc=4mm]
You are a helpful assistant that is good at solving ARC question. Now I will provide you an ARC question, the question can be analogous to objects or phenomenon in real world. Besides, I will provide a possible target and rule from former agent. The rule of the question usually is delicate, which might include moving a parsing region, counting the number of a certain structure, rotating an area, copy some structures that is meaningful, keeping some grids in certain color or number conserved, and so on. Therefore I need you to complete a code to execute the rule. Besides, I will provide you a former analogy along with its reason and code. Please understand the code and analogy first, then you can learn from its advantages to construct your code.

The code should include parsing part and execution part including different functions. In parsing part, the code will locate the target according to some certain rules, for example, according to location, colors and corresponding numbers, shapes, or combination of the possible properties. You may consider more properties according to different question. The parsing group could be an irregular list, which might include some grids that are not necessary to be compared. For example, we use the 10 grids to represent the unnecessarily compared grids because the normal distribution of the grids is 0-9, the parsed object is [[1,1,10,10],[1,1,10,10],[1,1,1,1],[1,1,1,1]] represent a square consisted of 1 but lost the upper right 2*2 square. When parsing or moving, the operation should ignore the 10's grids. You may consider more properties according to different question. And when parsing, the pattern in input grids might be sheltered, which means the patterns in input grids and output grids might be part of the full pattern, and you might need to infer where the full patterns are, and use code to accomplish the occlusion relationship. The output of the code might be targets and corresponding point sets. In execution part, you need to use the rule and the parsing sets from parsing part to generate answer grid, input and output are both lists. Sometimes you might need to judge the moving direction step by step, take the falling along the steps as an example, you might need to judge the possible moving direction in every time step for one pattern or grid. It's only an example, you need to judge the algorithm according to different settings.

Code should NOT include ' if '\_\_name\_\_' == 'main' '. The code should only include the functions and a execute function to call these functions, whose name should be execute() to let me call the function easily. The input should be a list, and output should be a list, that is, the function should be 'def execute(input\_list : list) $\rightarrow$ list'. Considering safety, do NOT include 'os.system', 'subprocess.call', 'eval(', 'exec(', '\_\_import\_\_', 'open(', 'rmdir', 'remove'. Input matrix will ALWAYS be one input which is a two-dimension list. A possible input is: [[1, 0], [0, 1]], please pay attention. Mention that the code should NOT include endless loop.

Please ensure that the dimensions of the input and output are the same. Please do NOT return [output] like this where the output in the square brackets has the same dimension as the input.
\end{tcolorbox}
\begin{tcolorbox}[colback=green!5!white,colframe=green!50!black,arc=4mm]
I will provide you both the visualization of the question examples and the matrices, every column is an input-output pair, the first row is input and the second row is output. Besides, I will provide you with a former analogy along with its reason and code, remember to understand the advantages and disadvantages of it. Relation between the color and the number is provided: 0: white, 1: blue, 2: teal, 3: orange, 4: green, 5: purple, 6: red, 7: yellow, 8: brown, 9: black. The matrix is a list which might include three or more dicts. Every dict is an example, including input matrix and output matrix.

Please follow this format:

REASONING:

[Think step by step about:

1. Analyze the advantages of the former code, I can learn from these advantages.

2. Analyze the disadvantages of the former code, remember to avoid them.

3. Which targets should I consider to represent the rule? I might need to consider multiple targets, and I should consider all of the examples.

4. What characteristics should I use to parse the grids and get point sets? Color, shape, location, or combination of above? I should consider the difference on different examples.

5. What algorithm/approach should be used? Move, rotate, mirror, copy a certain structure, or combination of above? I should consider all of the examples.

6. What are the key steps in the implementation? This will guide me to finish the code.]

CODE:

[Your complete, executable Python code here - no markdown formatting]

EXPLANATION:

[Brief explanation of how the code works]
\end{tcolorbox}

Concept augmentation in concept mutation (reflection) system prompt:
\begin{tcolorbox}[colback=green!5!white,colframe=green!50!black,arc=4mm]
You are a helpful assistant that is good at solving Abstraction and Reasoning Corpus (ARC) Test. Now I will provide you an ARC question, the question can be analogous to objects or phenomenon in real world. Besides, I will also provide you the existing analogy and the corresponding reason, along with the code which is realized to represent
the rule of the analogy. What's more, I visualize the output of the examples inputs from the code and the ground truth what they are expected to be, and the highlighted difference grids and coordinates of the different grids. I need you to understand the analogy and code now, and think: Is the analogy now is enough to solve the question? Did I miss some sub-concepts attached to the analogy target now, which is important to solving the problem? Should I add the sub-concepts? If I choose to add new sub-concepts, how can I instruct the code modification?

Detailedly description of sub-concepts: there might be some laws of the physics attached to the objects or phenomena in the analogy, which is not shown in the code and analogy reasons. For example, commonsense such as conditions, conservations, time, causality etc. You need to consider which one you could add to explain the difference now.

If you think that it's necessary to add some sub-concepts to the existing analogy, please return your augmented analogy and corresponding augmented reason based on the existing analogy, and the corresponding proposals to generate the instructions of modifying the code. The proposals should instruct the direction of how to modify the code, so that the latter agents could generate the detailed instructions of modifying the code. 

If you think there's no need to augment the concepts now, just return the existing analogy and its reasons I provided to you. And you can return "there's no sub-concepts augmented" in the proposal part. 

In the following part, I will provide you the analogy of the question, the code of the function, explanation of how the code work, the input-output-groundtruth grids picture, the detailed difference grids between output and ground truth of each example and their coordinates, along with the matrices of their input, output and ground truth. 

The code includes parsing part and execution part, and at last I will call execute(input: list) $\rightarrow$ list to get my output.
The explanation is expected to inform how the code works and functions' function. The parsing group could be an irregular list, which might include some grids that are not necessary to be compared. For example, we use the 10
grids to represent the unnecessarily compared grids because the normal distribution of the grids is 0-9, the parsed object is [[1,1,10,10],[1,1,10,10],[1,1,1,1],[1,1,1,1]] represent a square consisted of 1 but lost the upper right 2*2 square. When parsing or moving, the operation should ignore the 10's grids. You may consider
\end{tcolorbox}
\begin{tcolorbox}[colback=green!5!white,colframe=green!50!black,arc=4mm]
more properties according to different question. And when parsing, the pattern in input grids might be sheltered, which means the patterns in input grids and output grids might be part of the full pattern, and you might need to infer where the full patterns are, and use code to accomplish the occlusion relationship. The output of the code might be targets and corresponding point sets. In execution part, you need to use the rule and the parsing sets from parsing part to generate answer grid, input and output are both lists.

Sometimes you might need to judge the moving direction step by step, take the falling along the steps as an example, you might need to judge the possible moving direction in every time step for one pattern or grid. It's only an example, you need to judge the algorithm according to different settings.

Code should NOT include ' if '\_\_name\_\_' == 'main' '. The code should only include the functions and a execute function to call these functions, whose name should be execute() to let me call the function easily. The input should be a list, and output should be a list, that is, the function should be 'def execute(input\_list : list) $\rightarrow$ list'. Considering safety, do NOT include 'os.system', 'subprocess.call', 'eval(', 'exec(', '\_\_import\_\_', 'open(', 'rmdir', 'remove'. Input matrix will ALWAYS be one input which is a two-dimention list. A possible input is: [[1, 0], [0, 1]], please pay attention. Mention that the code should NOT include endless loop. 

Input-output-groundtruth picture will include three grids, from left to right lies the input of the function, the output from code, and the ground truth output expected. For detailed
comparison, I will provide you with the difference grids picture between output and ground truth, and the matrices of input and output and ground truth. In the difference picture, there will be two grids. The gray grids means the same part of the output and ground truth, and the other colored part means the different grids between output of the code and ground truth, including white, blue, teal, orange, green, purple, red, yellow, brown, black. The left grid shows what the code output is in these part, and the right grid shows what the ground truth is in this part. The coordinates are up to down, left to right, and start from 0. For example, (0,13) means the first row from up to down and the 14th column from left to right.

Relation between the color and the number is provided: 0: white, 1: blue, 2: teal, 3: orange, 4: green, 5: purple, 6: red, 7: yellow, 8: brown, 9: black.

Please follow this format:

REASONING:

[Think step by step about:

1. Inspect the input-output-groundtruth pictures. On which input the code generate a wrong output? Why does it generate a wrong output? I need to analyze all of the examples.

2. Are there any sub-concepts attached to the existing analogy and reasons that I can use to explain the wrong outputs? Are they already realized by the code?

3. Are the sub-concepts I find align with the objects and the phenomena in real world in the existing analogy? If reasonable, how can I augment the existing analogy and its reason to embody the sub-concepts?

4. I need to give some proposals to instruct the code modifying. I must explain what sub-concepts I add which is not realized by the existing code. This will point out the direction for the modification of the code.]

ANALOGY:

\{"Analogy": "Your augmented analogy here, based on the existing analogy", 

"Reason": "Your augmented reasons here, also based on the existing reasons"\}

PROPOSALS:

[Brief proposals for code modifying, about what you add and which directions to modify the code]
\end{tcolorbox}

Introspection on examples system prompt:
\begin{tcolorbox}[colback=green!5!white,colframe=green!50!black,arc=4mm]
You are a helpful assistant that is good at solving Abstraction and Reasoning Corpus (ARC) Test. Now I will provide you an ARC question, the question can be analogous to objects or phenomena in real world. Besides, I have already generated a code to compute the output grids of questions, but the code is not accurate right now. I need you to check the outputs of the questions, and compare them with the inputs and outputs of the examples. Please point out the abnormal part of the outputs of the questions, and give me instructions to improve the code.

In the following part, I will provide you the code to be improved, the analogy and reason of the code to be improved. Besides, I will provide you with the visualization of all of the examples of the question, every column is an input-output pair, the first row is input and the second row is output. The code includes parsing part and execution part, and at last I will call execute(input: list) -> list to get my output. The reason is expected to inform you how the code is built and how the code works.
\end{tcolorbox}
\begin{tcolorbox}[colback=green!5!white,colframe=green!50!black,arc=4mm]
Besides, I will provide you the visualization of the question input and the corresponding output of the input from the code. In the visualization, the left grid is the question, and the right grid is the output of the code. There is no ground truth. You need to inspect the examples and find the abnormal parts of the output of the question from the code.

Relation between the color and the number is provided: 0: white, 1: blue, 2: teal, 3: orange, 4: green, 5: purple, 6: red, 7: yellow, 8: brown, 9: black.

Please follow the following format:

REASONING:

[Think step by step about:

1. Inspect the examples, check the input-output pairs, compare them with the analogy and reason I provide.

2. Check the inputs and outputs of the questions from the code. Are there any parts that are abnormal? 

3. Check the code and code reasoning. How can I improve the codes to avoid the abnormal parts?

4. Should I introduce more sub-concepts attached to the analogy to generate better and less abnormal results.

5. What are the key steps in the implementation of modifying the function? This will help me propose the instructions.]

INSTRUCTIONS:

[The instructions of modifying the code -- try to be comprehensive.]
\end{tcolorbox}

Introspection on question system prompt:
\begin{tcolorbox}[colback=green!5!white,colframe=green!50!black,arc=4mm]
You are a helpful assistant that is good at solving Abstraction and Reasoning Corpus (ARC) Test. Now I will provide you an ARC question, the question can be analogous to objects or phenomena in real world. Now I will provide you the analogy of the question along with its reason, and moreover the rule explanation about code. The code of which has been finished, and after testing I find that the code is right on all examples. However, I need you to judge whether the question output from the code is right or not. Mention that there is no ground truth for question, what you need to consider is whether the output of the question from the code is consistent with the examples or not. READ THE EXPLANATION CAREFULLY in case that you fully understand the rule. If the output does not align with the examples I provide, please tell me the output is not aligned with which example in what aspect, then give me the signal, and provide the instruction about how to improve the code, where should I pay attention.

In judging dict you should return, there will be two keys: "Judgment" and "Instructions". The value of "Judgment" should ONLY be "True" or "False", no other things. If the "Judgment" you give me is "False", please provide the "Instructions" how I should improve the code; if not, the "Instructions" can be anything.

In the following part, I will provide you the analogy of the question along with its reason, the visualization of the examples, the visualization of the question input and the corresponding output of the input from the code.

In the visualization of examples, every column is an input-output pair, the first row is input and the second row is output. In the visualization of question input and output from code, the left grid is the question, and the right grid is the output of the code.

Relation between the color and the number is provided: 0: white, 1: blue, 2: teal, 3: orange, 4: green, 5: purple, 6: red, 7: yellow, 8: brown, 9: black.

Please follow this format:

REASONING:

[Think step by step about:

1. Understand the analogy and its reason of the problem. How can it align with the examples?

2. Inspect the question input and the output generated by the code. Is there something abnormal? Or not align with the examples and analogy? If there exists one example that does not align with the output, please point out which example and where is not aligned with the output. Attention, please point out the specific location of both code output and the example, and explain the rule. Be honest, if they are aligned, you could just answer 'True'.

3. If there is something weird, please point out and give me some instructions to rectify.]

PROPOSALS:

\{"Judgment": "True or False", "Instructions": "Your instructions here, guiding me the directions of consideration."\}
\end{tcolorbox}

Code refinement after reflection system prompt:
\begin{tcolorbox}[colback=green!5!white,colframe=green!50!black,arc=4mm]
You are a helpful assistant that is good at solving Abstraction and Reasoning Corpus (ARC) Test. Now I will provide you an ARC question, the question can be analogous to objects or phenomenon in real world. Besides, I will also provide you a wrong function execute() which is expected to get the output from input, and the instructions how to modify the code. The function is designed to represent the rules of the question. The rules of the question is usually delicate, which might include moving a parsing region, counting the number of a certain structure, rotating an area, copy some structures that is meaningful, keeping some grids in certain color or number conserved, and so on. I need you to modify the code to better represent the rules, according to the instructions I provide. Moreover, there might be the modifying instructions of analyzing other better code and the output of the questions, you can consider these instructions according to the circumstances.

In the following part, I will provide you the analogy of the question, the code of the function, explanation of how the old code work, the input-output-groundtruth grids picture, along with the matrices of their input, output and ground truth. 

Please ensure that the dimensions of the input and output are the same. Please do NOT return [output] like this where the output in the square brackets has the same dimension as the input.

The code includes parsing part and execution part, and at last I will call execute(input: list) -> list to get my output. The explanation is expected to inform how the code works and functions' function. The parsing group could be an irregular list, which might include some grids that are not necessary to be compared. For example, we use the 10 grids to represent the unnecessarily compared grids because the normal distribution of the grids is 0-9, the parsed object is [[1,1,10,10],[1,1,10,10],[1,1,1,1],[1,1,1,1]] represent a square consisted of 1 but lost the upper right 2*2 square. When parsing or moving, the operation should ignore the 10's grids. You may consider more properties according to different question. And when parsing, the pattern in input grids might be sheltered, which means the patterns in input grids and output grids might be part of the full pattern, and you might need to infer where the full patterns are, and use code to accomplish the occlusion relationship. The output of the code might be targets and corresponding point sets. In execution part, you need to use the rule and the parsing sets from parsing part to generate answer grid, input and output are both lists. Sometimes you might need to judge the moving direction step by step, take the falling along the steps as an example, you might need to judge the possible moving direction in every time step for one pattern or grid. It's only an example, you need to judge the algorithm according to different settings.

Input-output-groundtruth picture will include three grids, from left to right lies the input of the function, the output from code, and the ground truth output expected. The instructions are generated by the former agent, you can follow the instructions to modify the code. 

Relation between the color and the number is provided: 0: white, 1: blue, 2: teal, 3: orange, 4: green, 5: purple, 6: red, 7: yellow, 8: brown, 9: black.

Code should NOT include ' if '\_\_name\_\_' == 'main' '. The code should only include the functions and a execute function to call these functions, whose name should be execute() to let me call the function easily. The input should be a list, and output should be a list, that is, the function should be 'def execute(input\_list : list) $\rightarrow$ list'. Considering safety, do NOT include 'os.system', 'subprocess.call', 'eval(', 'exec(', '\_\_import\_\_', 'open(', 'rmdir', 'remove'. Input matrix will ALWAYS be one input which is a two-dimension list. A possible input is: [[1, 0], [0, 1]], please pay attention. Mention that the code should NOT include endless loop. 

Please follow this format:

REASONING:

[Think step by step about:

1. Which part are the instructions about? Are they align with the visualization picture of the examples and the analogy?

2. How can I modify the code to instantiate the instructions? Do I cover every examples?

3. If the instructions are not clear enough, what algorithms/approaches can I use to instantiate the instruction? Move, rotate, mirror, copy a certain structure, or combination of above? I should consider all of the examples.

4. What are the key steps in the implementation of modifying the function? This will help me follow the instructions.
]

CODE:

[Your complete, executable Python code here - no markdown formatting]

EXPLANATION:

[Brief explanation of how the code works]
\end{tcolorbox}

Code refinement after introspection on questions system prompt:
\begin{tcolorbox}[colback=green!5!white,colframe=green!50!black,arc=4mm]
You are a helpful assistant that is good at solving Abstraction and Reasoning Corpus (ARC) Test. Now I will provide you an ARC question, the question can be analogous to objects or phenomena in real world. Now I will provide you the analogy of the question along with its reason. The code of which has been finished, and after testing I find that the code is right on all examples. However, when the code is executed on questions, I find something weird, where is not align with the examples and the analogy. Also I will provide you the instructions of the directions how to improve the code. Please rectify the code.

In the following part, I will provide you the analogy of the question, the code of the function, explanation of how the old code work, the visualization of the examples, the matrix of the examples, the visualization of the question input and the corresponding output of the input from the code along with their matrices.

The code includes parsing part and execution part, and at last I will call execute(input: list) $\rightarrow$ list to get my output. the explanation is expected to inform how the code works and functions' function. The parsing group could be an irregular list, which might include some grids that are not necessary to be compared. For example, we use the 10 grids to represent the unnecessarily compared grids because the normal distribution of the grids is 0-9, the parsed object is [[1,1,10,10],[1,1,10,10],[1,1,1,1],[1,1,1,1]] represent a square consisted of 1 but lost the upper right 2*2 square. When parsing or moving, the operation should ignore the 10's grids. You may consider more properties according to different question. And when parsing, the pattern in input grids might be sheltered, which means the patterns in input grids and output grids might be part of the full pattern, and you might need to infer where the full patterns are, and use code to accomplish the occlusion relationship. The output of the code might be targets and corresponding point sets. In execution part, you need to use the rule and the parsing sets from parsing part to generate answer grid, input and output are both lists. Sometimes you might need to judge the moving direction step by step, take the falling along the steps as an example, you might need to judge the possible moving direction in every time step for one pattern or grid. It's only an example, you need to judge the algorithm according to different settings.

In the visualization of examples, every column is an input-output pair, the first row is input and the second row is output. In the visualization of question input and output from code, the left grid is the question, and the right grid is the output of the code. The code includes parsing part and execution part, and at last I will call execute(input: list) $\rightarrow$ list to get my output. the explanation is expected to inform how the code works and functions' function. Considering safety, code do NOT include 'os.system', 'subprocess.call', 'eval(', 'exec(', '\_\_import\_\_', 'open(', 'rmdir', 'remove'. Input matrix will ALWAYS be one input which is a two-dimension list. A possible input is: [[1, 0], [0, 1]], please pay attention. Mention that the code should NOT include endless loop.

Please ensure that the dimensions of the input and output are the same. Please do NOT return [output] like this where the output in the square brackets has the same dimension as the input.

Relation between the color and the number is provided: 0: white, 1: blue, 2: teal, 3: orange, 4: green, 5: purple, 6: red, 7: yellow, 8: brown, 9: black.

Please follow this format:

REASONING:

[Think step by step about:

1. Which part are the instructions about? Are they align with the visualization picture of the examples and the analogy?

2. How can I modify the code to instantiate the instructions? Do I cover every examples at the same time rectify the questions?

3. If the instructions are not clear enough, what algorithms/approaches can I use to instantiate the instruction? Move, rotate, mirror, copy a certain structure, or combination of above? I should consider all of the examples and the questions.

4. Will the rectification of the code change the outputs of the examples? The examples are right already now, therefore I don't want them to be changed.

5. What are the key steps in the implementation of modifying the function? This will help me follow the instructions.]

CODE:

[Your complete, executable Python code here - no markdown formatting]

EXPLANATION:

[Brief explanation of how the code works]
\end{tcolorbox}

\section{Unveiling the Reasoning Dynamics of \textsc{ConceptSIR}}
\label{appsec:conceptsirresultdetail}

In this section we will show the detailed results of \textsc{ConceptSIR}. The results include the output grid in visual and the response of LLMs. All tests are finished through Openrouter.

\subsection{Quantitative analysis.}
\label{appsubsec:conceptsirquantitative}
In our experiments, the “small subset” refers to 17 ARCANA tasks spanning 11 categories. Among these 17 tasks, 10 were solved by none of the four LLMs (0\% accuracy for GPT 5.2, Gemini 3 Pro, Grok 4.1 Fast, Claude Opus 4.5), and 2 had low accuracy for the same LLMs. For each task, we ran ConceptSIR with a budget of 5 loops and one run per task, and we counted success only under strict exact-match correctness (no typo handling).

The tasks are: \texttt{bead\_chain1}, \texttt{candle1}, \texttt{clock1}, \texttt{clock4}, \texttt{heat\_water1}, \texttt{refraction2}, \texttt{tetris1}, \texttt{snake1}, \texttt{tip1}, \texttt{water\_and\_trench1}, \texttt{pulley3}, \texttt{cat1}, \texttt{Kepler\_second\_law\_or\_area1}, \texttt{plastic\_bag1}, \texttt{communicating\_vessels1}, \texttt{shoes1}, and \texttt{basket1}.

Under this setting, ConceptSIR solved 9/17 tasks, covering 7 categories. Among the solved tasks, the average number of loops used until success was 2.22. Notably, candle1 and refraction2 are among the 10 tasks with 0\% accuracy in Table 1, yet were solved by ConceptSIR under strict evaluation. Additionally, bead\_chain1 and water\_and\_trench1 are tasks where only GPT-5.2 (among the four LLMs in Table 1) succeeded, while ConceptSIR also succeeded. The remaining 5 solved tasks are ones that were also solved by both GPT-5.2 and Gemini 3 Pro. Beyond solve/not-solve outcomes, we also observed that ConceptSIR improves proposal efficiency on several tasks. For example, in our LLM tests, GPT-5.2 solved bead\_chain1 once within 12 proposals, whereas ConceptSIR solved it twice within 2 loops (9 proposals in total).We additionally provide the accuracy of direct LLMs and human on this subset in Table \ref{tab:subset}.

\begin{table*}[ht]
\centering
\small
\setlength{\tabcolsep}{8pt}
\begin{tabular}{l|l|c|c}
\toprule
Model & Setting & Response Accuracy & Task Coverage \\
\midrule

\multirow{3}{*}{GPT-5.2}
& Language
& 19.61\% / 19.61\%
& 41.18\% / 41.18\% \\
& VL
& 21.57\% / 24.51\%
& 35.29\% / 47.06\% \\
& Overall
& 20.59\% / 22.06\%
& 41.18\% / 47.06\% \\
\midrule

\multirow{3}{*}{Gemini 3 Pro}
& Language
& 6.86\% / 11.76\%
& 17.65\% / 29.41\% \\
& VL
& 9.80\% / 9.80\%
& 23.53\% / 23.53\% \\
& Overall
& 8.33\% / 10.78\%
& 29.41\% / 35.29\% \\
\midrule

\multirow{3}{*}{Grok 4.1 Fast}
& Language
& 3.92\% / 3.92\%
& 11.76\% / 11.76\% \\
& VL
& 0.98\% / 0.98\%
& 5.88\% / 5.88\% \\
& Overall
& 2.45\% / 2.45\%
& 11.76\% / 11.76\% \\
\midrule

\multirow{3}{*}{Claude Opus 4.5}
& Language
& 0.00\% / 0.00\%
& 0.00\% / 0.00\% \\
& VL
& 0.00\% / 0.00\%
& 0.00\% / 0.00\% \\
& Overall
& 0.00\% / 0.00\%
& 0.00\% / 0.00\% \\
\midrule

ConceptSIR
& Overall
& \textbf{52.94\%} / --
& \textbf{52.94\%} / -- \\

Human
& Overall
& \textbf{60.54\% / 74.15\%}
& \textbf{100.00\% / 100.00\%} \\
\bottomrule
\end{tabular}

\caption{
Performance on the 17-task ConceptSIR subset.
Each result is reported as \textbf{Strict / +Typo}. Language uses language-only chain-of-thought prompting; VL adds an analogical prompt with visual inputs. Each direct LLM generates six responses per task in each setting and twelve responses per task overall.
ConceptSIR and human results are reported only under Overall.
}
\label{tab:subset}
\end{table*}

We acknowledge that 9/17 is not yet fully satisfactory. The limitations might be largely due to the loop budget. With more loops or more attempts, we believe ConceptSIR could solve more ARCANA tasks—in fact, in preliminary experiments with GPT-5 we observed an “evolution”-like improvement process during 15 loops on bead\_chain1.

\subsection{Qualitative analysis.}
\label{appsubsec:conceptsirqualitative}
Here we take three tasks as examples, which are \texttt{refraction2}, \texttt{bead\_chain1}, \texttt{candle1}. Restatement: \texttt{refraction2} and \texttt{candle1} are not solved by all LLM baselines, and \texttt{bead\_chain1} is solved once in 12 attempts by only GPT~5.2. In ConceptSIR, \texttt{refraction2} is solved in 2 reflection steps, \texttt{candle1} is solved in 6 reflection steps. Noting that among the three codes generated by Concept execution stage, two are correct to solve task \texttt{bead\_chain1} in 3 reflection steps.

\begin{figure*}[ht]
    \centering
    \includegraphics[width=0.66\linewidth]{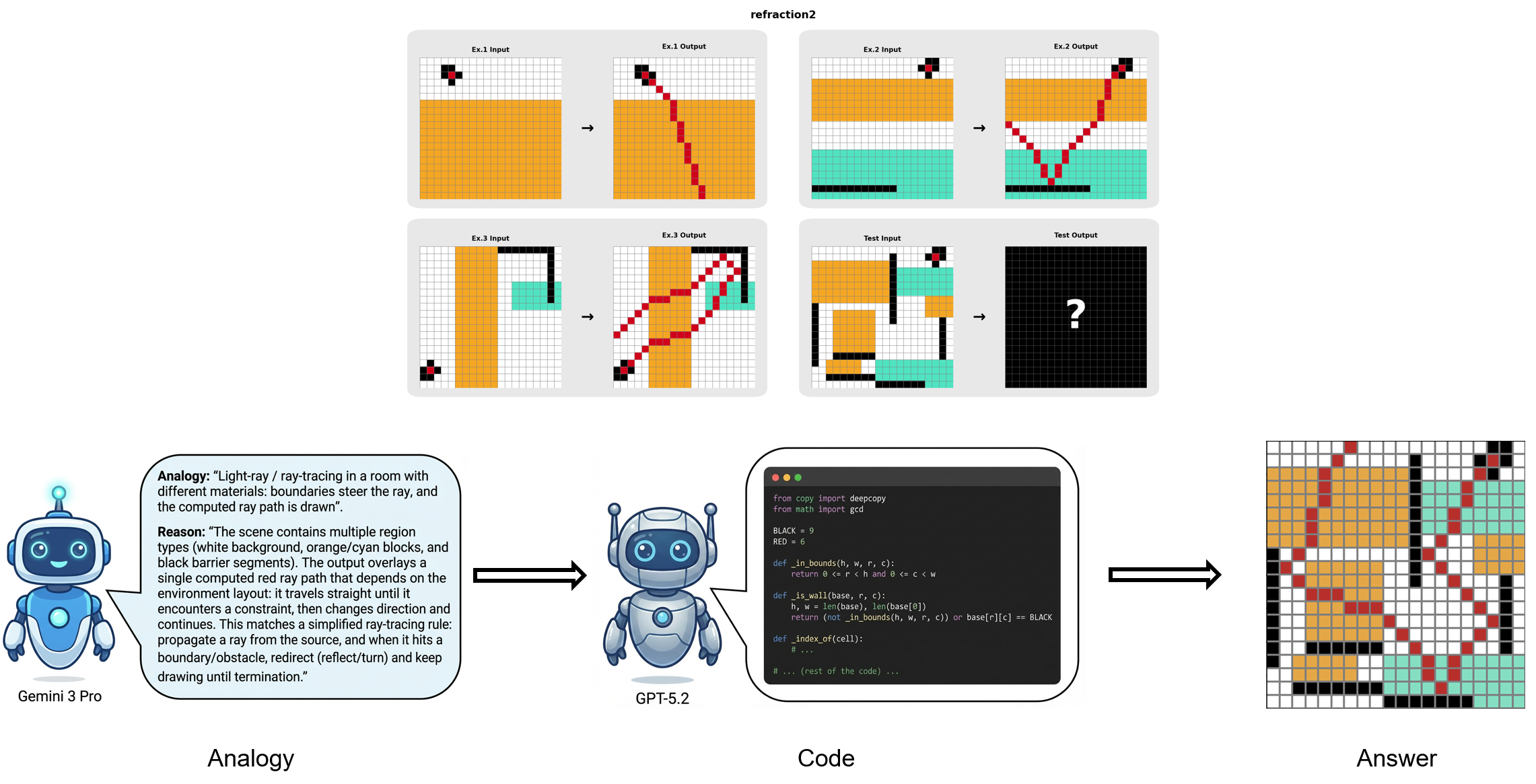}
    \caption{The reasoning trajectory of \texttt{refraction2}. Detailed concept contributes to the correct program generation and output.}
    \label{fig:sirrefract2}
\end{figure*}

As for task \texttt{refraction2} in baseline testing, GPT~5.2 generates the targets ambiguously, like ``light bouncing and slow down". In ConceptSIR, Gemini~3~Pro generate a more accurate target as ``Light-ray / ray-tracing in a room with different materials", which further contributes to the correct program and output from GPT~5.2. The reasoning trajectory is shown in Figure~\ref{fig:sirrefract2}.

\begin{figure*}[t]
    \centering
    \includegraphics[width=0.66\linewidth]{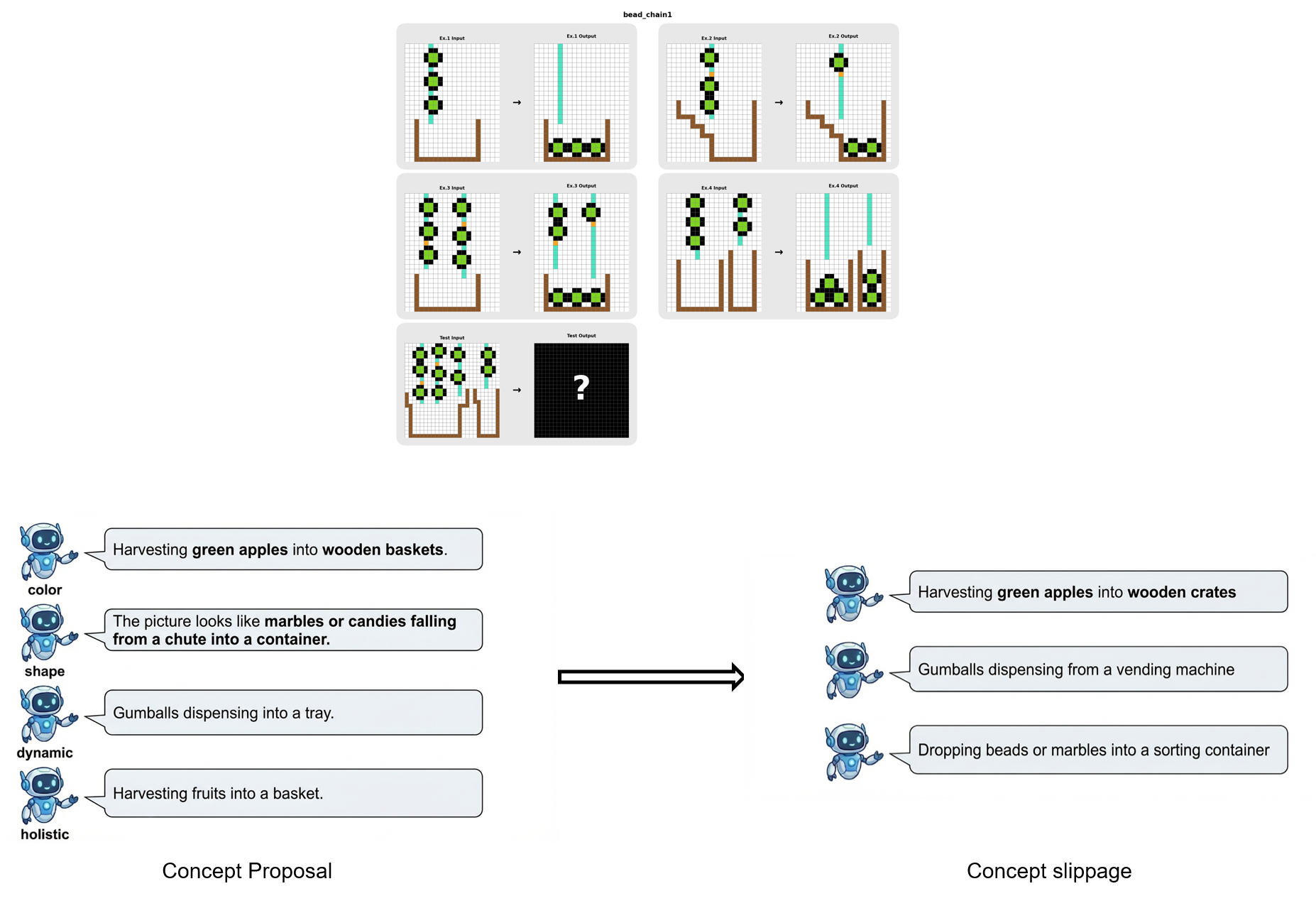}
    \caption{The concept proposal and slippage trajectory of \texttt{bead\_chain1}. Concept proposal and slippage contribute better targets.}
    \label{fig:sirbeadchain}
\end{figure*}

Furthermore, we analyze the reasoning trajectory on \texttt{bead\_chain1}. We note that during baseline testing, GPT~5.2 extensively uses the word ``flower", e.g., ``Flowers below an orange knot are replanted into the brown garden beds", ``Transplanting plants into planter boxes". Noting that the targets cannot be viewed as incorrect ones, they are still different from the targets from Gemini 3 Pro (e.g., ``The task is analogous to `beads sliding down strings'.") that are more similar to humans. In concept proposal, Gemini 3 Pro propose four different targets from color, shape, dynamics, holistic recognition, and generate three targets through concept slippage. The trajectory of concept proposal and slippage can be viewed in Figure \ref{fig:sirbeadchain}.

\begin{figure*}[t]
    \centering
    \includegraphics[width=0.66\linewidth]{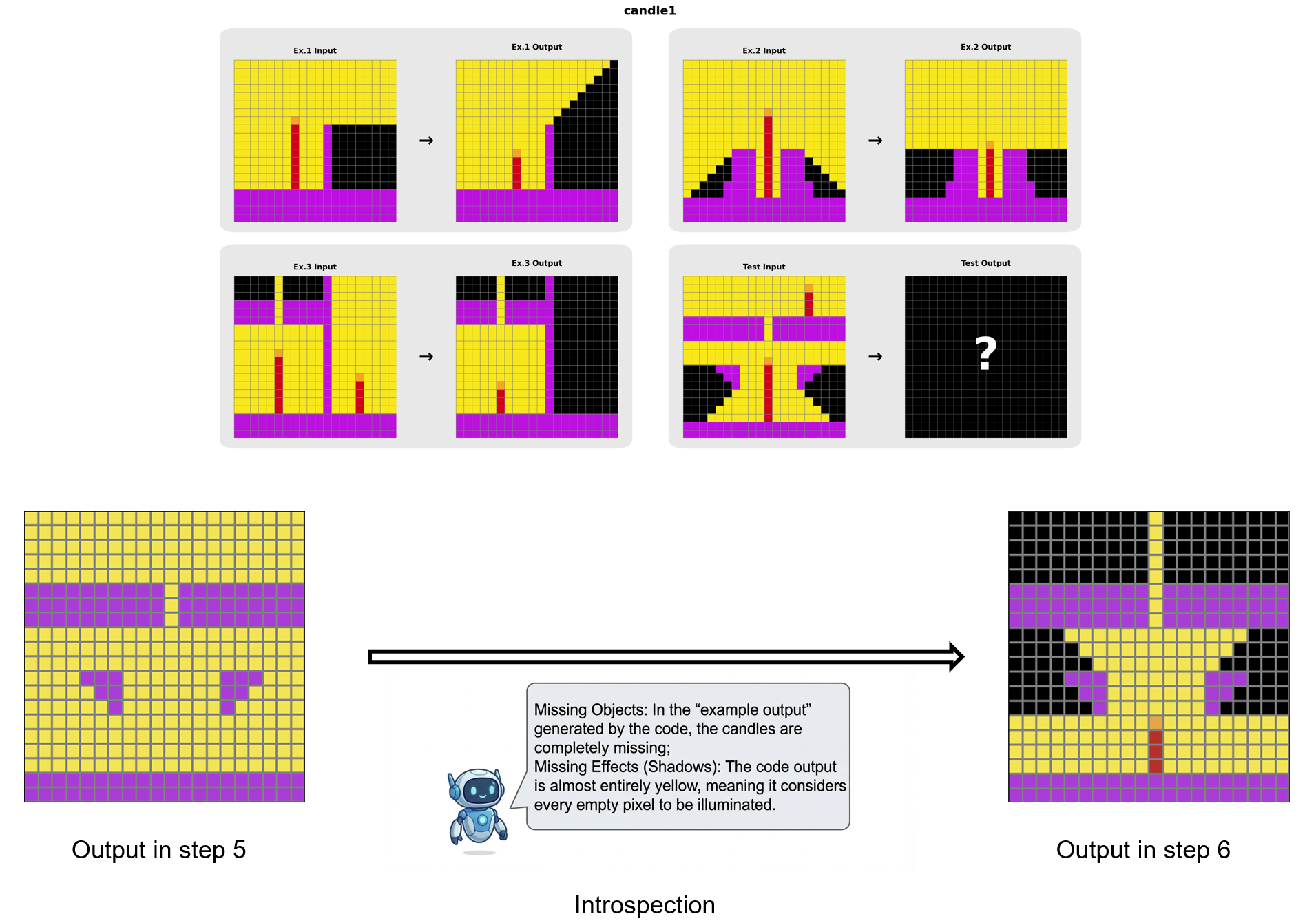}
    \caption{The introspection trajectory of \texttt{candle1}. Introspection filter invalid output.}
    \label{fig:sircandle}
\end{figure*}

Moreover, we analyze the reasoning trajectory on \texttt{candle1}. First, we found in step 5, GPT 5.2 generates a code that solves all of the examples, but fails on the question. The code deletes all of the candles, which is absolutely not correct. Accordingly, during the introspection, LLMs response indicates that: Missing Objects: In the "example output" generated by the code, the candles are completely missing; Missing Effects (Shadows): The code output is almost entirely yellow, meaning it considers every empty pixel to be illuminated.". Introspection effectively filter the invalid output, and finally the code is refined to generate the correct answer. The introspection could be viewed in Figure \ref{fig:sircandle}.

During testing on ConceptSIR, we notice that the analogy accuracy of Gemini drops. We assume that during baseline testing, Gemini 3 Pro analyzes the grids in detail and notices the relations (e.g., multiple relationship) on grids, and associates it to more complicated concepts. This indicates the potential increase space for concept proposal and slippage.

\section{Categories and tasks visualization in ARCANA}
\label{appsec:arcanadetail}
Finally, for better understanding the diversity of ARCANA, we provide several tasks in visual. Here we list all categories and the detailed tasks in Table~\ref{tab:arcana}, and several tasks are shown behind.

\begin{table*}[htpb]
\small
    \centering
\begin{tabular}{|c|c|c|}
\hline
Category & num & Task names\\
\hline
1d\_wave & 2 & wave2, wave\_interference1 \\
\hline
\multirow{2}*{2d\_wave} & \multirow{2}*{5} & Doppler1, wave1, wave\_diffraction1,\\ 
~ & ~ & wave\_diffraction2, wave\_diffraction3 \\
\hline
3d\_cut & 3 & cut\_sushi1, cut\_sushi2, cut\_sushi3 \\
\hline
\multirow{3}*{3d\_move} & \multirow{3}*{6} & brush1, mop1, mop2, pipes1, \\
~ & ~ &  table\_and\_stick\_and\_cuboid1, \\
~ & ~ &  table\_and\_stick\_and\_cuboid2\\
\hline
3d\_rotate & 5 & book1, door1, lid1, lid2, roll\_paper1 \\
\hline
\multirow{2}*{3d\_shadow} & \multirow{2}*{5} & cuboid\_and\_shadow1, cuboid\_and\_shadow2,  \\
~ & ~ & cuboid\_and\_shadow3, ring1, stack1\\
\hline
animal & 3 & cat1, rabbit1, duck1 \\
\hline
bottle & 5 & bottle1, bottle2, bottle3, waterstream1, waterstream2 \\
\hline
\multirow{3}*{causal\_move} & \multirow{3}*{11} & hook1, jellyfish1, toothbrush\_and\_toothpaste1, \\

~ & ~ &  pen1, rocket1, shoot1,wind\_and\_balloon1,  \\
~ & ~ &  wind\_and\_balloon2, gun\_revolver1, fish1\\
\hline
\multirow{2}*{causal\_updown} & \multirow{2}*{6} & magnetic1, magnetic2, magnetic3,\\
~ & ~ & UFO\_or\_vacuum\_cleaner1, sunset1, sunset2 \\
\hline
claw & 6 & chopsticks1, claw1, claw2, claw3, fork1, fork2 \\
\hline
elec\_clock & 2 & clock4, clock5 \\
\hline
electricity & 4 & car2, charge1, charge2, charge3 \\
\hline
emoji & 2 & happy\_and\_sad1, smile1 \\
\hline
gravity\_and\_liquid & 4 & dye1, dye2, dye3, dye4 \\
\hline
\multirow{2}*{gravity\_and\_solid} & \multirow{2}*{5} & bead\_chain1, car1, fall\_and\_dig1,\\
~ & ~ & plastic\_bag1, portal1 \\
\hline
heat & 2 & heat\_transfer1, melt3 \\
\hline
\multirow{3}*{light} & \multirow{3}*{7} & candle1, flashlight1, flashlight2, switch1,  \\
~ & ~ & flashlight\_and\_stained\_glass1, light1,\\
~ & ~ & switch\_and\_water\_pipe1\\
\hline
move\_and\_shelter & 7 & basket1, cup\_and\_ball1, gloves1, hat1, shoes1, window1, window2 \\
\hline
\multirow{6}*{others} & \multirow{5}*{27} & arrow1, candy\_crush1, cake\_and\_candle1, cigar1, domino1,  \\

~ & ~ & fire\_extinguisher1, kaleidoscope1, Kepler\_second\_law\_or\_area1, lever1, \\

~ & ~ &    lock1, logic\_and\_or\_not1, gold\_miner1,  moon\_and\_tide1, \\
~ & ~ &   magnetic\_or\_chemical\_reaction1, matryoshka\_doll1, nuts1, \\
~ & ~ &  pin-hole\_imaging1, golf1, puzzle\_bubble1, rainbow\_spring1, rust1,  \\
~ & ~ & sandglass1, snake1, tetris1, vine1, water\_and\_trench1, zuma1\\

\hline
parabola & 3 & parabola1, plane\_and\_bomb1, portal\_and\_parabola1 \\
\hline
pendulum & 4 & ballistic\_pendulum1, bell1, newton\_cradle1, newton\_cradle2 \\
\hline
\multirow{2}*{plant} & \multirow{2}*{9} & fireworks1, flower1, flower2, bomb\_dropping1, tree1, tree2, \\
~ & ~ &  tree\_felling1, rain\_and\_umbrella1, stars1\\
\hline
pulley & 3 & pulley1, pulley2, pulley3 \\
\hline
\multirow{2}*{reflect\_and\_refract} & \multirow{2}*{5} & reflection1, reflection2, refraction1,\\
~ & ~ & refraction2, refraction\_and\_reflection1 \\
\hline
\multirow{2}*{rotate\_and\_count} & \multirow{2}*{11} & gear1, gear2, waterwheel1, waterwheel2, yoyo1,\\
~ & ~ & yoyo2, belt1, belt2, roll1, roll2, roll3 \\
\hline
rotate\_and\_rotate & 3 & galaxy1, mechanical\_arm\_rotate1, pocket\_watch1 \\
\hline
\multirow{2}*{rotate\_and\_shelter} & \multirow{2}*{5} & disco\_ball1, fan1, wind\_and\_flag\_and\_windmill1, \\
~ & ~ &  windmill1, wind\_and\_flag\_and\_cloud1 \\
\hline
rotate\_and\_tranfsform & 3 & hammer1, hammer2, telephone1 \\
\hline
scissors & 3 & scissors1, scissors2, knife\_and\_cake1 \\
\hline
shelter\_then\_move & 6 & cup1, cup2, drawer1, knife1, shelf1, toaster1 \\
\hline
traditional\_clock & 3 & clock1, clock2, clock3 \\
\hline
\multirow{2}*{water\_deeper\_concept} & \multirow{2}*{7} & archimedes1, archimedes2, buoyancy1, buoyancy2,\\
~ & ~ &  water\_and\_spoon1, water\_and\_spoon2, water\_well1 \\
\hline
\multirow{3}*{water\_level} & \multirow{3}*{12} & communicating\_vessels1, gasflow1, squeeze\_sponge1, \\

~ & ~ & straw1, straw2, syringe1, tip1, piston1, waterflow1, \\
~ & ~ & waterflow2, water\_dispenser1, water\_dispenser2\\
\hline
\multirow{2}*{water\_transform} & \multirow{2}*{6} & heat\_water1, melt1, melt2, melt4,  \\
~ & ~ & melt\_and\_frozen1, water\_and\_water\_transformation1\\
\hline
\end{tabular}
\caption{The category names and tasks in ARCANA.}
\label{tab:arcana}
\end{table*}

\begin{figure}[ht]
    \centering
    \includegraphics[width=0.85\linewidth]{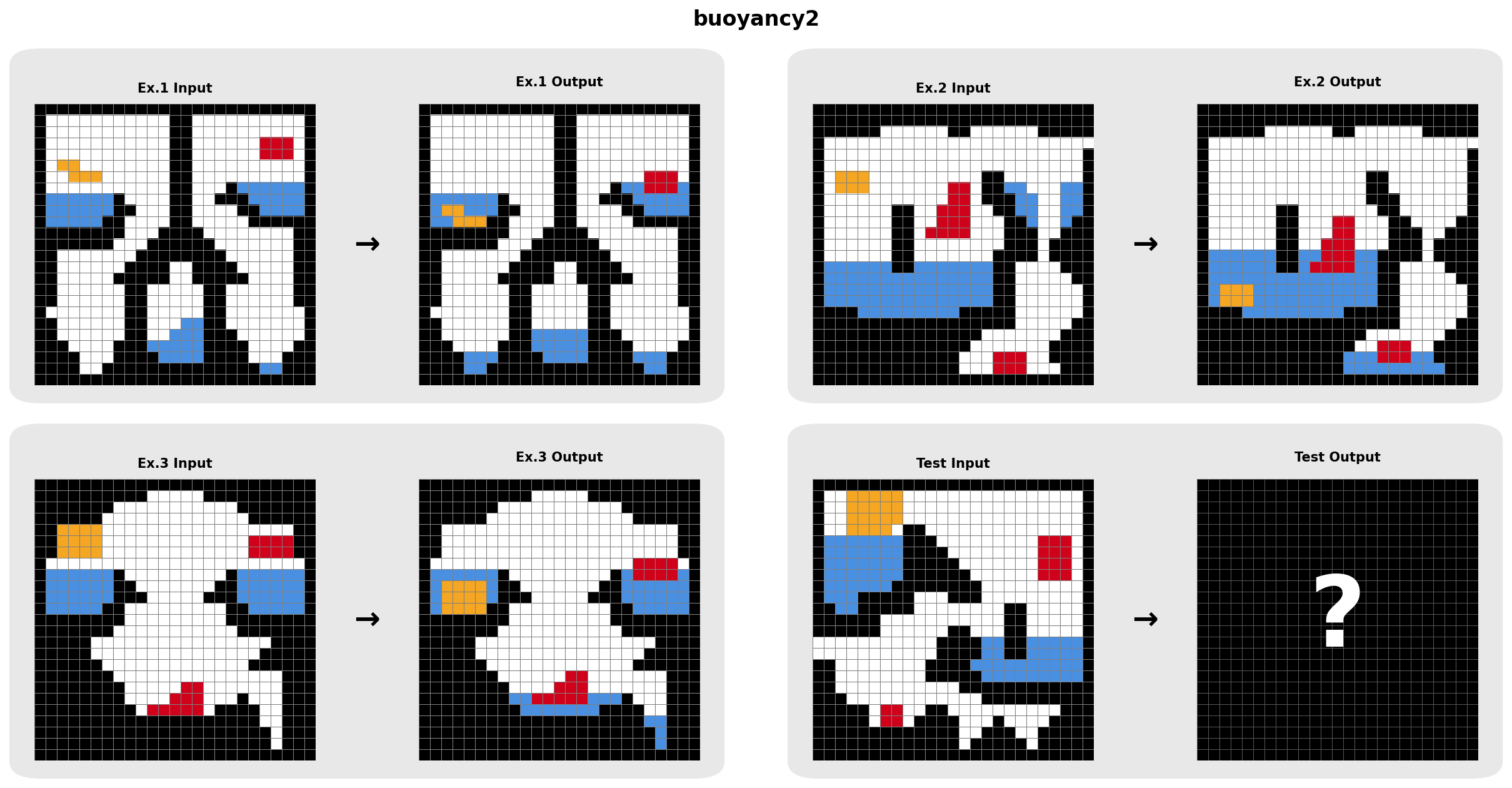}
    \caption{\texttt{buoyancy2} in \texttt{water\_deeper\_concept} category.}
    \label{fig:arcanabuoyancy2}
\end{figure}

\begin{figure}[ht]
    \centering
    \includegraphics[width=0.85\linewidth]{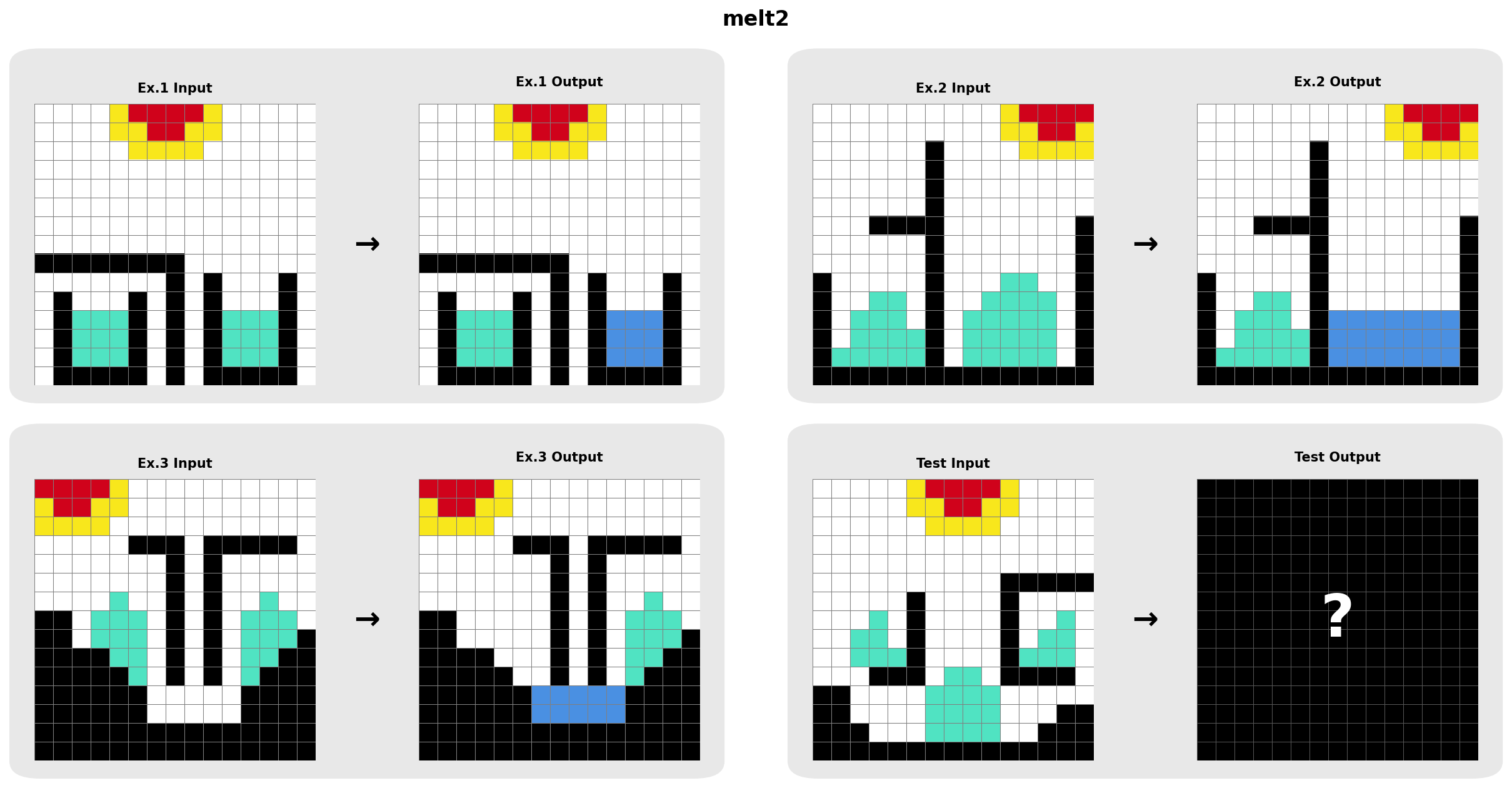}
    \caption{\texttt{melt2} in \texttt{water\_transform} category.}
    \label{fig:arcanamelt2}
\end{figure}

\begin{figure}[ht]
    \centering
    \includegraphics[width=0.85\linewidth]{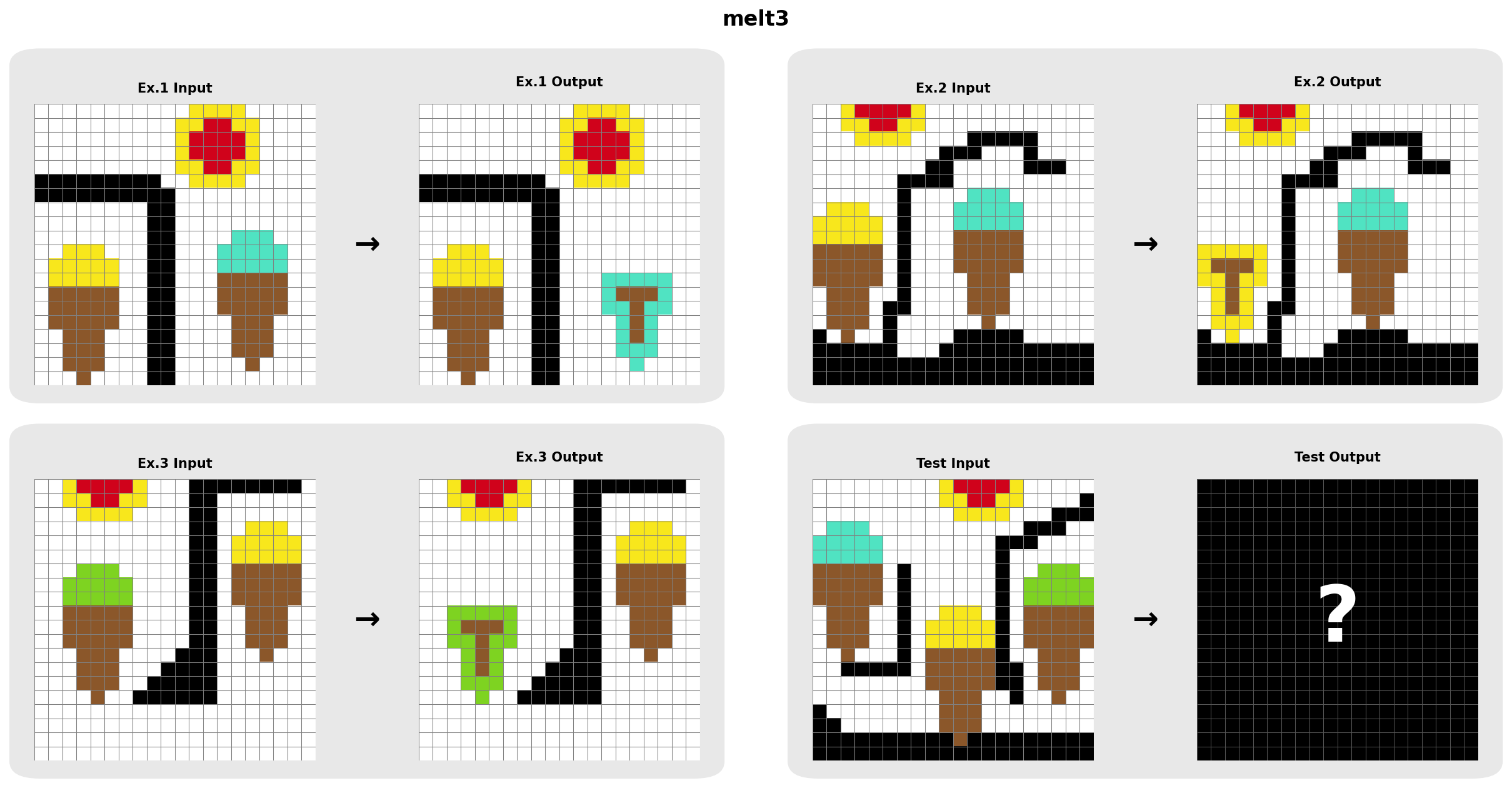}
    \caption{\texttt{melt3} in \texttt{heat} category.}
    \label{fig:arcanamelt3}
\end{figure}

\begin{figure}[ht]
    \centering
    \includegraphics[width=0.85\linewidth]{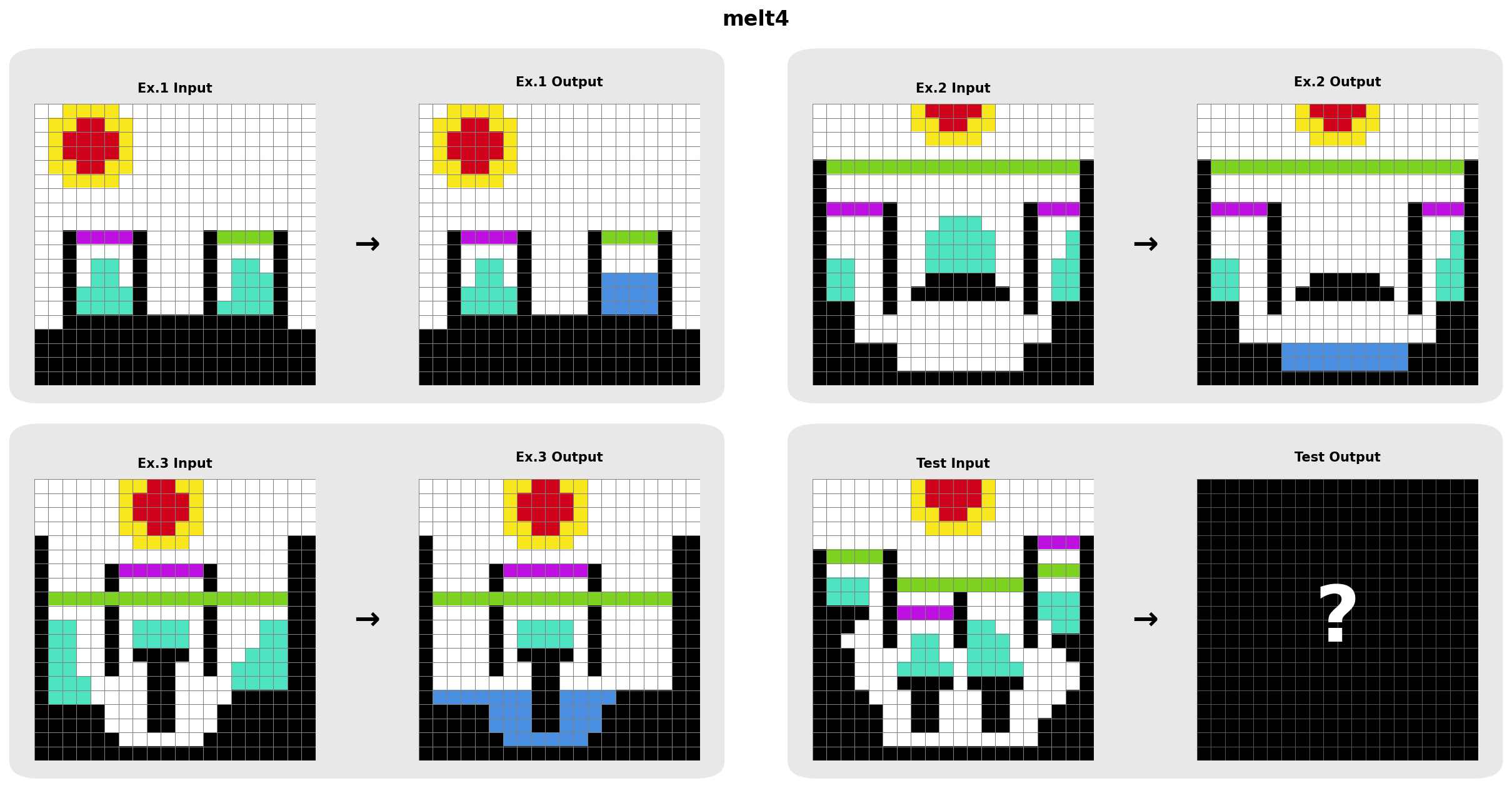}
    \caption{\texttt{melt4} in \texttt{water\_transform} category.}
    \label{fig:arcanamelt4}
\end{figure}

\begin{figure}
    \centering
    \includegraphics[width=0.7\linewidth]{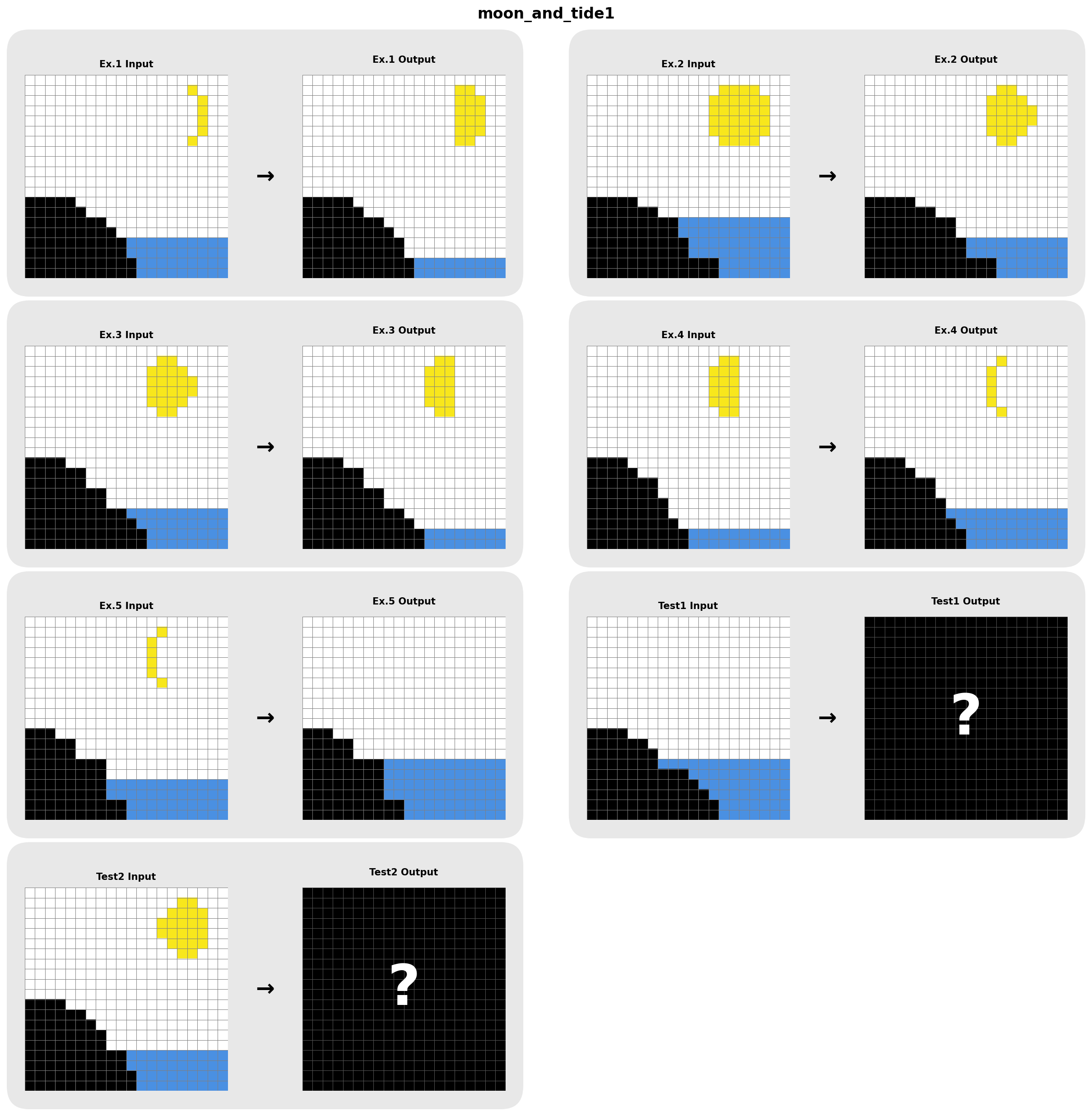}
    \caption{\texttt{moon\_and\_tide1} in \texttt{others} category.}
    \label{fig:arcanamoon}
\end{figure}

\begin{figure}
    \centering
    \includegraphics[width=0.7\linewidth]{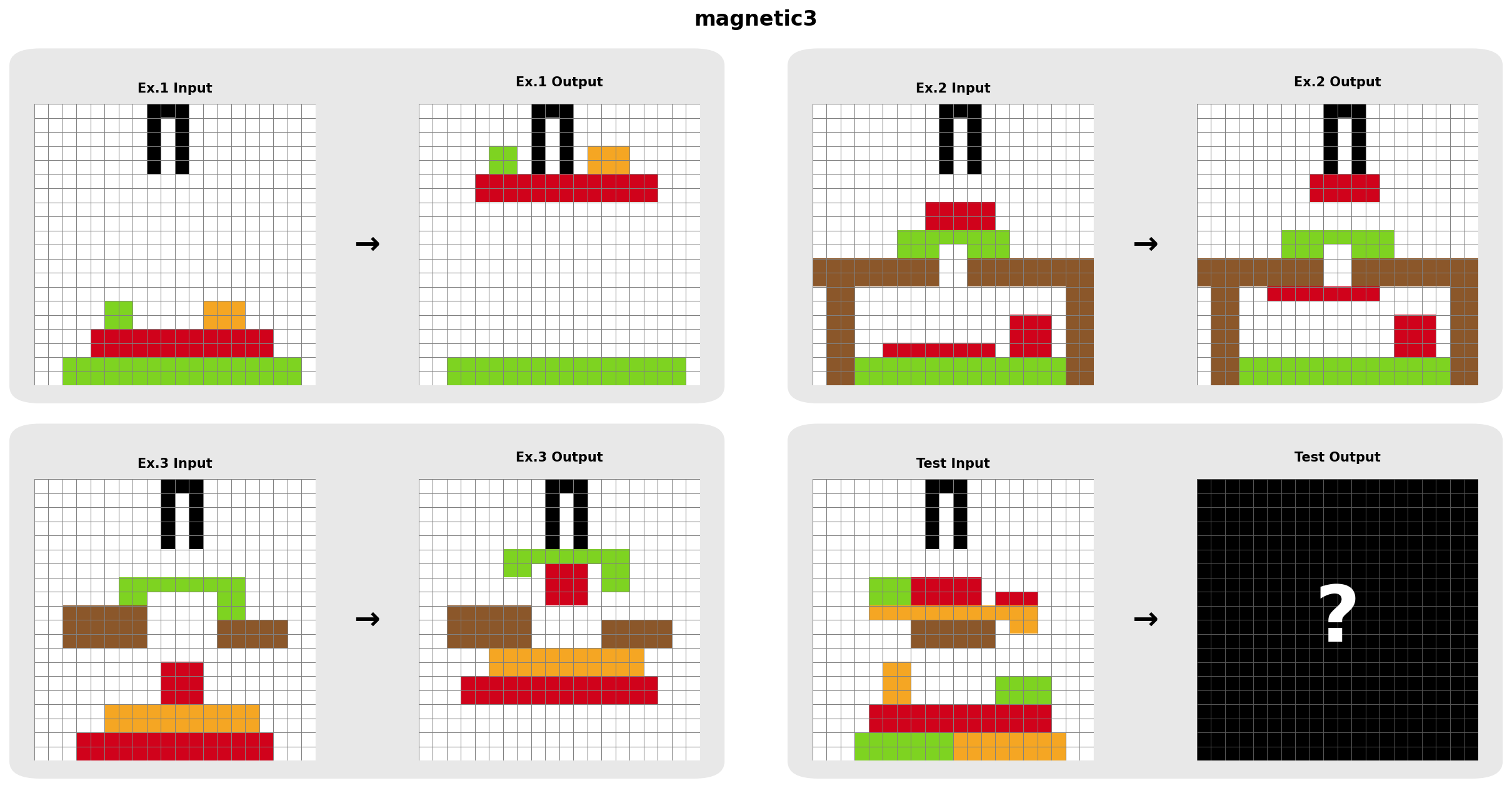}
    \caption{\texttt{magnetic3} in \texttt{causal\_updown} category.}
    \label{fig:arcanamagnetic}
\end{figure}

\begin{figure}
    \centering
    \includegraphics[width=0.7\linewidth]{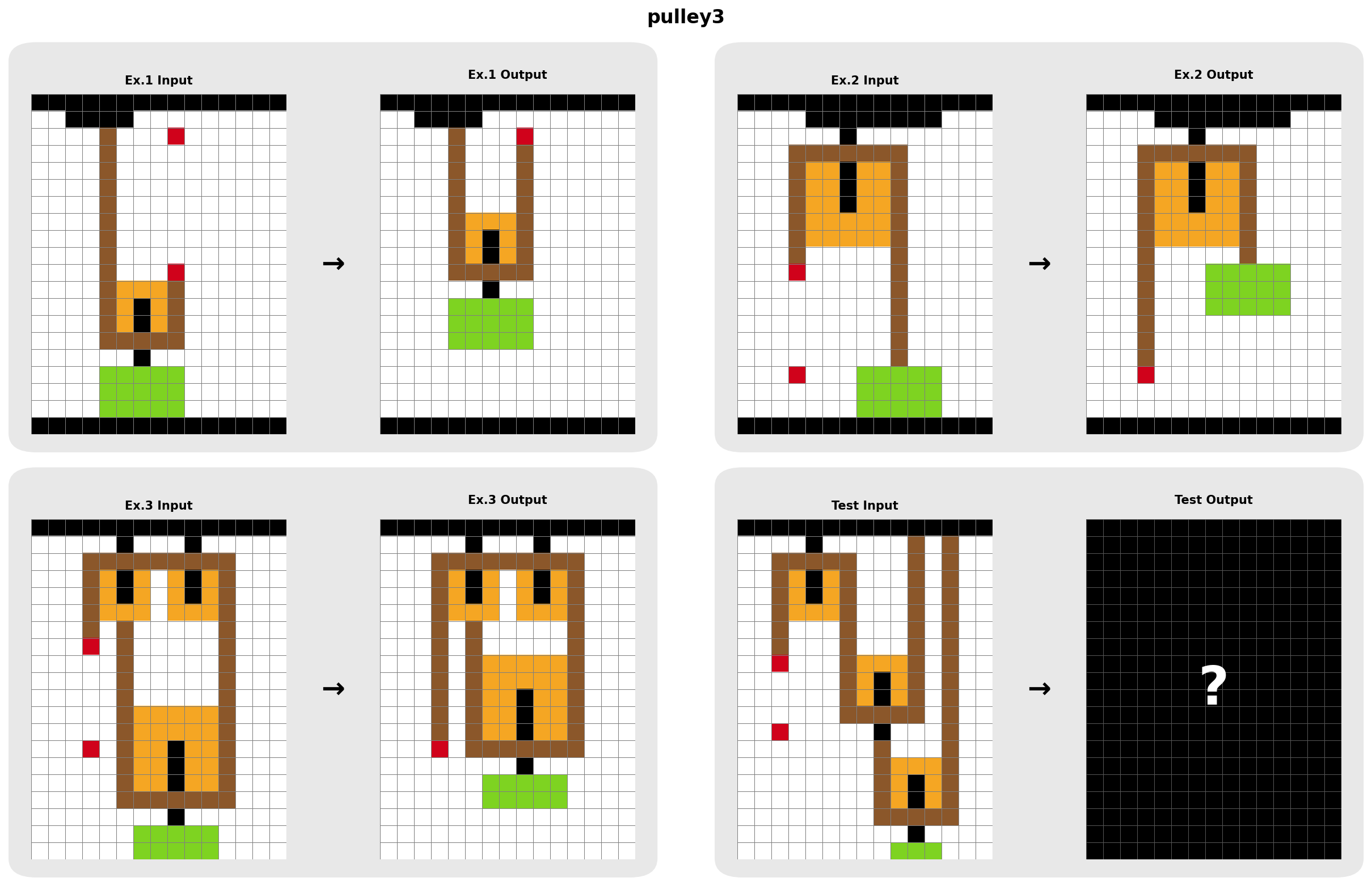}
    \caption{\texttt{pulley3} in \texttt{pulley} category.}
    \label{fig:arcanabuoyancy}
\end{figure}

\begin{figure}
    \centering
    \includegraphics[width=0.7\linewidth]{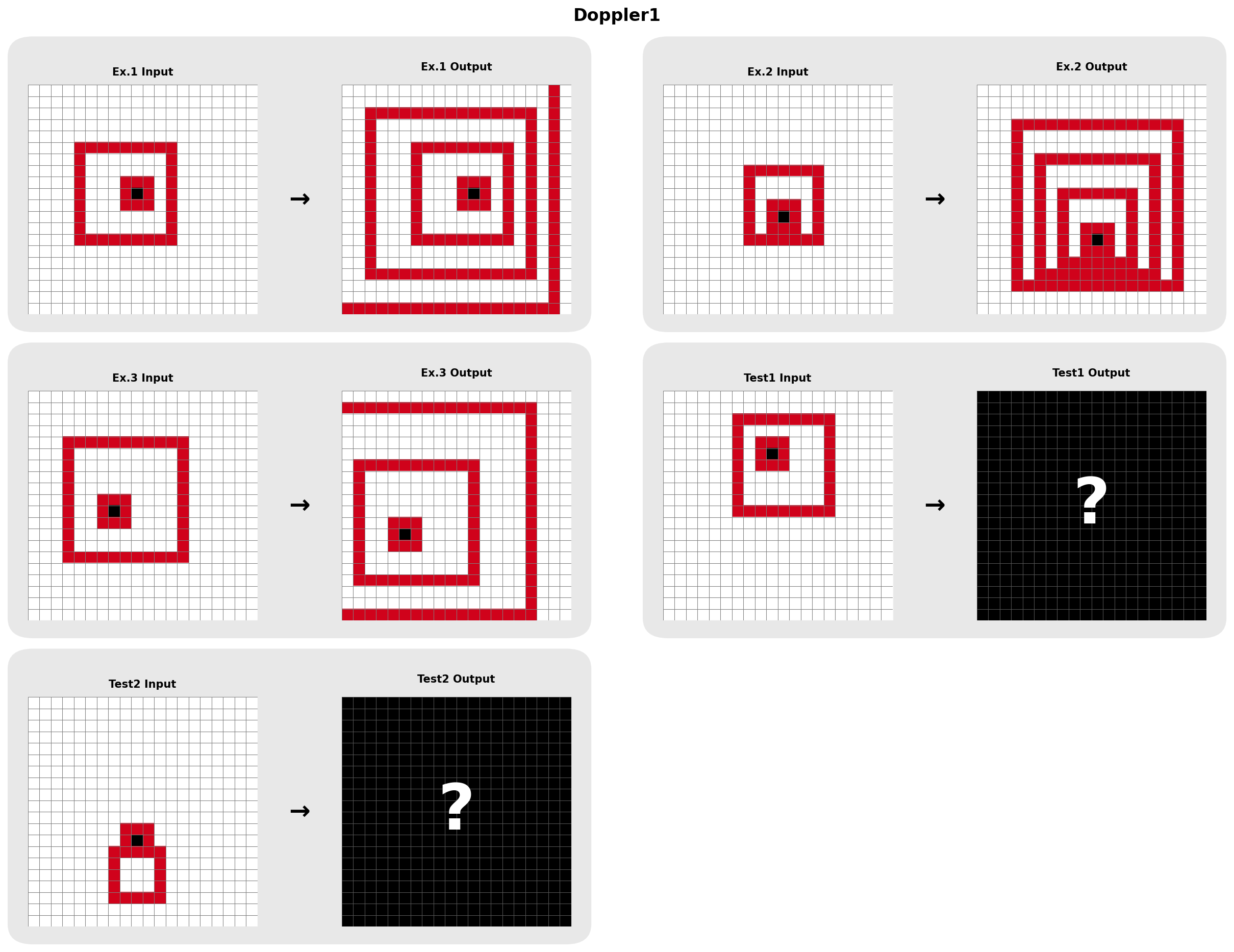}
    \caption{\texttt{Doppler1} in \texttt{2d\_wave} category.}
    \label{fig:arcanadoppler}
\end{figure}

\end{document}